%% file: main_springer.tex
\documentclass[sn-apa]{sn-jnl}

\jyear{2021}%
\input{math_commands.tex}

\newcommand{\pap}{ConCURL}
\newcommand{\iten}{ImageNet-10}
\newcommand{\idogs}{ImageNet-Dogs}
\newcommand{\stl}{STL-10}
\newcommand{\cten}{CIFAR-10}
\newcommand{\chundred}{CIFAR100-20}

\usepackage{times}
\usepackage{color,soul}
\usepackage{url}
\usepackage{epstopdf}
\usepackage{graphicx}
\usepackage{subfigure}
\usepackage{booktabs}
\usepackage[justification=centering]{caption}
\usepackage{algorithm}
\usepackage{algpseudocode}
\usepackage{upquote} 

\usepackage{listings}
\usepackage{tablefootnote}

\usepackage{multirow}
\usepackage{tablefootnote}

\usepackage{xcolor}
\definecolor{codeblack}{rgb}{0.0,0.0,0.0}
\newtheorem{definition}{Definition}

\theoremstyle{thmstyleone}%
\theoremstyle{thmstyletwo}%

\theoremstyle{thmstylethree}%

\begin{document}

\title[Representation Learning for Clustering via Building Consensus]{Representation Learning for Clustering via Building Consensus}


\author*[1]{\fnm{Aniket Anand} \sur{Deshmukh}}\email{aniketde@umich.edu}

\author[2]{\fnm{Jayanth Reddy} \sur{Regatti}}\email{regatti.1@osu.edu}
\equalcont{Work was done while author was at Microsoft}

\author[1]{\fnm{Eren} \sur{Manavoglu}}\email{ermana@microsoft.com}

\author[1]{\fnm{Urun} \sur{Dogan}}\email{urundogan@gmail.com}

\affil*[1]{ \orgname{Microsoft}, \orgaddress{\city{Mountain View}, \postcode{94043}, \state{CA}, \country{USA}}}

\affil[2]{ \orgname{The Ohio State University}, \orgaddress{ \city{Columbus}, \postcode{43210}, \state{OH}, \country{USA}}}


\abstract{ {In this paper, we focus on unsupervised representation learning for clustering of images.} Recent advances in deep clustering and unsupervised representation learning are based on the idea that different views of an input image (generated through data augmentation techniques) must be close in the representation space (exemplar consistency), and/or similar images must have similar cluster assignments (population consistency). We define an additional notion of consistency, \textit{consensus consistency}, which ensures that representations are learned to induce similar partitions for variations in the representation space, different clustering algorithms or different initializations of a single clustering algorithm. We define a clustering loss by executing variations in the representation space and seamlessly integrate all three consistencies (consensus, exemplar and population) into an end-to-end learning framework.
  The proposed algorithm, consensus clustering using unsupervised representation learning (\pap), improves upon the clustering performance of state-of-the-art methods on four out of five image datasets. Furthermore, we extend the evaluation procedure for clustering to reflect the challenges encountered in real-world clustering tasks, such as maintaining clustering performance in cases with distribution shifts. We also perform a detailed ablation study for a deeper understanding of the proposed algorithm. The code and the trained models are available at \href{https://github.com/JayanthRR/ConCURL_NCE}{https://github.com/JayanthRR/ConCURL_NCE}.}




\maketitle

\section{Introduction}
\label{sec:introduction}
The field of artificial intelligence (AI) advanced significantly in the previous decade due to developments in deep learning \citep{lecun2015deep}. 
In the early years of this field, deep learning methods exhibited stellar supervised learning performance, where each data sample was coupled with a ground truth (labeled data), e.g., each image was associated with a category. Unfortunately, generating labeled datasets is time consuming and expensive, and there may not be enough experts to label the data at hand (e.g., medical images). The straightforward solution is to use clustering for large-scale problems. 

In this work, we focus on representation learning for the unsupervised learning task of clustering images. Clustering is a ubiquitous task and has been actively used in many different scientific and practical pursuits \citep{frey2007clustering, masulli1999fuzzy, jain1999data, xu2005survey}. Clustering algorithms do not learn representations and are hence limited to data for which we have a good representation available.

Advancements in deep learning techniques have enabled the end-to-end learning of rich image representations for supervised learning. For the purposes of clustering, however, such features learned via supervised learning cannot be obtained due to lack of available labels. Therefore, supervised learning approaches fall short of providing a solution. Self-supervised learning addresses the issue of learning representations without labeled data. Self-supervised learning is a subfield of unsupervised learning in which the main goal is to learn general-purpose representations by exploiting user-defined tasks (pretext tasks) \citep{wu2018unsupervised,zhuang2019local,he2020momentum,zhuang2019local,chen2020simple,grill2020bootstrap,caron2020unsupervised}.
Representation learning algorithms have been shown to achieve good results when evaluated using a linear evaluation protocol, semisupervised training on ImageNet, or transfer to downstream tasks. A straightforward solution to the clustering problem is to use the features obtained via self-supervised learning and apply an out-of-the-box clustering algorithm (such as k-means) to compute data clusters. However, the performance of these features for clustering (using an out-of-the-box clustering algorithm) is not known, and as seen in our results, these features may be improved for clustering purposes.

On the other hand, deep clustering involves simultaneously learning cluster assignments and features using deep neural networks. Simultaneously learning the feature spaces with a clustering objective in deep clustering may lead to degenerate solutions, which until recently limited end-to-end implementations of clustering with representation learning approaches \citep{caron2018deep}. Subsequently, several works have been developed \citep{xie2016unsupervised, caron2018deep,shah2018deep, ji2019invariant,niu2020gatcluster,wu2019deep,huang2020deep,taoclustering}. We provide details regarding some of these works in Section~\ref{sec:relatedwork}. Our previous work shows some encouraging results but we extend the work substantially \citep{regatti2020consensus}. We categorize the current clustering and representation learning works based on the consistency constraints that are used to define their objective functions. We define an additional notion of consistency, consensus consistency, which ensures that representations are learned to induce similar partitions for variations in the representation space, different clustering algorithms or different initializations of a clustering algorithm. We use consensus consistency and propose an end-to-end learning approach that outperforms other end-to-end learning methods for image clustering. We summarize our contributions as follows:
\begin{enumerate}
    \item We introduce different notions of consistency (exemplar, population and consensus) that are used in unsupervised representation learning. 
    \item We propose a novel clustering algorithm that incorporates the above three consistency constraints and can be trained in an end-to-end way. An ensemble is generated in the consensus clustering objective by performing random transformations on the underlying embeddings. We combine several methods, which is not trivial, and this combination, along with our new consensus loss, is novel.  
    \item We show that the proposed algorithm \pap\ (consensus clustering with unsupervised representation learning) outperforms baselines on popularly used computer vision datasets when evaluated with clustering metrics.
	\item We demonstrate the clustering abilities of trained models under a data shift and argue for the need for different evaluation metrics for deep clustering algorithms.
    \item We study the impacts of various hyperparameters, data augmentation methods, and image resolutions on the clustering ability of the proposed algorithm.  
\end{enumerate}

\subsection{Related Work}
\label{sec:relatedwork}
\textbf{Self-Supervised Learning:}
Self-supervised learning is used to learn representations in an unsupervised way by defining some pretext tasks. There are many different flavors of self-supervised learning, such as instance discrimination (ID) tasks \citep{wu2018unsupervised,zhuang2019local} and contrastive techniques \citep{he2020momentum,chen2020simple}. In ID tasks, each image is considered its own category so that the learned embeddings are well separated \citep{wu2018unsupervised}. Building on the ID task, \cite{zhuang2019local} proposed a local aggregation (LA) method based on a robust clustering objective (using multiple runs of k-means) to move statistically similar data points closer in the representation space and dissimilar data points further away. In contrastive techniques such as simple contrastive learning of visual representations (SimCLR)~\citep{chen2020simple} and momentum contrast (MoCo)~\citep{he2019moco}, representations are learned by maximizing the agreement between different the augmented views of the same data example (known as positive pairs) and minimizing the agreement between the augmented views of different examples (known as negative pairs). Recent works, including Bootstrap Your Own Latent (BYOL) \citep{grill2020bootstrap} and swapping assignments between multiple views (SwAV) \citep{caron2020unsupervised}, have achieved state-of-the-art results without requiring negative pairs. Although self-supervised learning methods exhibit impressive performance on a variety of problems, it is not clear whether learned representations are good for clustering.

\noindent \textbf{Clustering with Representation Learning:}

DEC \citep{xie2016unsupervised} is one of the first algorithms to show that deep learning can be used to effectively cluster images in an unsupervised manner; this approach uses features learned from an autoencoder to fine-tune the cluster assignments. DeepCluster \citep{caron2018deep} shows that it is possible to train deep convolutional neural networks (DeCNNs) in an end-to-end manner with pseudolabels that are generated by a clustering algorithm. Subsequently, several works \citep{shah2018deep, ji2019invariant,niu2020gatcluster,wu2019deep,huang2020deep} have introduced end-to-end clustering-based objectives and achieved state-of-the-art clustering results.
For example, in the Gaussian attention network for image clustering (GATCluster) \citep{niu2020gatcluster}, training is performed in two distinct steps (similar to \cite{caron2018deep}), where the first step is to compute pseudotargets for a large batch of data and the second step is to train the model in a supervised way using these pseudotargets. Both DeepCluster and GATCluster use k-means to generate pseudolabels that may not scale well. \cite{wu2019deep} proposed deep comprehensive correlation mining (DCCM), where discriminative features are learned by taking advantage of the correlations among the data using pseudolabel supervision and the triplet mutual information among the features. However, DCCM may be susceptible to trivial solutions \citep{niu2020gatcluster}. Invariant information clustering (IIC) \citep{ji2019invariant} maximizes the mutual information between the class assignments of two different views of the same image (paired samples) to learn representations that preserve the commonalities between the views while discarding instance-specific details. It has been argued that the presence of an entropy term in mutual information plays an important role in avoiding degenerate solutions. However, a large batch size is needed for the computation of mutual information in IIC; this process may not be scalable for larger image sizes, which are common in popular datasets \citep{ji2019invariant,niu2020gatcluster}. \cite{huang2020deep} extended the celebrated maximal margin clustering idea to the deep learning paradigm by learning the most semantically plausible clusters through the minimization of a proposed partition uncertainty index. Their pixel intensity clustering (PICA) algorithm uses a stochastic version of this index, thereby facilitating minibatch training. PICA fails to assign a sample-correct cluster when that sample has either high foreground or background similarity to samples in other clusters.
In a more recent approach, contrastive clustering \citep{cons_clus}, a contrastive learning loss (as in SimCLR \citep{chen2020simple}) was adopted along with an entropy term to avoid degenerate solutions. Similarly, \cite{taoclustering} combined ID \citep{wu2018unsupervised} with novel softmax-formulated decorrelation constraints for representation learning and clustering. Their approach outperforms state-of-the-art methods and improves upon the instance discrimination method. Our method also improves upon ID and outperforms the method of~ \cite{taoclustering} on all datasets considered. There are other non-end-to-end approaches, such as SCAN \citep{van2020scan}, which use the learned representations from a pretext task to find the images that are semantically closest to the given image using the nearest neighbors algorithm. Similarly, one more state of the art non-end-to-end approach, SPICE \citep{niu2021spice} divides the clustering network in two parts - one to measure instance level similarity and one to identify cluster level discrepancy.

\section{Consensus Clustering}

One of the distinguishing factors between supervised learning and unsupervised learning is the existence of ground truth labels that construct a global constraint based on examples. In most self-supervised learning methods, the ground truth is replaced with some consistency constraint \citep{chen2020simple}. Without a doubt, the performance of any self-supervised method is a function of the power of the consistency constraint used.
We define two types of consistency constraints: exemplar consistency and population consistency.
\begin{definition}\label{def:exemplar_consistency}
  \textbf{Exemplar consistency}: Representation learning algorithms that learn closer representations (in terms of some distance metric) for different augmentations of the same data point are said to follow exemplar consistency.
\end{definition}
Examples of the usage of exemplar consistency include contrastive learning methods such as MoCo \citep{he2019moco} and SimCLR \citep{chen2020simple}. In these methods, a positive pair of images is defined as any two image augmentations of the same image, and a negative pair consists of any two different images.

\begin{definition}\label{def:weak_population}
\textbf{Population consistency}: Representation learning algorithms that ensure that learned representations satisfy the consistency constraint, where two similar data points or any augmentations of the same data points should belong to the same cluster (or population), are said to follow population consistency.
\end{definition}

Deep Cluster \citep{caron2018deep} is a prominent self-supervised method that utilizes population consistency, i.e., Definition \ref{def:weak_population}, by enforcing a clustering constraint on the input dataset. Please note that each cluster assignment contains data points that are similar to each other. Similarly, SwAV~\citep{caron2020unsupervised} is an example of the population consistency method.

\begin{definition}\label{def:consensus_consistency}
\textbf{Consensus consistency}: Representation learning algorithms that are able to learn representations that induce similar partitions for variations in the given representation space (subsets of features, random projections, etc. ), different clustering algorithms (k-means, Gaussian mixture models (GMMs), etc.) or different initializations of clustering algorithms are said to follow consensus consistency.
\end{definition}

Earlier works on \textit{consensus consistency} did not consider representation learning and used the knowledge reuse framework (see \cite{strehl2002cluster},\cite{ghosh2011cluster}), where the cluster partitions were available (the features were irrelevant) or the features of the data were fixed. For example, \cite{fern2003random} successfully applied random projections to consensus clustering by performing k-means clustering on multiple random projections of the fixed features of input data.
In contrast, the notion of consensus consistency here deals with learning representations that achieve a consensus regarding the cluster assignments of multiple clustering algorithms.
One example of a method that enforces consensus consistency is LA \citep{zhuang2019local}. LA builds on the ID task \citep{wu2018unsupervised} and was proposed as a method based on a robust clustering objective (using multiple runs of k-means) to move statistically similar data points closer in the representation space and dissimilar data points further away. However, \cite{zhuang2019local} did not evaluate the method with clustering metrics and only focused on linear evaluation using the learned features. Subsequently, we conducted a study to evaluate the clustering performance of these features (see Appendix) and observed that LA performed poorly when evaluated for clustering accuracy.
In Definition~\ref{def:consensus_consistency}, we inherently assume that the clustering algorithms under consideration have been tuned properly.
Unfortunately, the definition of consensus consistency is ill posed, and there can be arbitrarily many different partitions that can satisfy the given condition\footnote{a) Degenerate solution where all cluster assignments are the same. b) Random assignment can satisfy this condition given that all clustering process produces the same but random assignments}. We show that when exemplar consistency is used as an inductive bias, the resulting objective function achieves impressive performance on challenging datasets. Combining the exemplar and population constraints with consensus consistency seamlessly and effectively for clustering is the basis of our proposed method.

\subsection{Loss for Consensus and Population Consistency}
\label{sec:pop_cons}

We focus on learning generic representations that satisfy Definition \ref{def:consensus_consistency} for clustering. By using different clustering algorithms or different representation variations (such as projections), one can easily generate multiple different partitions of the same data. In unsupervised learning, it is not known which partitioning is correct. To tackle this problem, some additional assumptions are needed.

We assume that there is an underlying latent space $\mathcal{Z}^*$ (possibly not unique) such that all clusterings (based on latent space, algorithm or initialization variations) that take input data from this latent space produce similar data partitions. Furthermore, every clustering algorithm that also takes the true number of clusters as input produces the partition that is closest to the hypothetical ground truth. Moreover, we assume that there exists a function $h:X \rightarrow \mathcal{Z}^*$, where X represents the input space and $\mathcal{Z}^*$ represents the underlying latent space.
We call this assumption the \textbf{principle of consensus}.
The open question is how one constructs an efficient loss that reflects the \textbf{principle of consensus}. We define one such way below.

Given an input batch of images $\mathcal{X}_b\subset \mathcal{X}$, the goal is to partition these images into $K$ clusters. We obtain $p$ views of these images (by different image augmentation approaches) and define a loss such that cluster assignment of any of the $p$ views matches the target estimated from any other view. Without loss of generality, we define a loss for $p = 2$ views. The two views $\mathcal{X}_b^1, \mathcal{X}_b^2$ are generated using two randomly chosen image augmentations.

We learn a representation space $\mathcal{Z}_0$ at the end of every training iteration and obtain $M$ variations of $\mathcal{Z}_0$ as $\{ \mathcal{Z}_1, \mathcal{Z}_2, ... , \mathcal{Z}_M\}$ (e.g., random projections). The goal is to build an efficient loss according to the \textbf{principle of consensus} among $\mathcal{Z}_0$ and its $M$ variations $\{ \mathcal{Z}_1, \mathcal{Z}_2, ... , \mathcal{Z}_M\}$ such that we learn the latent space $\mathcal{Z}^*$ at the end of training (i.e., the learned features lie in the latent space described above). For a given batch of images $\mathcal{X}_b$ and a representation space $\mathcal{Z}_m, \forall m \in [1,...,M]$, we denote the cluster assignment probability of image $i$ and cluster $j$ for view $1$ as $\mathbf{p}_{i,j}^{1}(\mathcal{Z}_m)$ and that for view 2 as $\mathbf{p}_{i,j}^{2}(\mathcal{Z}_m)$. We concisely use $\tilde{\mathbf{p}}^{(1,m)},\tilde{\mathbf{p}}^{(2,m)}$ when we talk about all the images and all the clusters. Here, we define a loss that incorporates ``population consistency" and ``consensus consistency". We assume that the target cluster assignment probabilities for the representation $\mathcal{Z}_0$ are given (as in DeepCluster \citep{caron2018deep}), and they are denoted as $\mathbf{q}_{i,j}^{1}$ for view 1 and $\mathbf{q}_{i,j}^{2}$ for view 2.

We define the loss for any representation space $\mathcal{Z}$ and batch of images $\mathcal{X}_b$ as
\begin{align}
\label{eqn:ConCURLloss}
    \begin{split}
        L_{\mathcal{Z}_m}^1 &= - \frac{1}{2B}\sum_{i=1}^{B}\sum_{j=1}^K \mathbf{q}^{2}_{ij} \log \mathbf{p}^{1}_{ij}(\mathcal{Z}_m) \\
        L_{\mathcal{Z}_m}^2  &= - \frac{1}{2B}\sum_{i=1}^{B}\sum_{j=1}^K \mathbf{q}^{1}_{ij} \log  \mathbf{p}^{2}_{ij}(\mathcal{Z}_m), \\
        L_{\mathcal{Z}} &= \sum_{m = 1}^M \Big(  L_{\mathcal{Z}_m}^1 + L_{\mathcal{Z}_m}^2 \Big).
    \end{split}
\end{align}

\begin{figure}
    \centering
    \includegraphics[width=\textwidth]{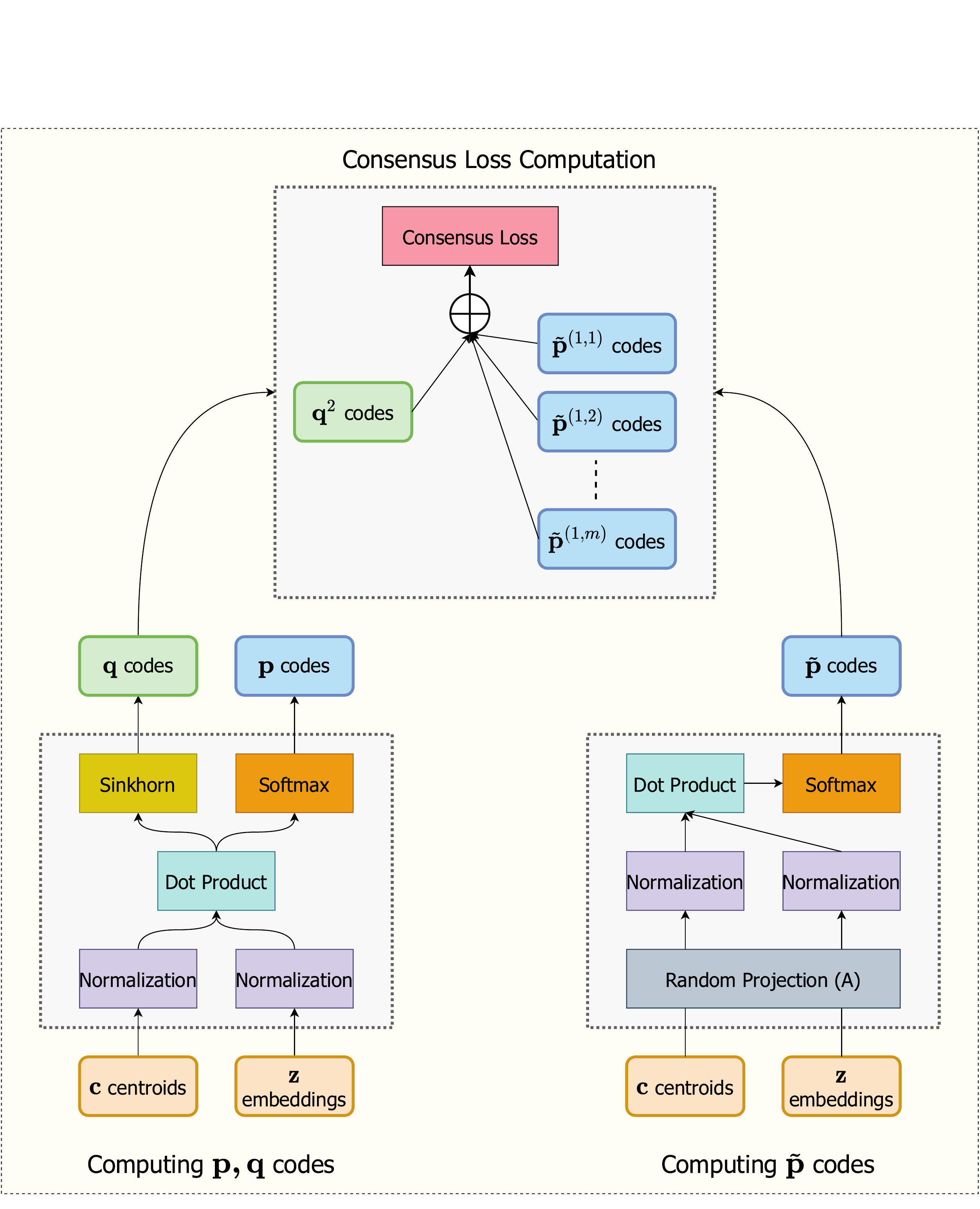}
    \caption{An illustration of the consensus loss part of \pap}
    \label{fig:concurl_flowchart}
\end{figure}

Note that here, consensus among the clustering results is defined via the number of common targets $\mathbf{q}$. An overview of the procedure is shown in Figure~\ref{fig:concurl_flowchart}. The exact details regarding how to obtain variations of $\mathcal{Z}_0$ and calculate the cluster
assignment probabilities $\mathbf{p}$ and targets $\mathbf{q}$ are described in the next section.

\subsection{End-to-End Stochastic Gradient Descent (SGD)-Based Trainable Consensus Loss}

In this section, we propose an end-to-end trainable algorithm and define a way to compute $\mathbf{p}$ and $\mathbf{q}$. When the cluster assignment probabilities $\mathbf{p}$ can take any values in the set $[0,1]$, we refer to the process as soft clustering, and when $\mathbf{p}$ is restricted to the set $\{0,1\}$, we refer to the process as hard clustering. 

Without loss of generality, in this paper, we focus on soft clustering, which makes it easier to define a loss function using the probabilities and update the parameters using the gradients to enable end-to-end learning. We follow the soft clustering framework presented in SwAV~\citep{caron2020unsupervised}, which is a centroid-based technique that aims to maintain consistency between the clusterings of the augmented views $\mathcal{X}_b^{1}$ and $\mathcal{X}_b^{2}$. We store a set of randomly initialized prototypes $C_0=\{ \mathbf{c}_0^1,\cdots,\mathbf{c}_0^K \} \in \mathbb{R}^{d\times K}$, where $K$ is the number of clusters and $d$ is the dimensionality of the prototypes. These prototypes are used to represent clusters and define a ``consensus consistency" loss. We compute $M$ variations of $C_0$ as $C_1,...,C_M$ exactly as we compute the $M$ variations of $\mathcal{Z}_0$.

\subsubsection{Cluster assignment probability $\mathbf{p}$}
We use a two-layer multilayer perceptron (MLP) $g$ to project the features $\mathbf{f}^1 = f_\theta(\mathcal{X}_b^1)$ and $\mathbf{f}^2 = f_\theta(\mathcal{X}_b^2)$ to a lower-dimensional space $\mathcal{Z}_0$ (of size $d$). The outputs of this MLP (referred to as cluster embeddings) are denoted as ${Z}_0^1 = \{\mathbf{z}_0^{1,1}, \ldots, \mathbf{z}_0^{1,B} \}$ and ${Z}_0^2  = \{\mathbf{z}_0^{2,1}, \ldots, \mathbf{z}_0^{2,B} \}$ for view $1$ and view $2$, respectively. Note that $h: \mathcal{X} \rightarrow \mathcal{Z}$ defined in \ref{sec:pop_cons} is equivalent to the composite function of $f: \mathcal{X} \rightarrow \Phi$ and $g: \Phi \rightarrow \mathcal{Z}$.

For a latent space $\mathcal{Z}$, we compute the probability of assigning a cluster $j$ to image $i$ using the normalized vectors $\bar{\mathbf{z}}^{1,i} = \frac{\mathbf{z}^{1,i}}{\|\mathbf{z}^{1,i}\|}$, $\bar{\mathbf{z}}^{2,i} = \frac{\mathbf{z}^{2,i}}{\|\mathbf{z}^{2,i}\|}$ and $\bar{\mathbf{c}}_j = \frac{\mathbf{c^j}}{\|\mathbf{c^j}\|}$ as

\begin{align}
\label{eqn:softmaxprob}
\begin{split}
        \mathbf{p}^1_{i,j}(\mathcal{Z},C)  &= \frac{\exp (\frac{1}{\tau} \langle \mathbf{\bar{z}}_i^1 , \mathbf{\bar{c}}_j \rangle)}{\sum_{j^{'}} \exp (\frac{1}{\tau} \langle \mathbf{\bar{z}}^1_{i} , \mathbf{\bar{c}}_{j^{'}} \rangle)} ,   \quad \\
        \mathbf{p}^2_{i,j}(\mathcal{Z},C) &= \frac{\exp (\frac{1}{\tau} \langle \mathbf{\bar{z}}_i^2 , \mathbf{\bar{c}}_j \rangle)}{\sum_{j^{'}} \exp (\frac{1}{\tau} \langle \mathbf{\bar{z}}^2_{i} , \mathbf{\bar{c}}_{j^{'}} \rangle)}.
\end{split}
\end{align}

We concisely write $ \mathbf{p}^1_{i}(\mathcal{Z}) = \{ \mathbf{p}^1_{i,j}(\mathcal{Z},C) \}_{j = 1}^K $ and $ \mathbf{p}^2_{i} = \{ \mathbf{p}^2_{i,j}(\mathcal{Z},C) \}_{j = 1}^K $. 
Here, $\tau$ is a temperature parameter, and we set its value to $0.1$, similar to \cite{caron2020unsupervised}. Note that we use $\mathbf{p}_{i}$ to denote the predicted cluster assignment probabilities for image $i$ (when not referring to a particular view), and the shorthand notation $\mathbf{p}$ is used when $i$ is clear from context.

\subsubsection{Targets $\mathbf{q}$}
The idea of predicting the assignments $\mathbf{p}$ and then comparing them with the high-confidence estimates $\mathbf{q}$ (referred to as codes henceforth) of the predictions was proposed by \cite{xie2016unsupervised}. While \cite{xie2016unsupervised} used pretrained features (from autoencoders) to compute the predicted assignments and the codes, the use of their approach in an end-to-end unsupervised manner might lead to degenerate solutions. \cite{asano2019self} avoided such degenerate solutions by enforcing an equipartition constraint (the prototypes equally partitioned the data) during code computation using the Sinkhorn-Knopp algorithm \citep{cuturi2013sinkhorn}. \cite{caron2020unsupervised} followed a similar formulation but computed the codes for the two views separately in an online manner for each minibatch. The assignment codes are computed by solving the following optimization problem:

\begin{align}
    \label{eqn:sinkhorn}
\begin{split}
        Q^1 &= \argmax_{Q\in \mathcal{Q}} \text{Tr}(Q^TC_0^TZ_0^1) + \epsilon H(Q) \\
    Q^2 &= \argmax_{Q\in \mathcal{Q}} \text{Tr}(Q^TC_0^TZ_0^2) + \epsilon H(Q),
\end{split}
\end{align}

where $ Q = \{\mathbf{q}_1, \ldots, \mathbf{q}_B \} \in \mathbb{R}_{+}^{K\times B}$, $\mathcal{Q}$ is the transportation polytope defined by
\begin{equation*}
\mathcal{Q} =  \{\mathbf{Q}\in \mathbb{R}^{K\times B}_{+}~\text{s.t}~  \mathbf{Q}\mathbf{1}_B = \frac{1}{K}\mathbf{1}_K, \mathbf{Q}^T\mathbf{1}_K = \frac{1}{B}\mathbf{1}_B \}
\end{equation*}

$\mathbf{1}_K$ is a vector of ones of dimension $K$ and $ H(Q) = -\sum_{i,j}Q_{i,j}\log Q_{i,j}
$. The above optimization is computed using a fast version of the Sinkhorn-Knopp algorithm \citep{cuturi2013sinkhorn}, as described by \cite{caron2020unsupervised}.

After computing the codes $Q^1 $ and $Q^2$, to maintain the consistency between the clustering results of the augmented views, the loss is computed using the probabilities $\mathbf{p}_{ij}$ and the assigned codes $\mathbf{q}_{ij}$ by comparing the probabilities of view $1$ with the assigned codes of view $2$ and vice versa, as in \eqref{eqn:ConCURLloss}.

\subsubsection{Defining variations of $Z_0$ and $C_0$}
To compute $\{Z_1,...,Z_M \}$, we project the $d$-dimensional space $Z_0$ to a $D$-dimensional space using a random projection matrix. We follow the same procedure to compute $\{C_1,...,C_M \}$ from $C_0$. At the beginning of the algorithm, we randomly initialize $M$ such transformations and fix them throughout training. Suppose that by using a particular random transformation (a randomly generated matrix $A$), we obtain $\Tilde{\mathbf{z}} = A\mathbf{z},\; \Tilde{\mathbf{c}} = A\mathbf{c}$. We then compute the softmax probabilities using the normalized vectors $\Tilde{\mathbf{z}}/\|\Tilde{\mathbf{z}}\|$ and $\Tilde{\mathbf{c}}/\|\Tilde{\mathbf{c}}\|$. This step is repeated with the $M$ transformation results in the $M$ predicted cluster assignment probabilities for each view. When the network is untrained, the embeddings $\mathbf{z}$ are random, and applying the random transformations, followed by computing the predicted cluster assignments, leads to a diverse set of soft cluster assignments. The parameter weights are trained by using the stochastic gradients of the loss for updates.

\subsubsection{Backbone loss}
To better capture exemplar consistency, based on previous evidence of successful clustering with the ID approach \citep{taoclustering}, we use ID \citep{wu2018unsupervised} as one of the losses, as in \cite{taoclustering}. The exemplar objective of ID is to classify each image as its own class.

Given $n$ images and a neural network $f_{\theta}$ for calculating features, we first normalize the features $\bar{f}_{\theta}(x) = \frac{f_{\theta}(x)}{\| f_{\theta}(x) \|}$. Then, ID defines the probability of an example $x$ being recognized as the $i$-th example as
\begin{equation}
\label{eqn:p_softmax}
	P(i \vert f_{\theta}(x)) =
	\frac{\exp \left( \langle \bar{f}_{\theta}(x_i), \bar{f}_\theta(x) \rangle / \tau \right) }
	{\sum_{j=1}^n \exp \left( \langle \bar{f}_{\theta}(x_j), \bar{f}_\theta(x) \rangle / \tau \right)  }.
\end{equation}

ID then uses the uniform distribution as a noise distribution $P_n = \frac{1}{n}$ to compute the probability that data example $x$ comes from a data distribution $P_d$ as opposed to the noise distribution $P_n$ as $h(i, f_{\theta}(x)) := \frac{P(i\vert f_{\theta}(x))}{P(i\vert f_{\theta}(x)) + m P_n(i)}$. Assuming that the noise samples are $m$ times more frequent than actual data samples, the ID loss is defined as

\begin{align}
\label{eqn:backbone_loss}
    \begin{split}
        L_{b} &= - E_{P_d} \left[\log h(i, x)\right]  - m  E_{P_n} \left[\log (1 - h(i, x')) \right],
    \end{split}
\end{align}
where $x'$ is the feature from a randomly drawn image other than image $x$ in a given dataset. We exactly follow the framework developed in \cite{wu2018unsupervised} to implement the ID loss.

The final loss that we seek to minimize is the combination of the losses $L_{\mathcal{Z}}$ (\eqref{eqn:ConCURLloss}) and $L_b$ (\eqref{eqn:backbone_loss}),
\begin{equation}
    \label{eqn:totalloss}
    L_{\text{total}} = \alpha L_{\mathcal{Z}} + \beta L_b.
\end{equation}

where $\alpha, \beta$ are nonnegative constants. Details of the algorithm are given Algorithm~\ref{alg_1} and we also provide  a PyTorch-style psuedocode in Algorithm \ref{alg:concurl} in the Appendix.

\begin{algorithm}[h]
\scriptsize
\caption{\scriptsize{{Consensus Clustering algorithm (\pap)}}}
\label{alg_1}
\begin{algorithmic}[1]
    \State Dataset $\mathcal{X}=\{x_i\}_{i=1}^N$, number clusters $K$, batch size $B$, weights $(\alpha,\beta)$, number of random transformations $M$, projection dimension $d$
    \State Randomly initialize network the networks $f_{\theta},g$; $K$ prototypes  ($\mathbf{c}_{1:K}$), and $M$ random projection matrices to dimension $d$ ($R_{1:M}$)  and epoch $e = 0$\;
\While{$e < $ total epoch number}

    \For{$b \in \{1, 2, \dots, \lfloor\frac{N}{B}\rfloor\}$}
            \State Select $B$ samples as $\mathcal{X}_b$ from $\mathcal{X}$ \;
            \State Make a forward pass on the two views: $f_\theta(\mathcal{X}_b^1)$ and $ f_\theta(\mathcal{X}_b^2)$
            \State $\mathbf{z}^{1}_{1:B}\longleftarrow g(f_\theta(\mathcal{X}_b^1)), \;\mathbf{z}^{2}_{1:B}\longleftarrow g(f_\theta(\mathcal{X}_b^2))$ \;
            \State Compute loss $L_b$ using (\ref{eqn:p_softmax},~\ref{eqn:backbone_loss})
            \State Compute $\mathbf{p}^1_{i,j},\mathbf{p}^2_{i,j}$ for both views using (\ref{eqn:softmaxprob})\;
            \State Compute codes of the current batch $\mathbf{q}$ using (\ref{eqn:sinkhorn})\;
            \For{$m \in \{1,2,\dots,M$ \}}
                \State $\Tilde{\mathbf{z}}\longleftarrow R_m \mathbf{z},\;\Tilde{\mathbf{c}} \longleftarrow R_m\mathbf{c}$
                \State Compute ($\Tilde{\mathbf{p}}^{(1,m)}_{i,j},\Tilde{\mathbf{p}}^{(2,m)}_{i,j} $) for normalized $\Tilde{\mathbf{z}}, \Tilde{\mathbf{c}}$ using (\ref{eqn:softmaxprob})\;
            \EndFor
            \State Compute loss $L_{\mathcal{Z}}$ using  (\ref{eqn:ConCURLloss}) \;
            \State Compute total loss (\ref{eqn:totalloss}). Update the network $f_{\theta},g$ parameters, and prototypes using gradients
    \EndFor
    \State $e := e + 1$
    \EndWhile
    \State Make forward pass on all the data and store the features\;
    \State Compute k-means clustering of the features

\end{algorithmic}
\end{algorithm}

\subsubsection{Computing the cluster metrics}
\label{sec:compute_clustering}
In this section, we describe the approach used to compute the cluster assignments and the metrics chosen to evaluate their quality. Note that we assume that the number of true clusters ($K$) in the data is known.

There are two ways to compute the cluster assignments. The first way is to use the embeddings generated by the backbone; here, the embeddings are the outputs of the ID block $f_{\theta}(x)$. The embeddings of all the images are computed, and then we perform k-means clustering.

The second method is to use the soft clustering block to compute the cluster assignments. It is sufficient to use the computed probability assignments $\{\mathbf{p}_i\}_{i=1}^N$ or the computed codes $\{\mathbf{q}_i\}_{i=1}^N$ and assign the cluster index as $c_i = \arg \max_{k} \mathbf{q}_{ik}$ for the $i^{\text{th}}$ data point. Once the model is trained, in this second approach, cluster assignment can be performed online without requiring the computation of the embeddings of all the input data.

We evaluate the quality of the clusterings using metrics such as the cluster accuracy, normalized mutual information (NMI), and adjusted Rand index (ARI). To compute the clustering accuracy, we are required to solve an assignment problem (computed using a Hungarian match~ \citep{kuhn1955hungarian, kuhn1956variants}) 
between the true class labels and the cluster assignments. In our analysis, we observe that using k-means with the embeddings produced by the ID block achieves better clustering accuracy, and we use this method throughout the paper while evaluating our proposed algorithm.

\subsection{Generating Multiple Clustering Results}
\label{sec:generate_ensemble}

\begin{table}[h]
\begin{center}
\caption{Different ways to generate ensembles}\label{tab:diffwaysensembles}
{%
\begin{tabular}{p{4.45cm}p{6.7cm}}\\\toprule  
Data Representation & Clustering algorithms \\\midrule
Different data preprocessing techniques & Multiple clustering algorithms (k-means, GMMs, etc)\\  \midrule
Subsets of features & Same algorithm with different parameters or initializations\\  \midrule
Different transformations of the features & Combination of multiple clustering results and different parameters or initializations \\  \bottomrule
\end{tabular} 
}
\end{center}
\end{table}

\cite{fred2005combining} discussed different ways to generate cluster ensembles; these methods are tabulated in Table \ref{tab:diffwaysensembles}. In our proposed algorithm, we focus on choosing of the appropriate data representation to generate cluster ensembles.

By fixing a stable clustering algorithm, we can generate arbitrarily large ensembles by applying different transformations on the embeddings. Random projections were previously successfully used in consensus clustering \citep{fern2003random}. By generating ensembles using random projections, we have control over the amount of diversity we can induce into the framework by varying the dimensionality of the random projection. In addition to random projections, we also use diagonal transformations \citep{hsu2018unsupervised} where different components of the representation vector are scaled differently. \cite{hsu2018unsupervised} illustrated that such scaling enables a diverse set of clusterings, which is helpful for the meta learning task. We study ablations over the number of transformations needed and the dimensions of these transformations in Section \ref{sec:ablation}.

\section{Understanding the Consensus Objective}
We investigate a potential hypothesis regarding ``training driven by noisy cluster assignments" that can shed light on the success of \pap\footnote{ A theoretically grounded explanation of \pap\ is considered future work due to the nonconvexity of deep learning methods and the nonconvexity of the proposed loss.}. The hypothesis stems from the following intuition. \textit{Using different clustering algorithms, the generated cluster assignments are noisy versions of the hypothetical ground truth; as the training process progresses, the noise in the cluster assignments is reduced, and eventually all the different clustering algorithms considered generate similar cluster assignments}.

\begin{figure}[h]
\centering
\subfigure[]{
    \label{fig:synthetic_data_original}
    \includegraphics[width=0.4\columnwidth]{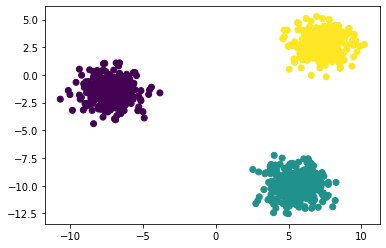}
    } 
\subfigure[]{
    \label{fig:synthetic_data_normalized}
    \includegraphics[width=0.4\columnwidth]{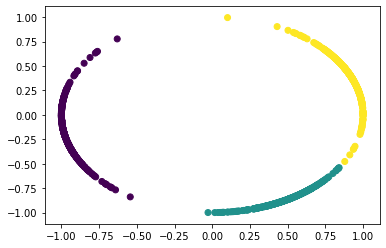}
}
\caption{a: Three-cluster dataset = z, b: Normalized data}
\end{figure}

\begin{table*}[htb]
\caption{Predicted cluster assignment probabilities and target probabilities obtained from the Sinkhorn algorithm for four data points}
\label{tab:synth_data}
\begin{center}
\resizebox{\textwidth}{!}{
\begin{tabular}{ ccc }   
Cluster assignment probability $\mathbf{p}$ & Cluster assignment probability $\tilde{\mathbf{p}}$ & Targets $\mathbf{q}$ from Sinkhorn \\  
\begin{tabular}{ |c| ccc| } 
    \hline
    & Cluster 1 &  Cluster 2 & Cluster 3  \\
    \hline
    1  &
        0.6473 & 0.2587 & 0.0940  \\
    \hline
    2  &
         0.7180 & 0.1812 & 0.1008 \\
    \hline
    3 &
        0.0832 & 0.3160 &  0.6008\\
    \hline
    4 &
        {0.1543} & {0.5917} & {0.2541}\\      
\hline
\end{tabular} &  
\begin{tabular}{ |c| ccc| }  
    \hline
    & Cluster 1 &  Cluster 2 & Cluster 3  \\
    \hline
    1 &
        0.6764 & 0.2271 & 0.0965  \\
    \hline
   2   &
         0.7305 & 0.1680 & 0.1015 \\
    \hline
    3 &
        0.0802 & 0.3371 &  0.5827\\
    \hline
    4 &
        0.1357 & 0.5784 & 0.2860\\ 
\hline
\end{tabular} &  
\begin{tabular}{ |c| ccc| } 
    \hline
    & cluster 1 &  cluster 2 & cluster 3  \\
    \hline
   1  &
        1.0 & 8.9e-09 & 1.7e-17  \\
    \hline
    2  &
         1.0 & 9.1e-13 & 8.7e-18 \\
    \hline
    3 &
        6.8e-18 & 2.2e-6 &  1.0\\
    \hline
  4 &
        2.5 e-12 & 1.0 & 5.4e-8\\ 
\hline
\end{tabular} \\
\end{tabular}
}
\end{center}
\end{table*}

We verify this hypothesis empirically with the help of the following experiments on the \stl\ dataset: (i) we observe the noisy clusterings generated by using random projections and (ii) verify that the noise in the cluster assignments is reduced as training progresses.

For the purpose of demonstrating noisy cluster assignments, we use synthetic data as follows.
We generate three clusters in $\mathbb{R}^2$, as shown in Figure~\ref{fig:synthetic_data_original}, and compute the centroids of each cluster. Here, the centroids act as the prototypes. We then generate a Gaussian random projection matrix $A$ with a dimensionality of $\mathbb{R}^{2\times2}$. We first normalize the embeddings (2-dimensional features) and the centroids (see Figure~\ref{fig:synthetic_data_normalized}).
Using the matrix $A$, we transform both the embeddings and prototypes for the new space and normalize the resultant vectors. 

We follow the soft clustering framework discussed earlier and compute the soft cluster assignments for the original and transformed data. We observe that the cluster assignment probabilities in the new space are noisy versions of the cluster assignment probabilities in the original space (see Table~\ref{tab:synth_data}). 

\begin{figure}[htb!]
\centering
\label{fig:NMI_analysis}
\subfigure[]{
    \label{fig:NMI_mean}
    \includegraphics[width=0.45\columnwidth]{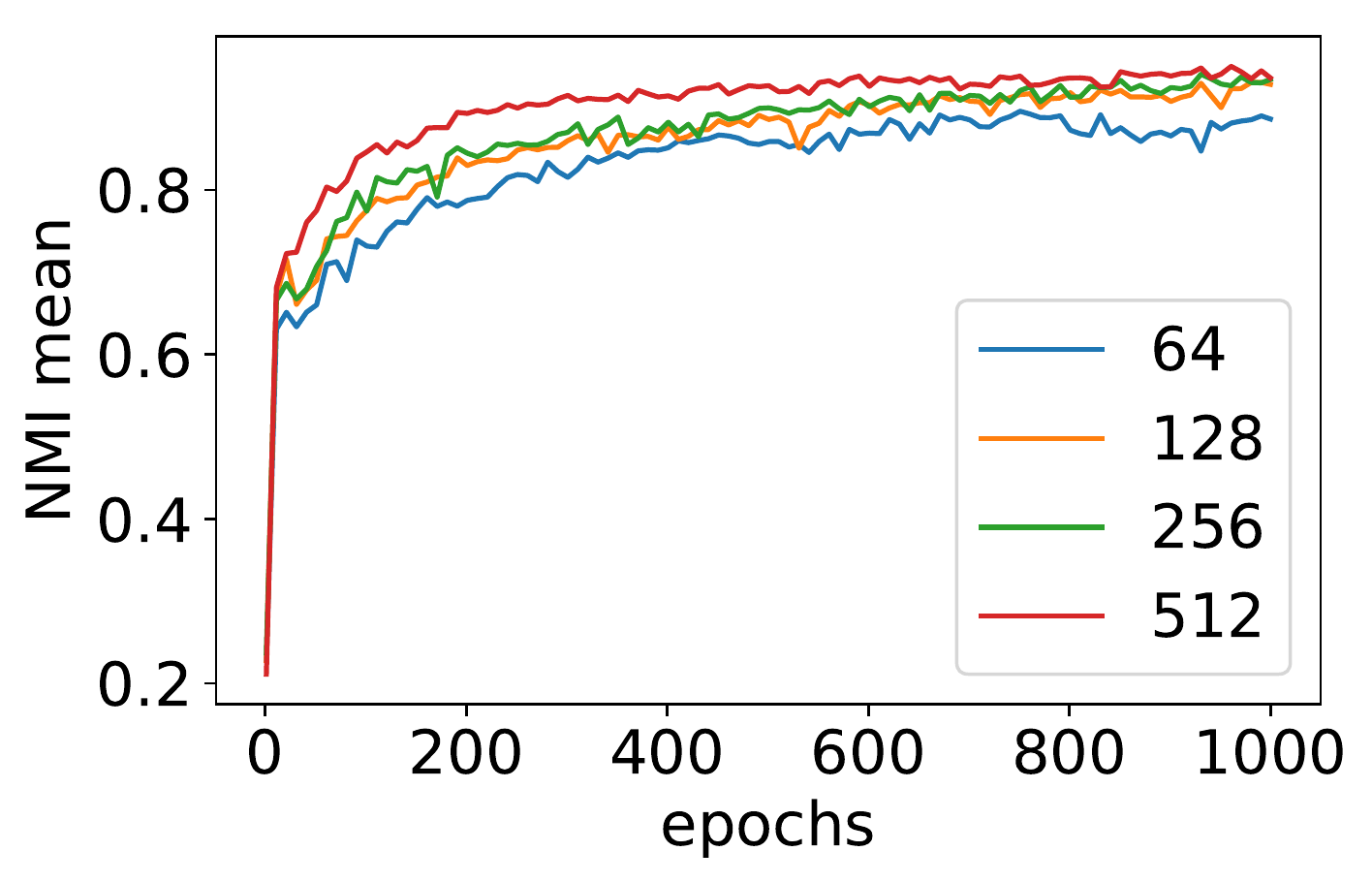}
    } 
\subfigure[]{
    \label{fig:NMI_std}
    \includegraphics[width=0.45\columnwidth]{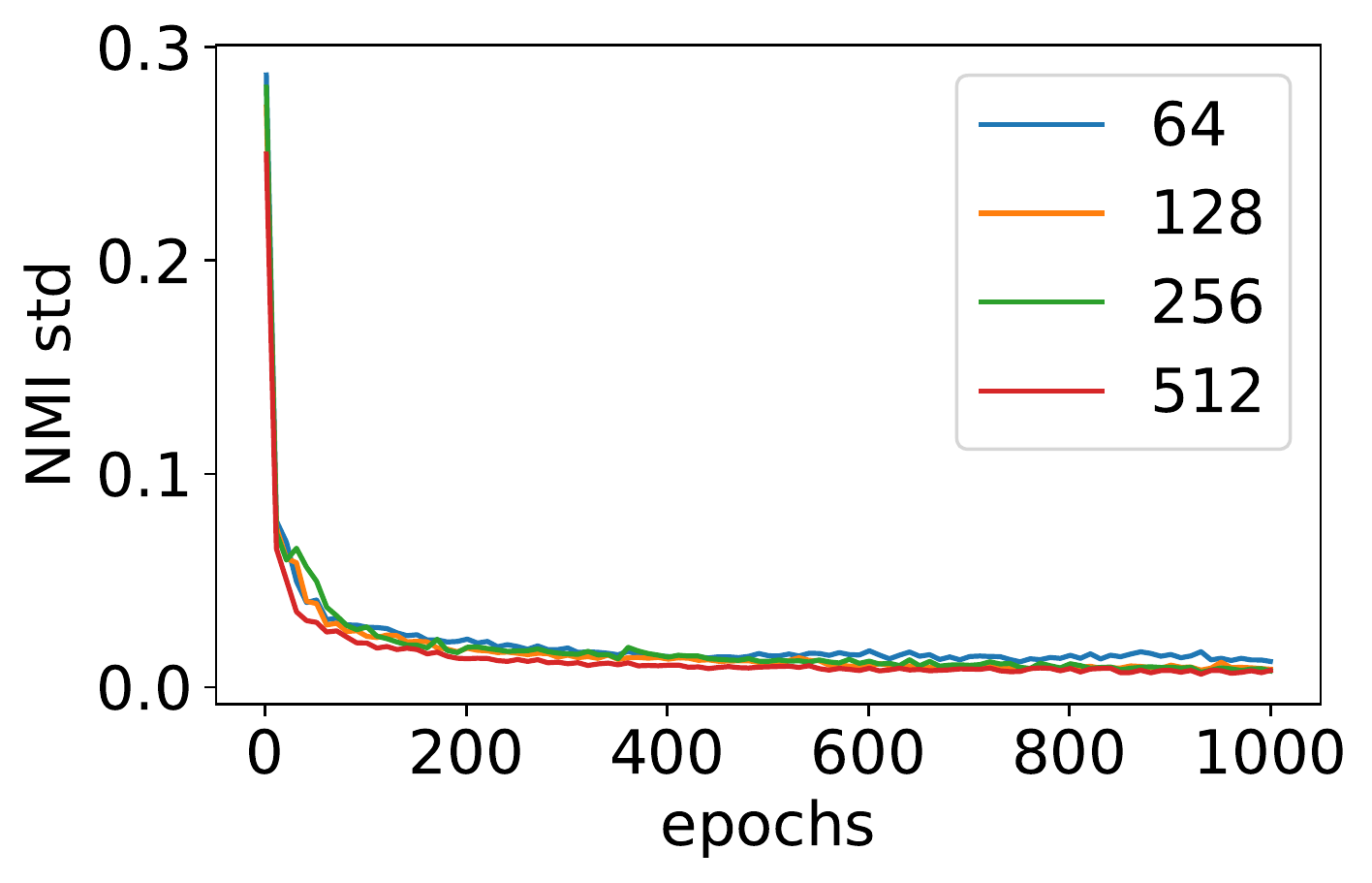}
}
\caption{Pairwise NMI values as a way to measure the diversity in the ensemble; the results are obtained for the \stl\ dataset, and the pairwise NMI values for different random projection dimensions are shown (the original dimensionality of $\mathbf{z}$ is 256)}
\end{figure}

To verify that the noise in the cluster assignment probabilities is reduced as training progresses, we perform the following experiment. We measure the similarity among the cluster assignments at every epoch to observe the effect of consensus as training progresses. For each random projection used, we use the cluster assignment probabilities $\tilde{\mathbf{p}}$ and compute cluster assignments by taking an $\argmax$ on $\tilde{\mathbf{p}}$ for each image. We obtain M such cluster assignments due to the M random projections. We then compute a pairwise NMI (similar to the analysis of \cite{fern2003random}) between every two cluster assignments and compute the average and standard deviation of the pairwise NMI values across the $\frac{M(M-1)}{2}$ pairs. An NMI score of $1.0$ signifies that the two clusters perfectly correlate with each other, and a score of $0.0$ implies that the two clusters are uncorrelated. We observe from Figure 2 that the pairwise NMI increases as training progresses and becomes closer to 1. At the beginning of training, the cluster assignments are very diverse (small NMI scores with a large standard deviation), and as training progresses, the diversity is reduced (large NMI scores with a smaller standard deviation). This observation leads us to conclude that for the applied clustering algorithms (defined using different random projections), we have learned an embedding space where the different cluster assignments concur. In other words, ``consensus consistency" is achieved. Additionally, it is evident from our empirical results in Section~\ref{sec:empirical_evaluation} that we achieve an improved overall clustering accuracy. 

If noisy cluster assignments are the reason behind the improved performance, one might wonder if it is sufficient to simply add noise to the original cluster assignments rather than computing multiple cluster assignments. However, this may not be fruitful because if noise is added externally, one must define a scheduler to reduce the noise as training progresses. However, in the case of \pap, the end-to-end learning algorithm determines the rate of consensus or agreement between $\mathbf{p} $ and $\tilde{\mathbf{p}}$ itself. In the next section, we provide empirical evidence of the effectiveness of our method.

\section{Empirical Evaluation}
\label{sec:empirical_evaluation}
Evaluating clustering algorithms is a notoriously hard problem. The reference text \cite{jain1988algorithms} states the following:
\small
\textit{The validation of clustering structures is the most difficult and frustrating part of cluster analysis. Without a strong effort in this direction, cluster analysis will remain a black art accessible only to those true believers who have experience and great courage.}
\normalsize

In the literature on representation learning for clustering, e.g., \cite{cons_clus, Huang_2020_CVPR, taoclustering}, to evaluate the performance of different algorithms, the following methodology has been used: as a set of models with some hyperparameters are trained, these models are sorted by the observed clustering performance. Finally the best model's results are reported. This methodology is called \textbf{max performance} in the remainder of the paper. We assess the quality of the learned embeddings by using five challenging image datasets for clustering and report their performance with the max-performance strategy. Although the max-performance procedure provides some insights into the performance of the method under consideration, we provide additional insights by significantly extending the evaluations. In practice, it is desirable that the learned models can be utilized for different datasets other than the training dataset. However, the max-performance
method may not be suitable for this purpose. To address
this, we design two additional experiments that focus on the performance of cross-model features under a distribution shift. Furthermore, we also assess the quality of the learned embeddings in image retrieval tasks. Finally, we present a detailed ablation study to assess the impact of the loss terms, data augmentation methods, hyperparameters and architecture choices utilized to obtain a more complete picture.

\subsection{Image Clustering with the Max-Performance Strategy}
We evaluate our algorithm and compare it with existing methods on some popular image datasets, namely, \iten, \idogs, \stl, \cten, and \chundred.

For \chundred, we use the 20 meta classes as the class labels while evaluating the clustering results. For \stl, similar to the earlier PICA \citep{huang2020deep} and GATCluster \citep{niu2020gatcluster} approaches, we use both training and testing splits for training and evaluation. Note that PICA also uses an unlabeled data split with 100k points in \stl, which we do not use. \iten\ and \idogs\ are subsets of ImageNet, and we use only the training splits for these two datasets \citep{deng2009imagenet}. We use the same classes as \cite{Chang_2017_ICCV} for evaluation on the \iten\ and \idogs\ datasets. The dataset summary is given in Table \ref{tab:dataset_summary}. We evaluate the cluster accuracy, NMI, and ARI of each computed cluster assignment (see the Appendix for details).
\begin{table}[ht]
\begin{center}
\caption{{Dataset summary}}
  \label{tab:dataset_summary}
{%
\begin{tabular}{|c | c | c | c | c |} 
 \hline
 Dataset & Classes & Split & Samples & Resolution
 \\ [0.5ex] 
 \hline
 \iten
 & 10 & train & 13000 & 160 $\times$ 160 \\ 
 \hline
 \idogs 
 & 15 & train & 19500  & 160 $\times$ 160 \\
 \hline
 \stl
 & 10 & train+test & 13000 & 96$\times$ 96 \\
 \hline
 \cten 
 & 10 & train & 50000 & 32$\times$ 32 \\
 \hline
 \chundred 
 & 20 & train & 50000 & 32$\times$ 32 \\
 \hline
\end{tabular}}
\end{center}
\end{table}

\subsubsection{Comparison with state-of-the-art baselines}

\begin{table}[h]
\caption{Methods Compared }
\label{tab:DAType}
\centering
\begin{tabular}{|c | c |} 
\hline
Method    & Reference   \\ \hline
k-means       & \cite{kmeans1967}   \\ \hline
SC          &      \cite{Spectral2002}  \\ \hline
AC  & \cite{Pasi2006Fast}        \\ \hline
NMF    & \cite{NMF}      \\ \hline
AE          & \cite{Bengio2007Greedy}    \\ \hline
SDAE     & \cite{SDAE2010}       \\ \hline
DeCNN   & \cite{Zeiler2010Deconvolutional}       \\ \hline
JULE   & \cite{Yang2016Joint}       \\ \hline
DEC       & \cite{Xie2016}      \\ \hline
DAC   & \cite{DAIC2017}     \\ \hline
Deep Cluster     & \cite{vf2018}        \\ \hline
DDC    & \cite{chang2019deep}      \\ \hline
IIC      & \cite{IIC2019}    \\ \hline
DCCM   &  \cite{Wu_2019_ICCV}      \\ \hline
GATCluster      & \cite{gatcluster}     \\ \hline
PICA    & \cite{Huang_2020_CVPR}      \\ \hline
CC     &  \cite{cons_clus}     \\ \hline
ADC      & \cite{ADC2019}        \\ \hline
ID   & \cite{taoclustering}      \\ \hline
IDFD    & \cite{taoclustering}      \\ \hline
\end{tabular}
\end{table}

\begin{table*}[h!]
\scriptsize
\renewcommand\tabcolsep{3.1pt}
\caption{Clustering with the max-performance strategy}
\label{tab:comparison_sota}
\centering
\resizebox{\linewidth}{!}
{
\begin{tabular}{|c|ccc|ccc|ccc|ccc|ccc|}
\hline
\multirow{2}{*}{Method}             & \multicolumn{3}{c|}{\stl}&\multicolumn{3}{c|}{\iten}&\multicolumn{3}{c|}{\idogs}&\multicolumn{3}{c|}{\cten}&\multicolumn{3}{c|}{\chundred}\\
\cline{2-16} & ACC  & NMI  & ARI& ACC&NMI&ARI   & ACC&NMI&ARI  &ACC&NMI&ARI   &ACC&NMI&ARI \\
\hline\hline
k-means             & 0.192  & 0.125 & 0.061   & 0.241 &0.119& 0.057   & 0.105 &0.055&0.020   & 0.229 &0.087&0.049   & 0.130 &0.084&0.028   \\
SC                 & 0.159  & 0.098 & 0.048   & 0.274 &0.151&0.076    & 0.111 &0.038&0.013   & 0.247 &0.103&0.085   & 0.136 &0.090&0.022   \\
AC                  & 0.332  & 0.239 & 0.140   & 0.242 &0.138&0.067    & 0.139 &0.037&0.021   & 0.228 &0.105&0.065   & 0.138 &0.098&0.034   \\
NMF                           & 0.180  & 0.096 & 0.046   & 0.230 &0.132&0.065    & 0.118 &0.044&0.016   & 0.190 &0.081&0.034   & 0.118 &0.079&0.026   \\
AE             & 0.303  & 0.250 & 0.161   & 0.317 &0.210&0.152    & 0.185 &0.104&0.073   & 0.314 &0.239&0.169   & 0.165 &0.100&0.048   \\
SDAE                    & 0.302  & 0.224 & 0.152   & 0.304 &0.206&0.138    & 0.190 &0.104&0.078   & 0.297 &0.251&0.163   & 0.151 &0.111&0.046   \\
DeCNN    & 0.299  & 0.227 & 0.162   & 0.313 &0.186&0.142    & 0.175 &0.098&0.073   & 0.282 &0.240&0.174   & 0.133 &0.092&0.038   \\
JULE                & 0.277  & 0.182 & 0.164   & 0.300 &0.175&0.138    & 0.138 &0.054&0.028   & 0.272 &0.192&0.138   & 0.137 &0.103&0.033   \\
DEC                       & 0.359  & 0.276 & 0.186   & 0.381 &0.282&0.203    & 0.195 &0.122&0.079   & 0.301 &0.257&0.161   & 0.185 &0.136&0.050   \\
DAC                      & 0.470  & 0.366 & 0.257   & 0.527 &0.394&0.302    & 0.275 &0.219&0.111   & 0.522 &0.396&0.306   & 0.238 &0.185&0.088   \\
Deep Cluster                & 0.334  & N/A   &  N/A    & N/A &N/A&N/A          & N/A &N/A&N/A         & 0.374& N/A   &  N/A  & 0.189 & N/A &  N/A   \\
DDC                  & 0.489  & 0.371 & 0.267   & 0.577 &0.433&0.345    & N/A &N/A&N/A         & 0.524 &0.424&0.329   & N/A &N/A&N/A         \\
IIC                       & 0.610  & N/A   &  N/A    & N/A &N/A&N/A          & N/A &N/A&N/A         & 0.617& N/A & N/A     & 0.257& N/A & N/A     \\
DCCM               & 0.482  & 0.376 & 0.262   & 0.710 &0.608 &0.555   & 0.383&0.321 &0.182   & 0.623& 0.496&0.408   & 0.327 &0.285&0.173   \\
GATCluster            & 0.583  & 0.446 & 0.363   & 0.762 &0.609 &0.572   & 0.333&0.322 & 0.200  & 0.610&0.475 &0.402   & 0.281 &0.215&0.116   \\
PICA             & 0.713  & 0.611 & 0.531   & 0.870 &0.802 &0.761   & 0.352&0.352 & 0.201  & 0.696&0.591 &0.512   & 0.337 &0.310&0.171   \\
CC                              & \textbf{0.850}  & \textbf{0.746 }& \textbf{0.726}   & 0.893 &0.859 &0.822   & 0.429&0.445 & 0.274  & 0.790&0.705 &0.637   & 0.429 &0.431&0.266   \\
ADC                     & 0.530  &N/A&N/A          & N/A &N/A&N/A          & N/A &N/A&N/A         & 0.325 & N/A& N/A     &0.160  &N/A&N/A  \\
ID & 0.726&0.64&0.526&0.937&0.867&0.865&0.476&0.47&0.335&0.776&0.682&0.616&0.409&0.392&0.243\\

IDFD  & 0.756&0.643&0.575&0.954&0.898&0.901&0.591&0.546&0.413&0.815&0.711&0.663&0.425&0.426&0.264\\
\hline
\pap     & 0.749&0.636&0.566&\textbf{0.958}&\textbf{0.907}&\textbf{0.909}&\textbf{0.695}&\textbf{0.63}&\textbf{0.531}&\textbf{0.846}&\textbf{0.762}&\textbf{0.715}&\textbf{0.479}&\textbf{0.468}&\textbf{0.3034}\\
\hline
\end{tabular}
}
\end{table*}

In our comparison, we consider some state-of-the-art methods that were developed for image clustering problems and targeted for end-to-end training scenarios with random initialization. We should note that we do not consider baselines that use prior information, e.g., the nearest neighbors algorithm derived by using pretrained models. The implementation details of \pap\ are provided in the Appendix, and the results are presented in Table~\ref{tab:comparison_sota}.

We observe that \pap\ outperforms the baseline algorithms considered in terms of all three metrics for all the datasets except \stl. \pap\ improves the state-of-the art clustering accuracy by approximately $17.5\%$ on \idogs, by $12.7\%$ on \chundred\ and by $3.8\%$ on \cten. Although \pap\ improves upon the results of ID \citep{taoclustering}, please note that ID is the backbone used in this paper and is slightly worse than IDFD, as shown in \cite{taoclustering}.

The proposed method achieves good clustering performance on popular computer vision datasets. 
Similar to all the algorithms considered, we assume that $K$, the number of clusters, is known. However, this may not hold true in practice in real-world applications. In such a case, we may assume an estimate for the upper bound on the number of clusters to use as the number of prototypes. Additionally, we also assume that the dataset is equally distributed among the $K$ clusters. 
If this assumption (also common in the literature \citep{huang2020deep,niu2020gatcluster}) does not hold, the fast Sinkhorn-Knopp algorithm used to solve Eq. \ref{eqn:sinkhorn} may not be optimal.

\begin{figure}
    \centering
    \subfigure[\iten]{\includegraphics[width=0.43\textwidth]{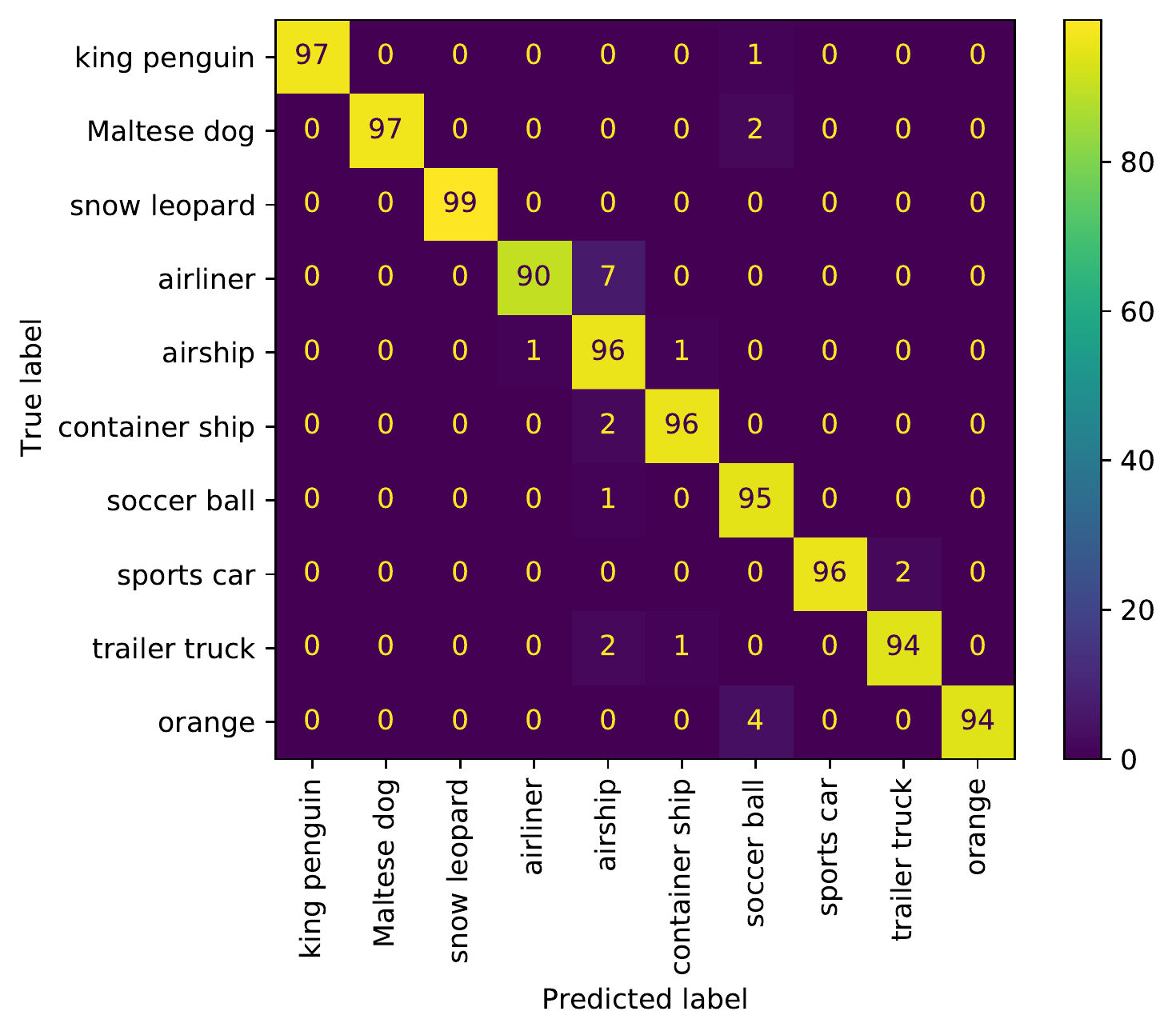}} 
    \subfigure[\idogs]{\includegraphics[width=0.43\textwidth]{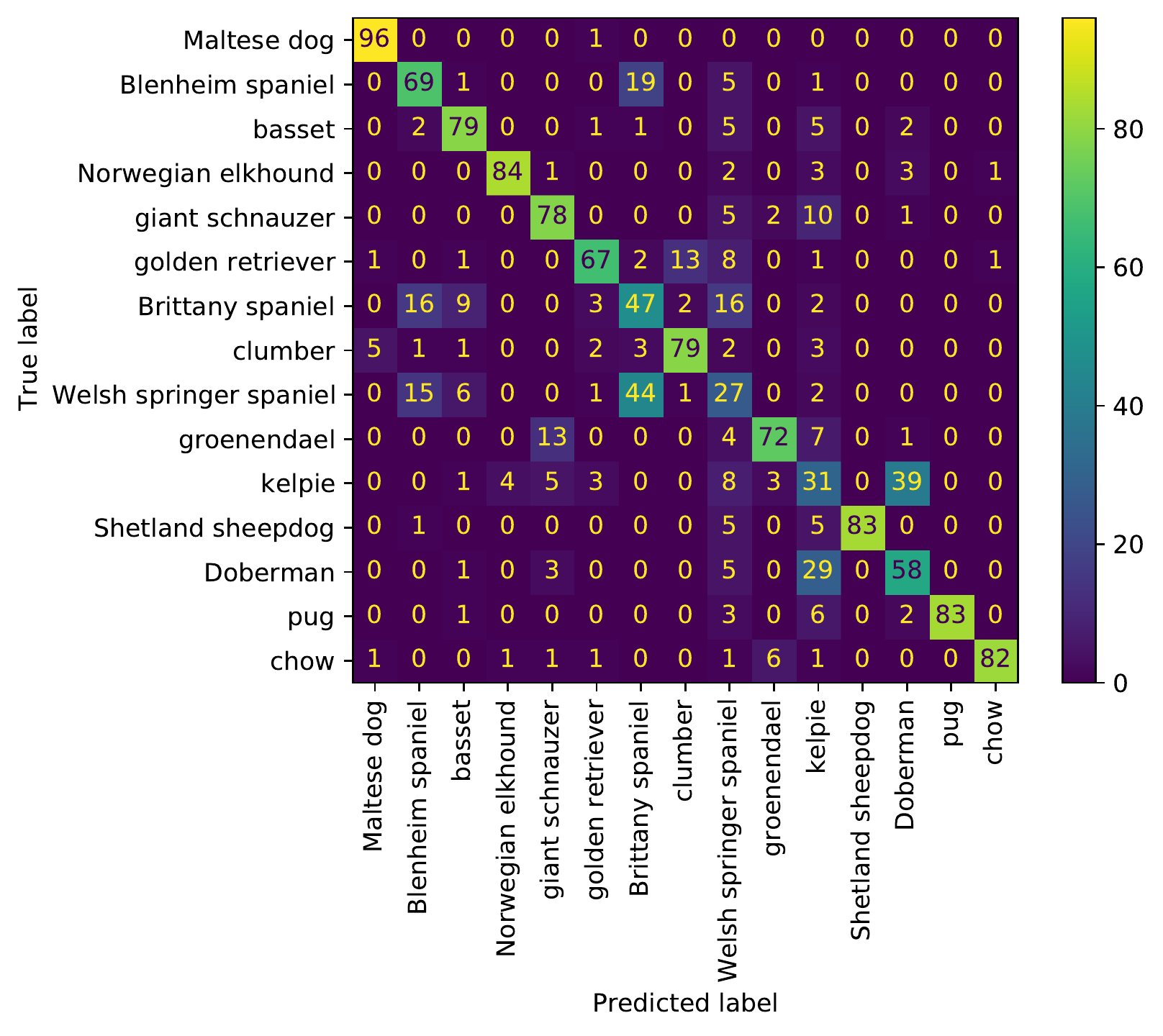}} 
    \subfigure[\cten]{\includegraphics[width=0.43\textwidth]{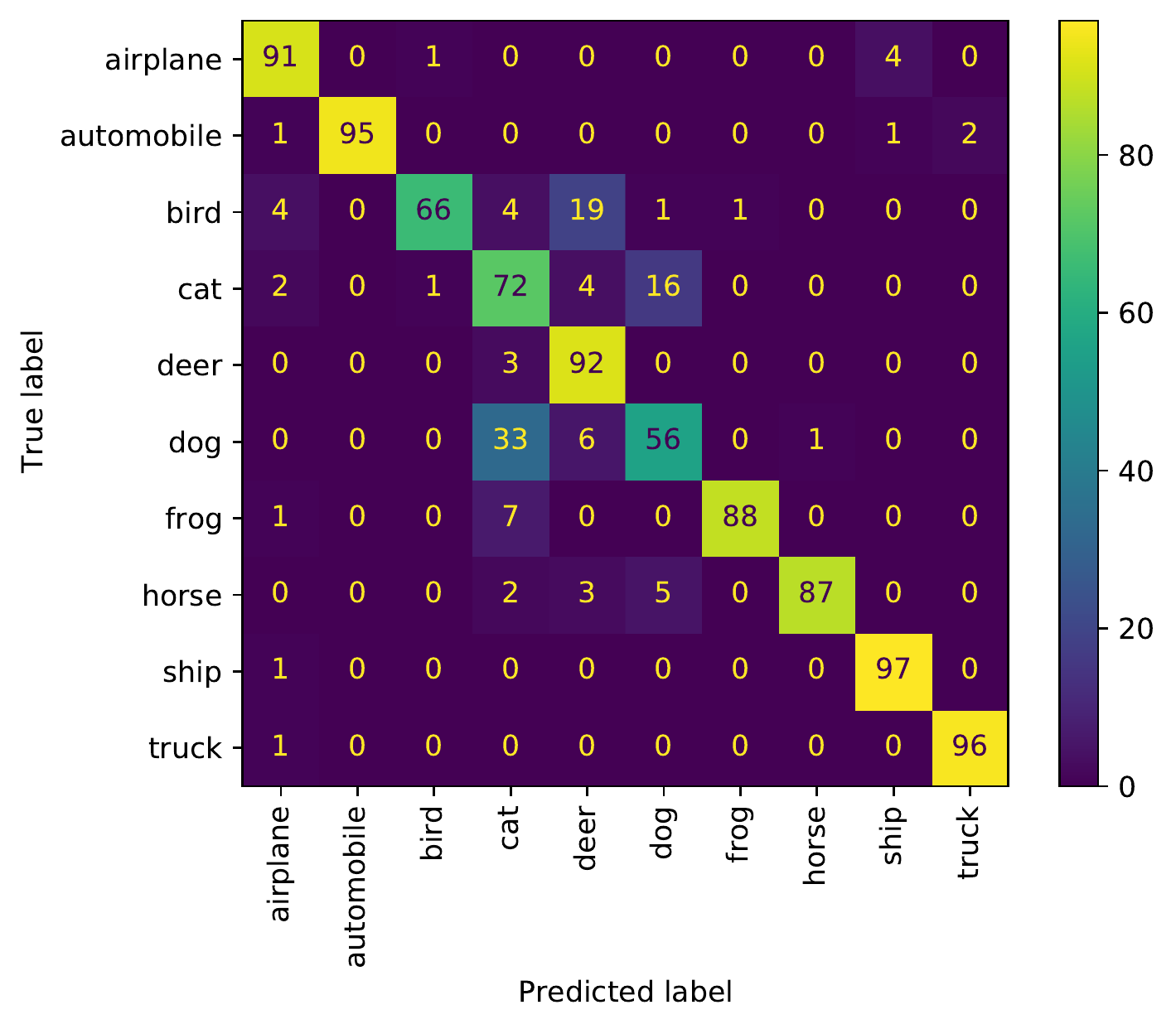}}
    \subfigure[\stl]{\includegraphics[width=0.43\textwidth]{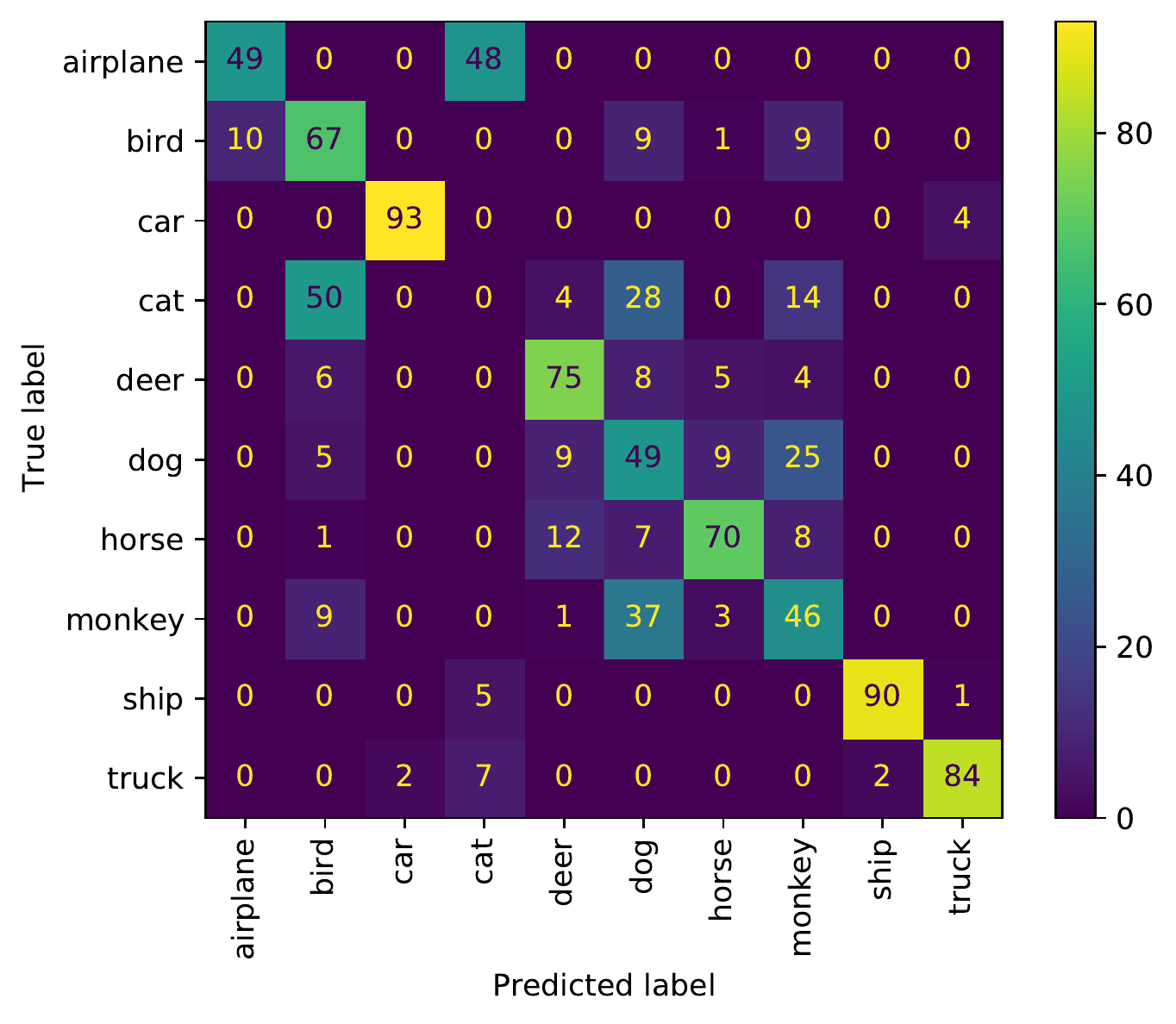}}
    \subfigure[\chundred]{\includegraphics[width=0.65\textwidth]{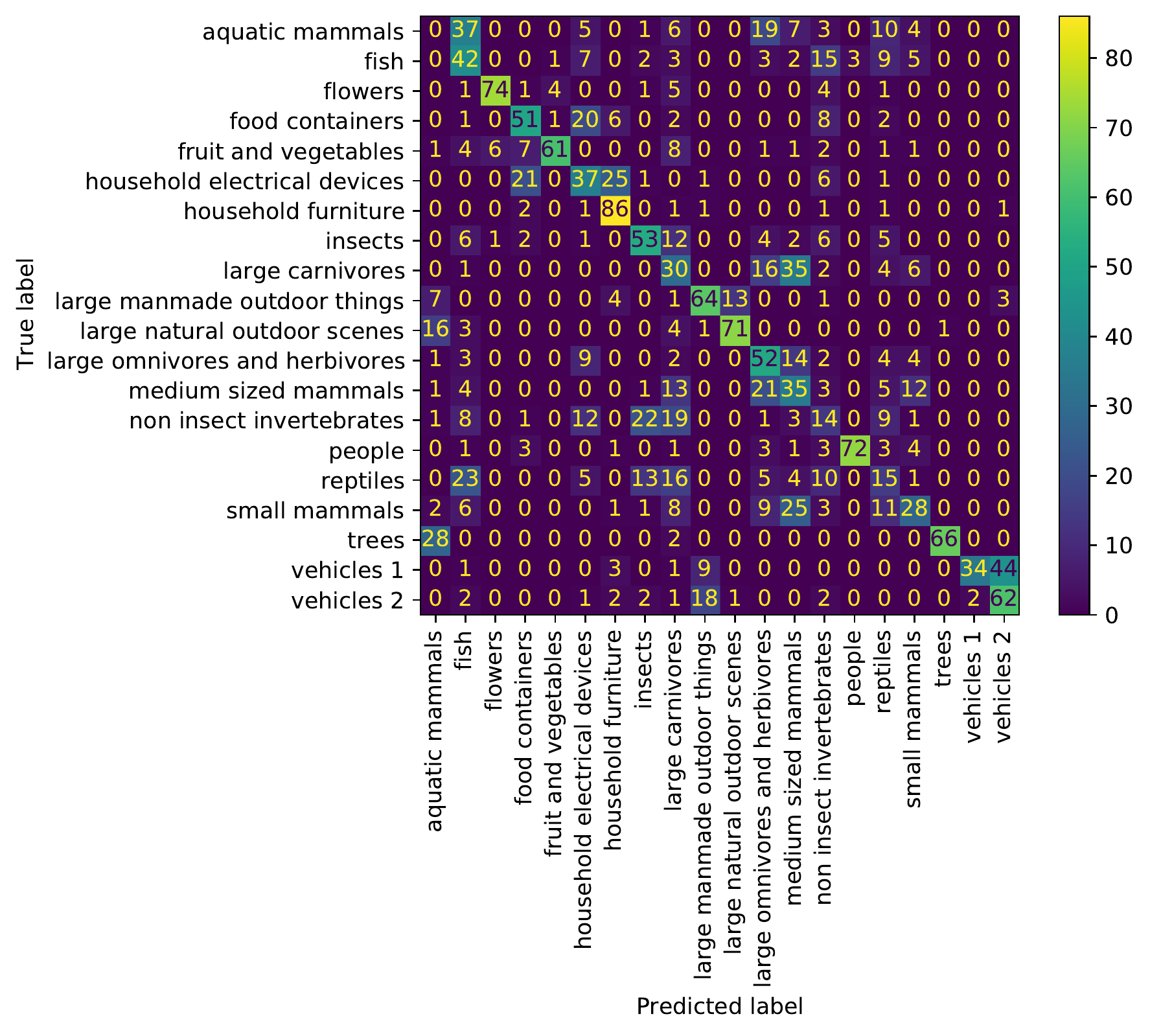}} 
    \caption{
Confusion matrices 
}
    \label{fig:confusion_matrix}
\end{figure}

\subsubsection{{Performance on the Test split}}
\label{sec:test_acc}

\begin{table}[htb!]

\renewcommand\tabcolsep{3.1pt}
\caption{{Clustering on the test dataset}}
\label{table:test_accs}
\centering
\begin{tabular}{|c|ccc|ccc|}
\hline

\multirow{2}{*}{Dataset}             & 
\multicolumn{3}{c|}{Train Dataset}&\multicolumn{3}{c|}{Test Dataset}\\
\cline{2-7} & ACC  & NMI  & ARI& ACC&NMI&ARI  \\
\hline\hline
        \cten & 0.846 & 0.762 & 0.715 & 0.838 & 0.744 & 0.700 \\
        \chundred & 0.479 & 0.468 & 0.303 & 0.427 & 0.430 & 0.254 \\
        \iten & 0.958 & 0.908 & 0.910 & 0.914 & 0.876 & 0.838 \\
        \idogs & 0.695 & 0.630 & 0.532 & 0.660 & 0.675 & 0.515 \\
\hline
\end{tabular}
\end{table}

{
In the previous section, we studied the clustering performance on the training split used to train the algorithm. Here, we shall evaluate the clustering performance on a held out test set. We shall use the standard test split for the datasets \cten,~\chundred,~\iten, and \idogs. We used the trained models to extract the features for each test dataset and compute the clustering as above. We observe from Table~\ref{table:test_accs} that the performance doesn't get affected much on the test set. This shows that the algorithm is able to extract good feature representations for clustering on data not used for training which shows good generalization ability of the algorithm when the data is drawn from the same distribution.
}

\subsubsection{Class-specific accuracy}
We present the class-specific accuracies (percentage) and confusion matrices in Figure~\ref{fig:confusion_matrix}. In each row $i$, $j^{th}$ entry in the matrix represents the percentage of samples from category $i$ belonging to the cluster of category $j$. For better visualization, we round each percentage to the nearest integer.{Row sum may not be equal to 100 because we are rounding to the nearest integer}. For perfect clustering, all elements along the diagonal should be equal to 100. Here, we note some interesting observations. For \iten, the airliner category shows the worst performance, with 7\% of airliner samples being confused with the airship category. Additionally, 4\% of the samples from the orange category are categorized as soccer balls. In \idogs, there are three types of spaniels- Blenheim spaniels, Brittany spaniels and Welsh springer spaniel, which look very similar to each other. Nineteen percent of the samples from the Blenheim spaniel category are categorized as Brittany spaniels, and 5\% are categorized as Welsh springer spaniels. Similarly, 16\% of the samples from the Brittany spaniel category are categorized as Blenheim spaniels, and 16\% more are categorized as Welsh springer spaniels. Forty-four percent of the samples from the Welsh springer spaniel category are categorized as Brittany spaniels, and 15\% are categorized as Blenheim spaniels. Kelpies and Dobermans are also confused with each other, where 39\% of the kelpie samples are categorized as Dobermans, and 29\% of the Doberman samples are categorized as kelpies. For \cten, 33\% of the samples from the dog category are categorized as cats, and 16\% of the samples from the cat category are categorized as dogs.

For \stl, none of the samples from the cat category are categorized correctly. Even though \stl\ and \cten\ have the same list of categories, \stl\ seems harder to cluster than \cten. Note that \stl\ has only 13000 images to learn representations, while in \cten, 60000 images are used. For \chundred, none of the samples from aquatic mammals are categorized correctly. Thirty-seven percent of the samples from aquatic mammals are categorized as fish, and 19\% are categorized as large omnivores and herbivores. In the case of reptiles, only 15\% of the examples are categorized correctly, but 23\% are categorized as fish, 16\% as large carnivores and 13\% as insects. Surprisingly, 28\% of the samples from trees are categorized as aquatic mammals.

\subsection{Out-of-Distribution Results}
In this section, we evaluate \pap~ by performing clustering on dataset that is not used during training. We focus mainly on studying the clustering performance on datasets that maybe similar to the training dataset and datasets that may have a different number of clusters than the training dataset.

\subsubsection{Cross-Model Accuracy}
Here, we calculate the clustering performance achieved when the model is trained on one dataset but evaluated on a different dataset that may have a different number of clusters. For example, the first row in Table~\ref{tab:cross_model_imagenet_10_dogs} gives the performance of the model trained on \iten\ and evaluated on both \iten\ and \idogs. Similarly, the second row shows the performance of the model trained on \idogs. { We find that performance on \iten\ is decreased to 35.6\% when the model trained on \idogs\ is used instead of the model trained on \iten. Similarly, the performance on \idogs\ is decreased to 17.7 \% when the model trained on \iten\ is used instead of the model trained on \idogs}. Table~\ref{tab:cross_model_cifar_10_10020} provides the same performance metrics for \cten\ and \chundred. 

{
For cross-model performance to be high, the embedding function must be generalizable to the out of distribution dataset. It is important to observe that for each pair of datasets considered, the distributions of the datasets are very different due to the classes being completely different in both cases (\iten~vs~\idogs, and \cten~vs~\chundred). However, since we are considering datasets with a small number of datapoints and small number of classes (see Table~\ref{tab:dataset_summary}), the representation power of the learnt embeddings is limited and this affects the cross-model accuracy. Moreover, the consensus loss $L_{\mathcal{Z}}$ here assumes knowledge of the number of clusters in the dataset. Therefore, the embeddings learnt by optimizing the $L_{total}$ loss on one dataset may be sub-optimal for evaluating clustering on a dataset with different number of clusters.
}

It is clear that these performance drops are significant, and the generalization performance of the learned embeddings needs to be assessed by taking out-of-distribution datasets into account. However, since there are only 2 groups of different datasets, it is difficult to reach a definitive conclusion. Hence, in the following section, we propose a new evaluation methodology that sheds light on the out-of-distribution performance of the learned embeddings.

\begin{table}[htb!]
\renewcommand\tabcolsep{3.1pt}
\caption{\iten\ vs. \idogs: cross-model performance}
\label{tab:cross_model_imagenet_10_dogs}
\centering
\begin{tabular}{|c|ccc|ccc|}
\hline

\multirow{2}{*}{Model Training Dataset}             & 
\multicolumn{3}{c|}{\iten}&\multicolumn{3}{c|}{\idogs}\\
\cline{2-7} & ACC  & NMI  & ARI& ACC&NMI&ARI  \\
\hline\hline
\iten             & 0.958  & 0.908 & 0.910   & 0.177  & 0.127 & 0.068    \\
\idogs                 & 0.356 & 0.298 & 0.184   & 0.695 & 0.630 & 0.532   \\
\hline
\end{tabular}
\end{table}

\begin{table}[htb!]
\renewcommand\tabcolsep{3.1pt}
\caption{\cten\ vs. \chundred: cross-model performance}
\label{tab:cross_model_cifar_10_10020}
\centering
\begin{tabular}{|c|ccc|ccc|}
\hline

\multirow{2}{*}{Model Training Dataset}             & \multicolumn{3}{c|}{\cten}&\multicolumn{3}{c|}{\chundred  }\\
\cline{2-7} & ACC  & NMI  & ARI& ACC&NMI&ARI  \\
\hline\hline
\cten            & 0.846 & 0.762  & 0.715 &  0.178 & 0.158 & 0.061      \\
\chundred               &  0.464 & 0.359  & 0.250 & 0.480 & 0.468 & 0.304   \\
\hline
\end{tabular}
\end{table}

\subsubsection{ImageNet random-10 and random-15 accuracies}
\begin{figure}[htb!]
    \centering
    \subfigure[]{\includegraphics[width=0.24\textwidth]{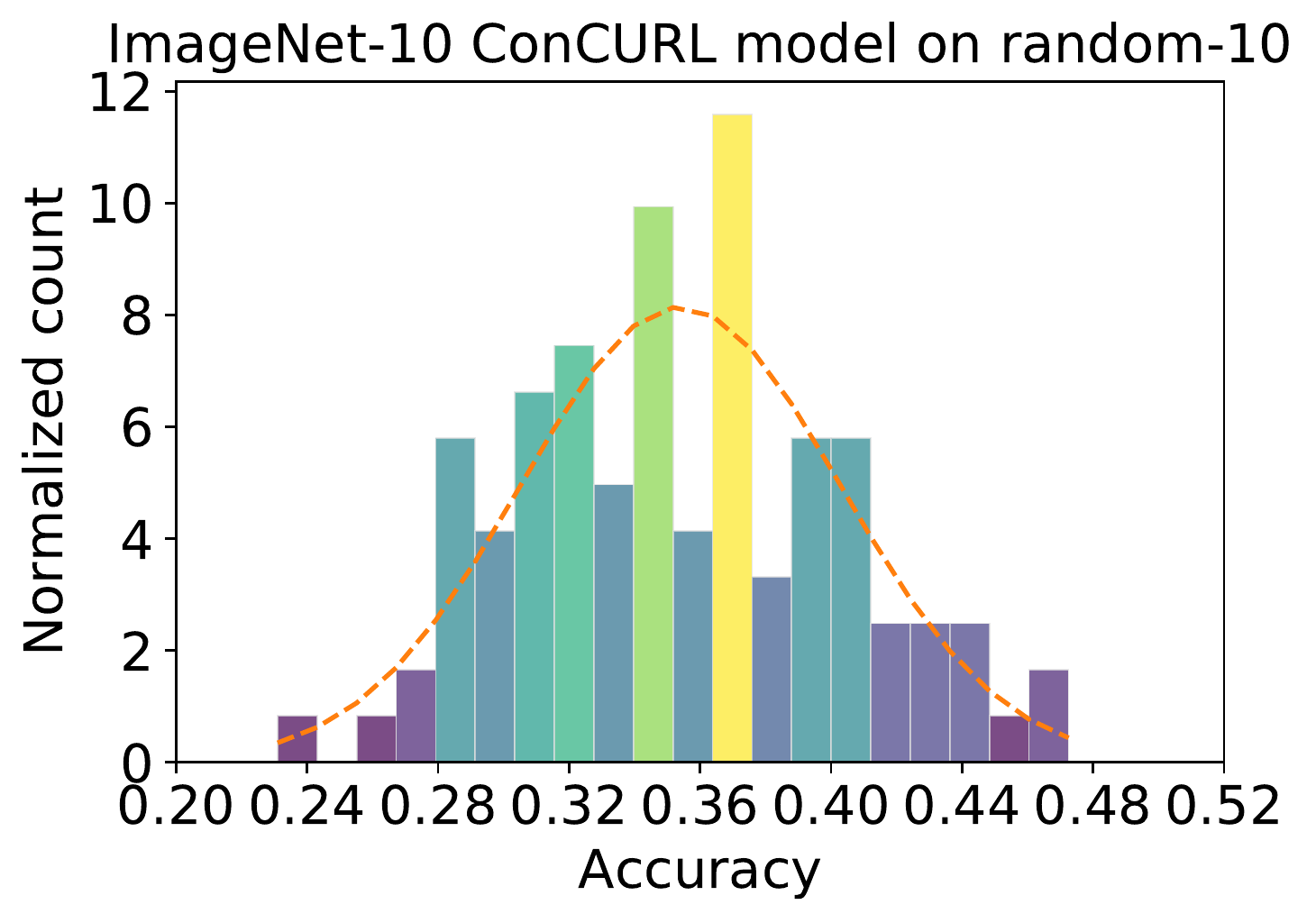}} 
    \subfigure[]{\includegraphics[width=0.24\textwidth]{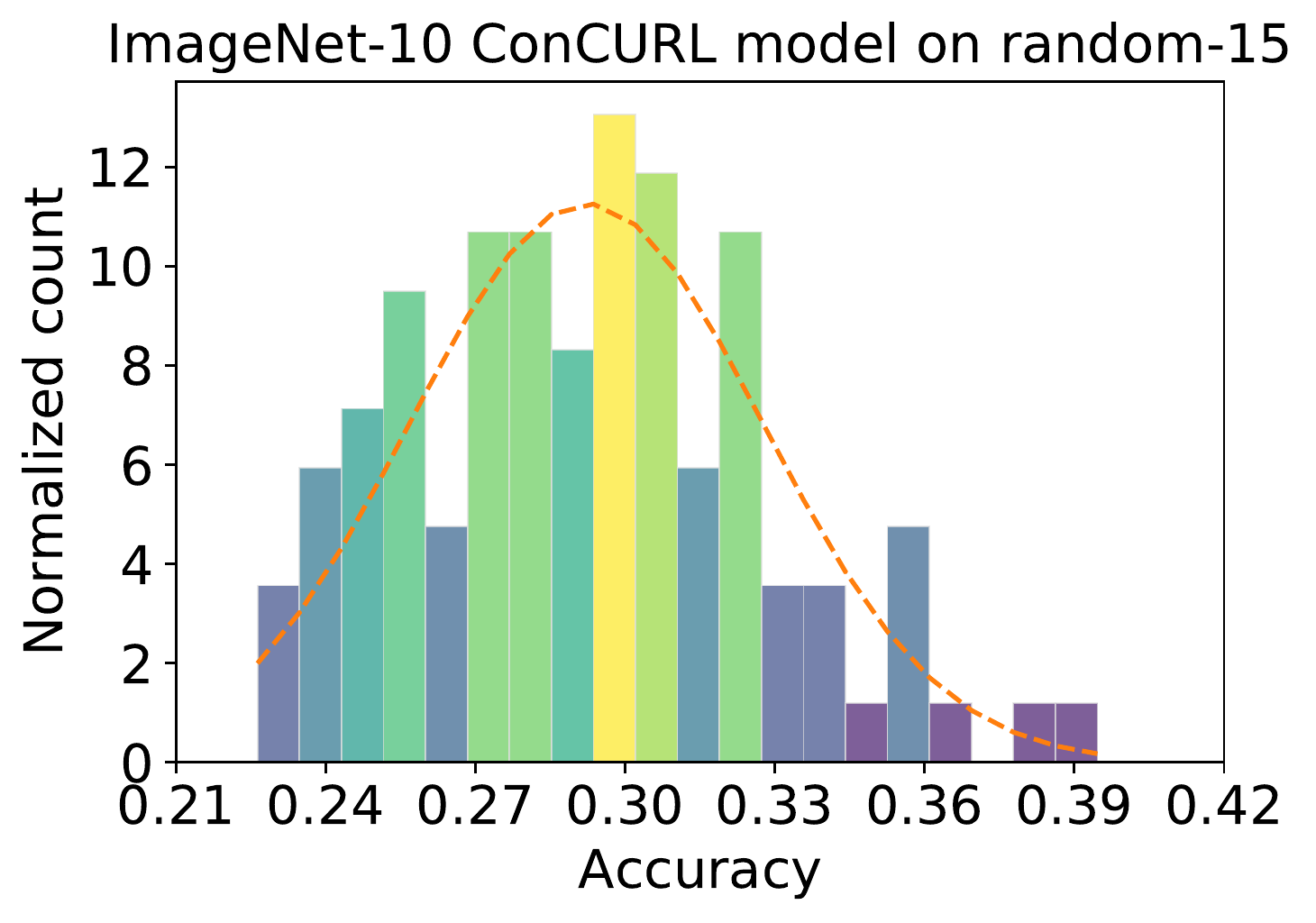}} 
    \subfigure[]{\includegraphics[width=0.24\textwidth]{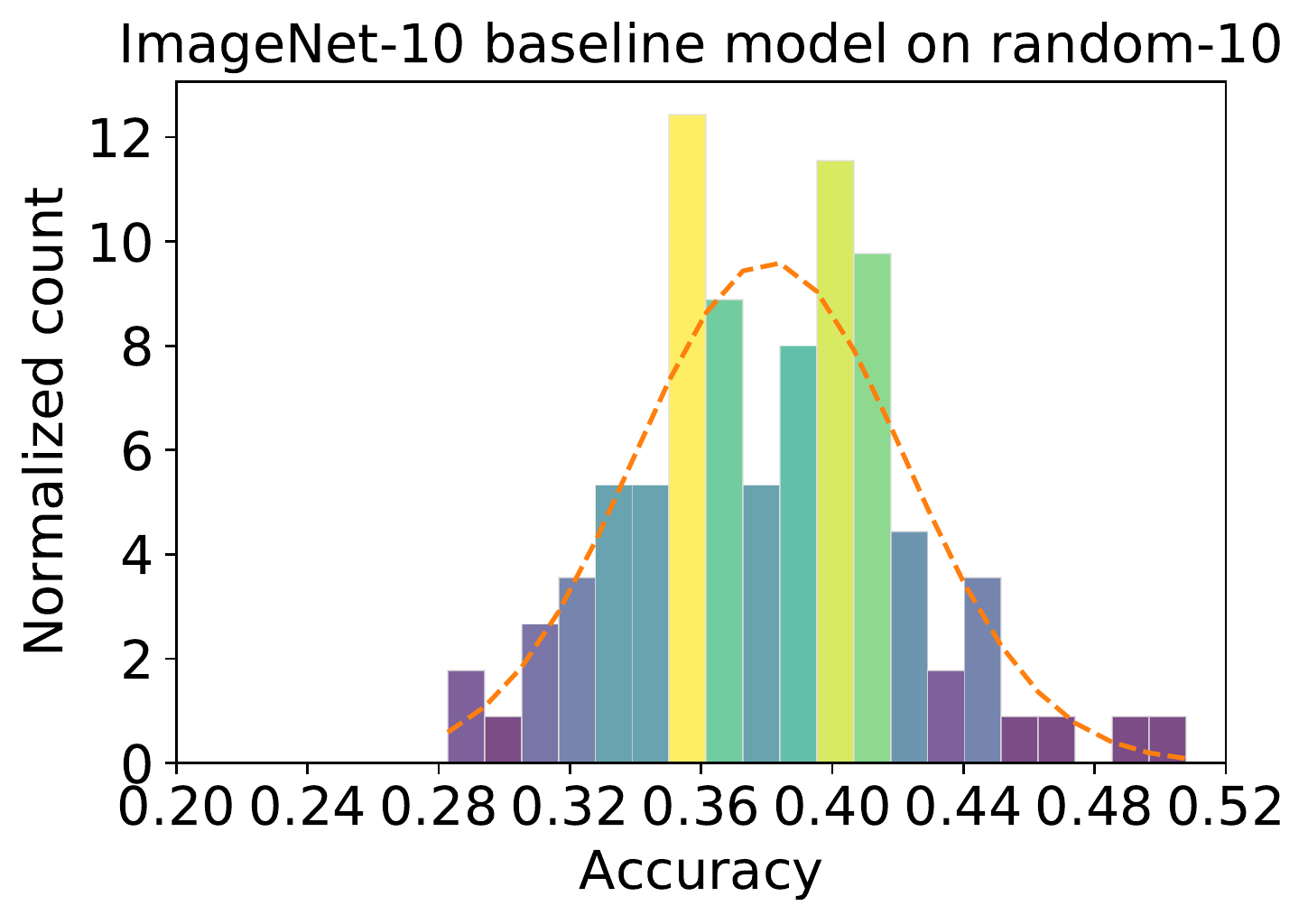}} 
    \subfigure[]{\includegraphics[width=0.24\textwidth]{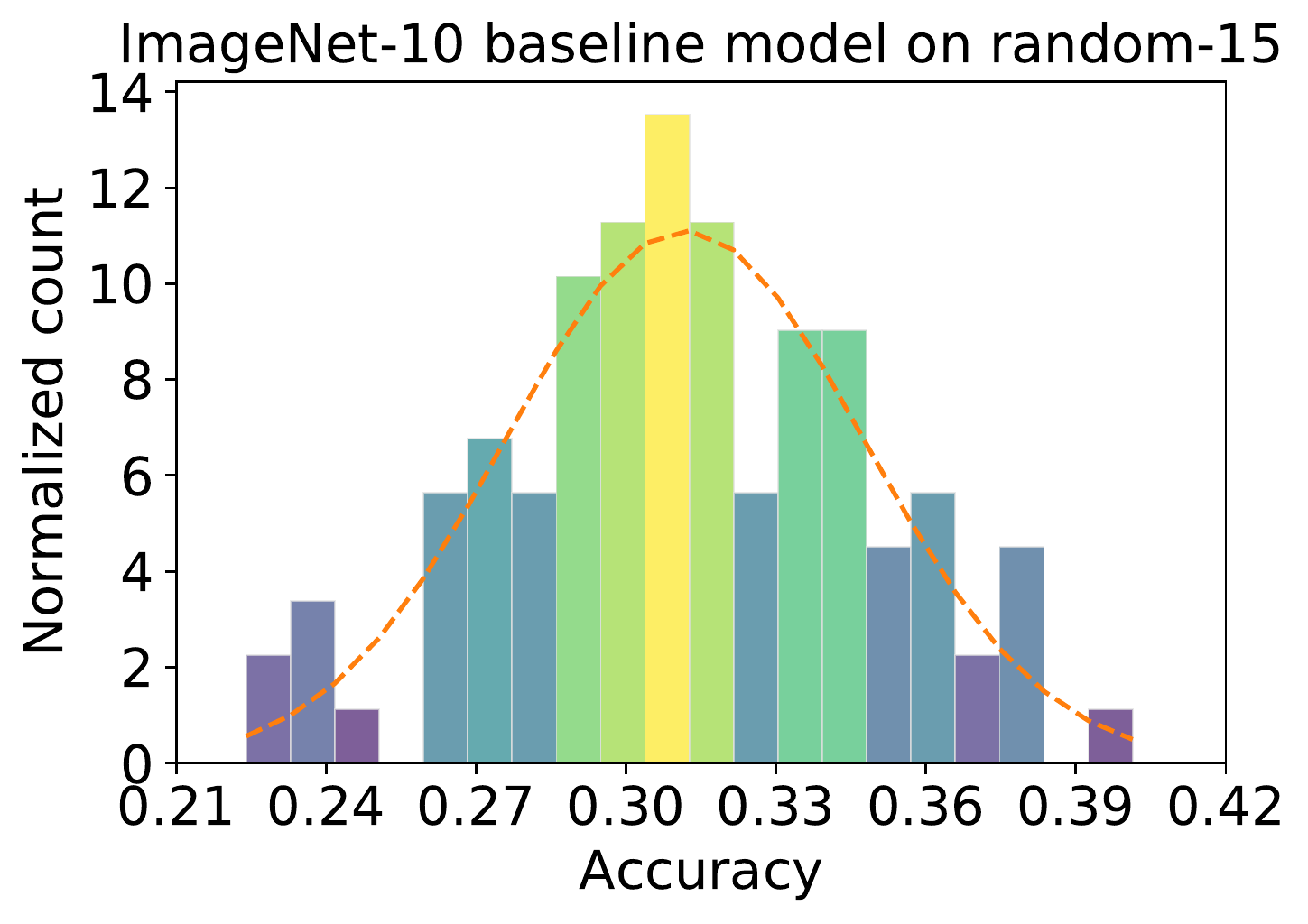}} 
    \caption{Histogram of clustering accuracies for models trained on \iten: (a) \pap\ model evaluated on random-10, (b) \pap\ model evaluated on random-15, (c) Baseline (ID) model evaluated on random-10, (d) Baseline (ID) model evaluated on random-15}
    \label{fig:ImageNet10ConcurlvsIDrandom}
\end{figure}

\begin{figure}[htb!]
    \centering
    \subfigure[]{\includegraphics[width=0.24\textwidth]{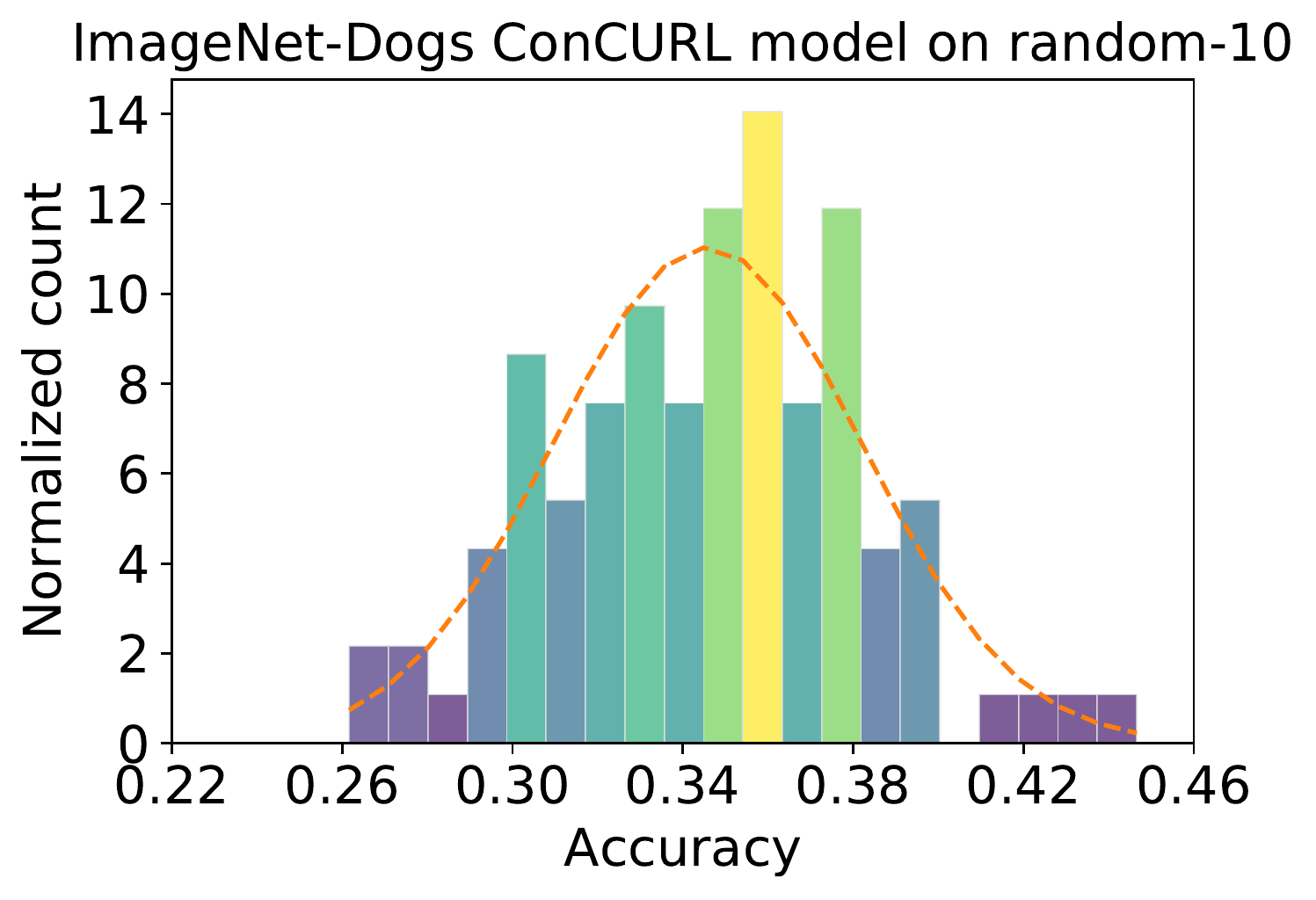}}
    \subfigure[]{\includegraphics[width=0.24\textwidth]{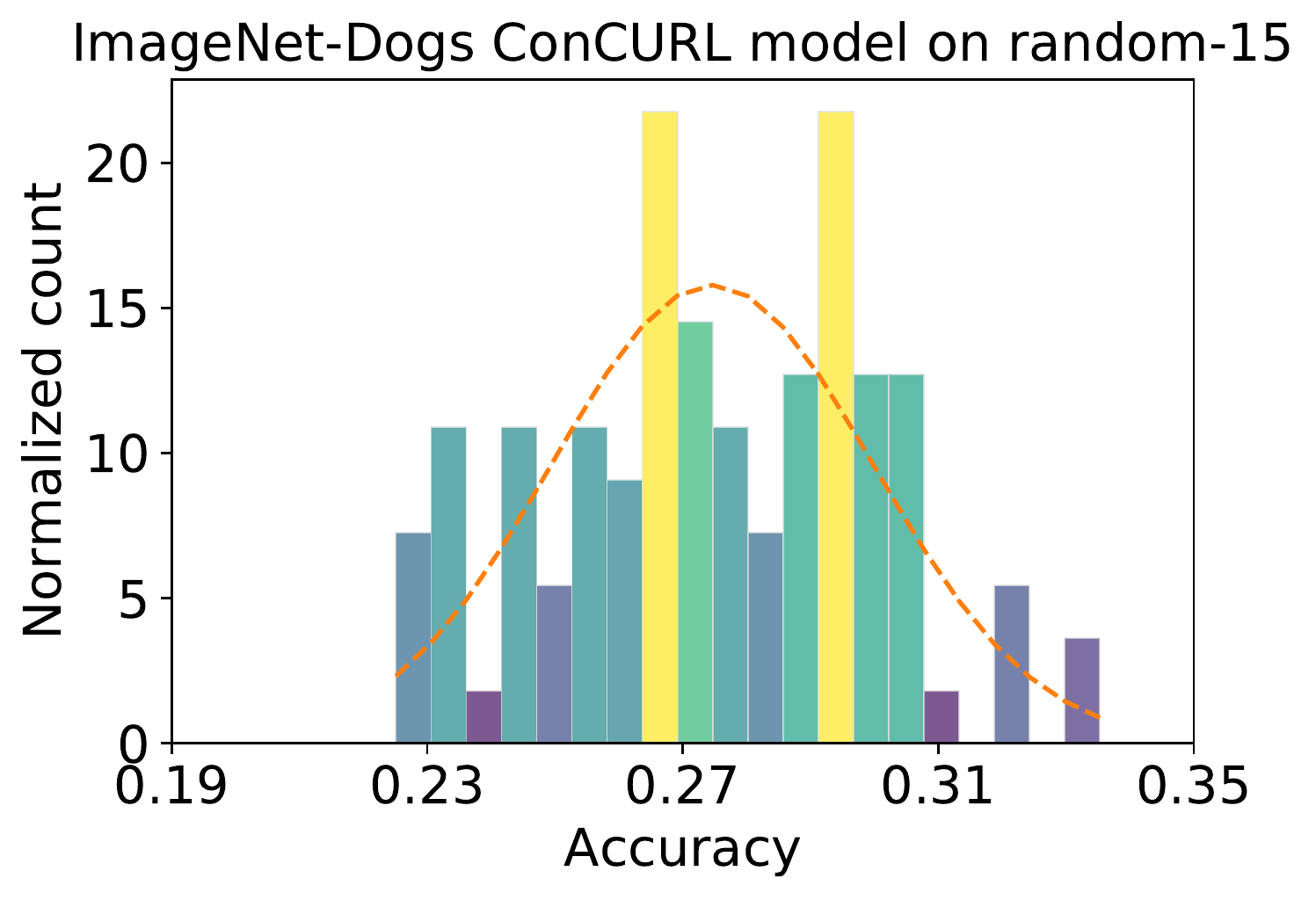}}
    \subfigure[]{\includegraphics[width=0.24\textwidth]{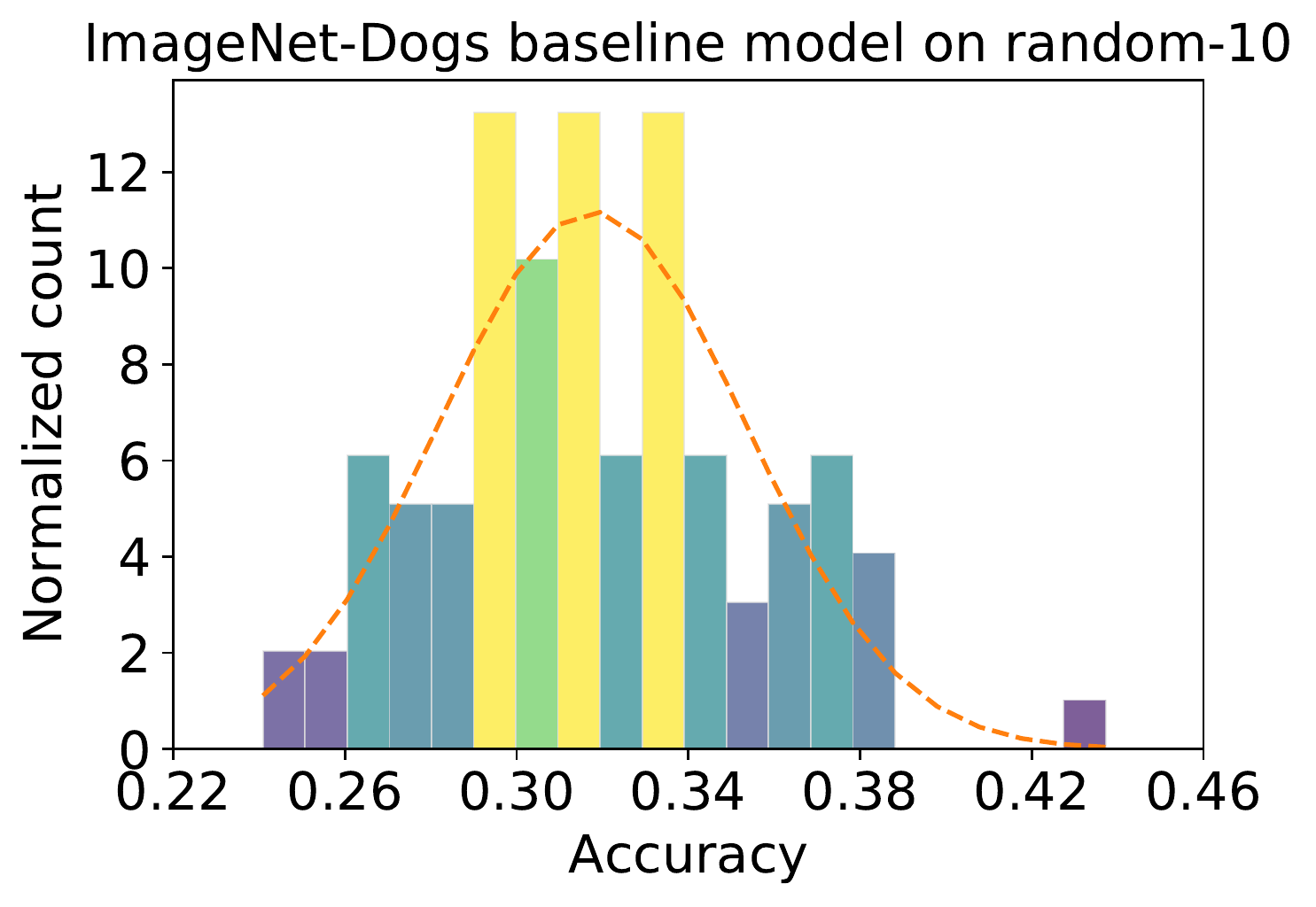}}
    \subfigure[]{\includegraphics[width=0.24\textwidth]{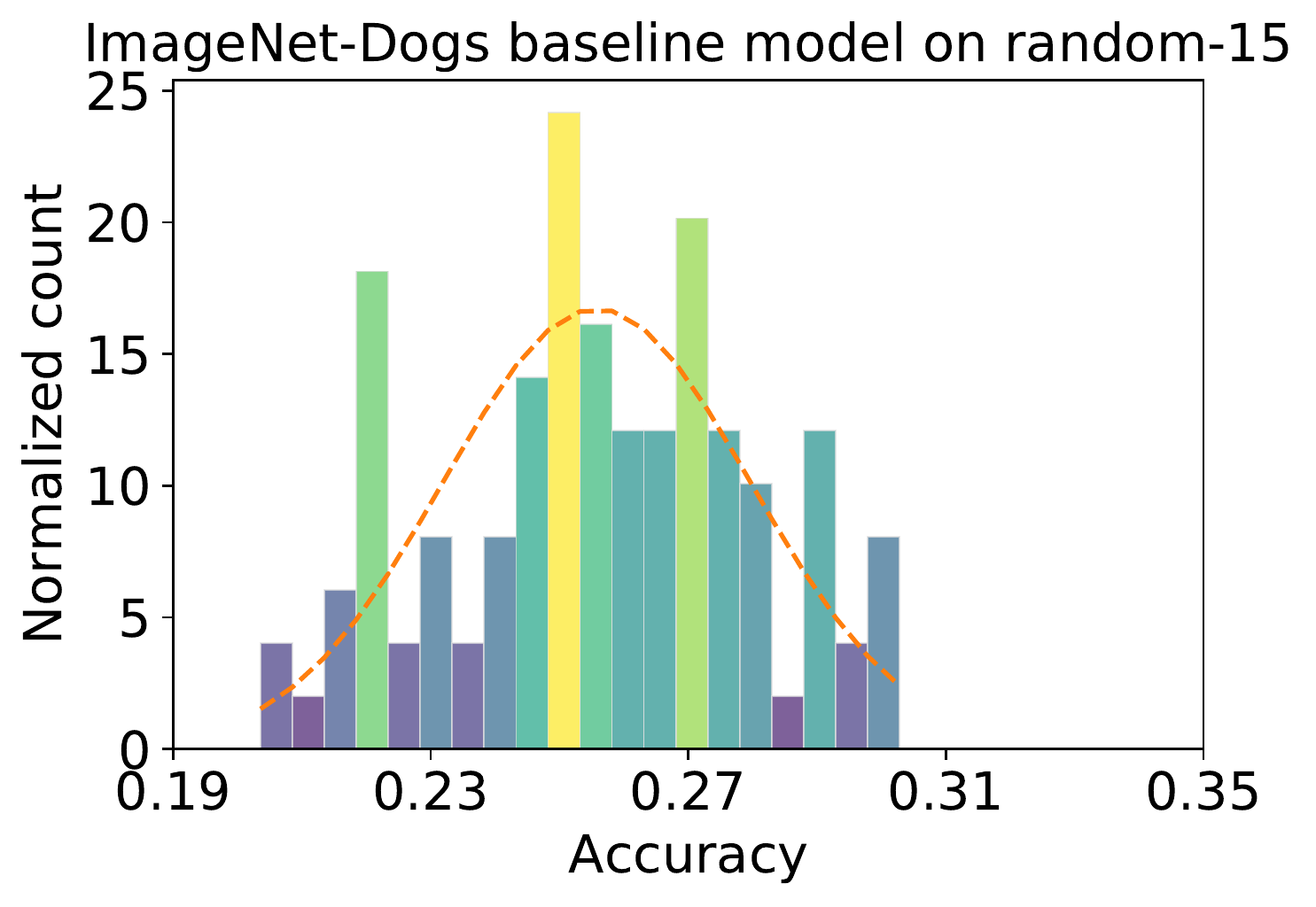}}
    \caption{Histogram of clustering accuracies for models trained on \idogs: (a) \pap\ model evaluated on random-10, (b) \pap\ model evaluated on random-15, (c) Baseline (ID) model evaluated on random-10, (d) Baseline (ID) model evaluated on random-15}
    \label{fig:ImageNetDogsConcurlvsIDrandom}
\end{figure}

Here, we compare the baseline model trained with ID \citep{wu2018unsupervised} with our proposed method \pap. We randomly sample 10 and 15 classes from the 1000-class ImageNet data and evaluate the clustering accuracy obtained on the training split of the data using the model trained on the original \iten\ and \idogs\ sets. We repeat the process 100 times for both the 10-class and 15-class datasets and call them the random-10 and random-15 datasets, respectively. Note that we do not retrain the model on the randomly sampled dataset; we only evaluate the model on this set. We show the histogram of the obtained accuracies for these 100 random datasets. In Figure~\ref{fig:ImageNet10ConcurlvsIDrandom}, we compare the accuracy of the \pap\ model and the baseline ID model trained on \iten\ on both random-10 and random-15. Along with the histogram, we show a Gaussian distribution (along the red dotted line) with first and second moments equal to the average and standard deviation of all accuracies, respectively. Similarly, in Figure~\ref{fig:ImageNetDogsConcurlvsIDrandom}, we show the accuracies obtained based on models trained on \idogs. Among the models trained on \iten, the baseline ID model performs slightly better than the proposed \pap\ model. The trend is reversed for the evaluation based on the model trained on \idogs, where \pap\ performs better than the baseline model. Even though the proposed method performs best with the max-performance strategy, it performs slightly worse on random-10. This result strengthens our argument regarding the need to go beyond the traditional reporting of maximum performance based on the ACC, NMI and ARI metrics.

\begin{figure}[htb!]
    \centering
    \includegraphics[width=1.\columnwidth]{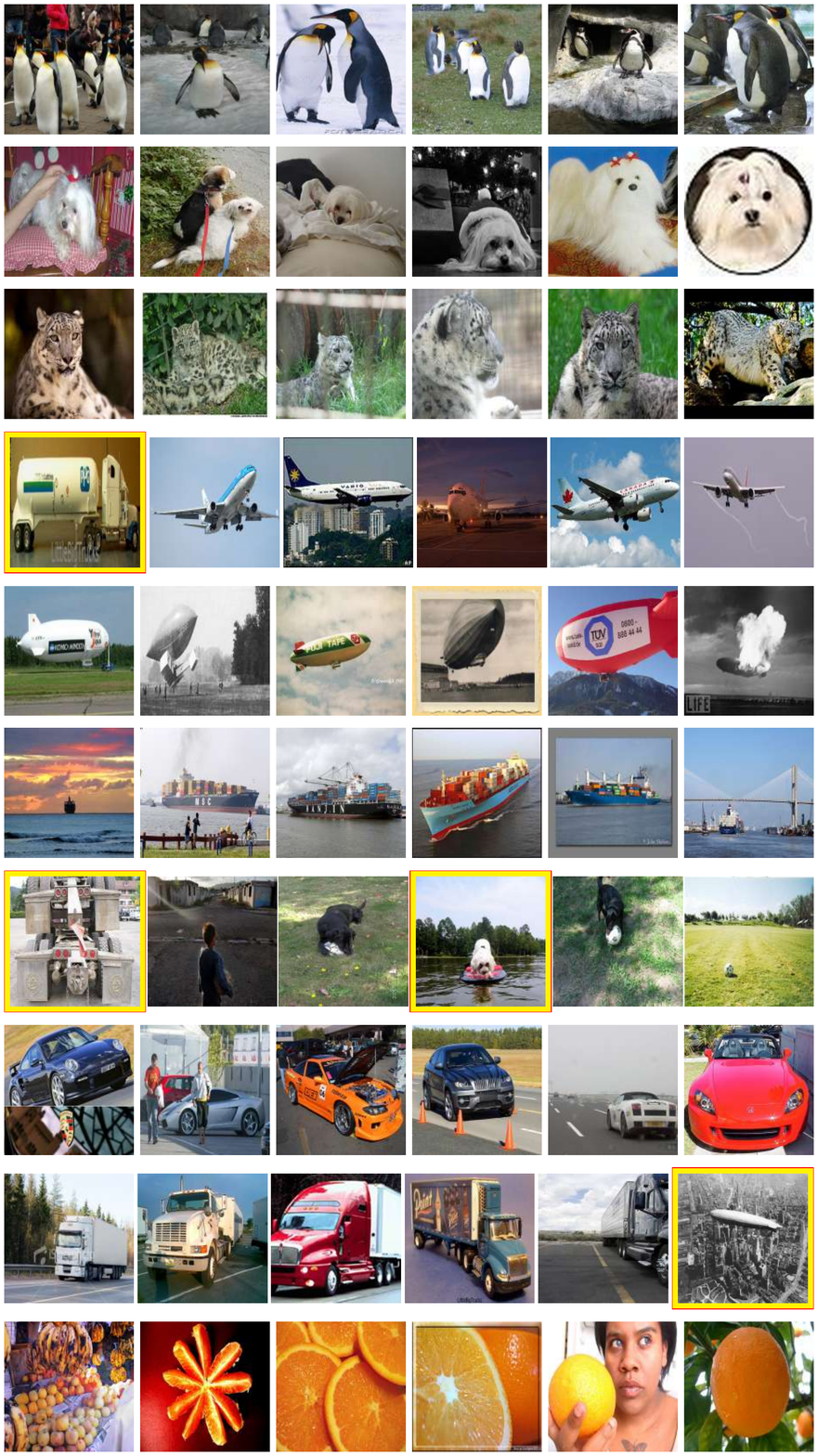}
    \caption{{Images from the same cluster: In \iten, we randomly sample six images from all ten clusters and show them above. Each row presents one cluster and images that have red-and-yellow borders are categorized incorrectly and should belong to different clusters.} }
    \label{fig:images_same_cluster}
\end{figure}

\begin{figure}[htb!]
    \centering
    \includegraphics[width=1.\columnwidth]{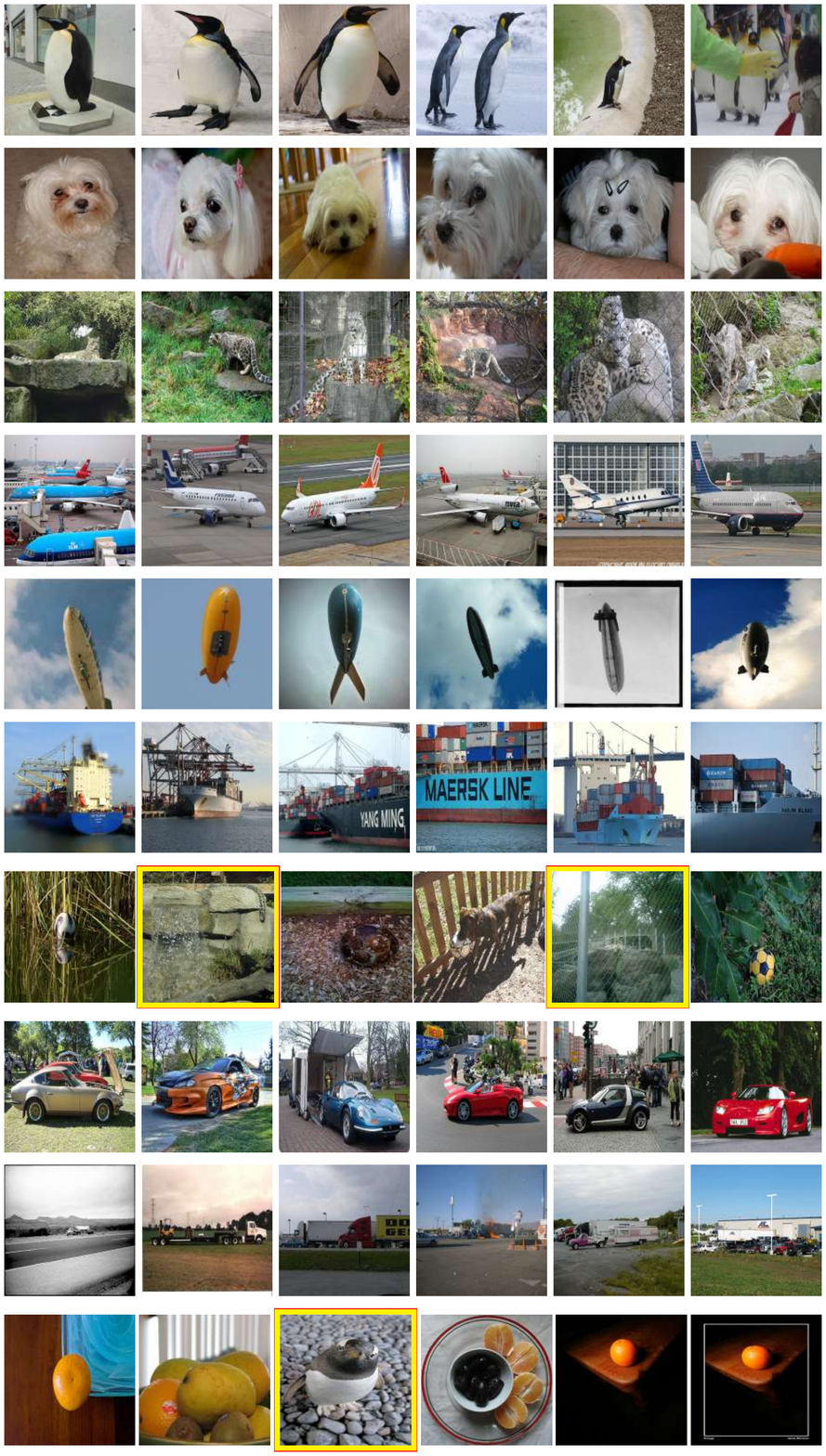}
    \caption{{Image retrieval: The first image in each row was used as a query (random samples from the dataset \iten), and the five images nearest to the query image were retrieved using their representations. Images with red-and-yellow borders are retrieved incorrectly}}
    \label{fig:images_retrieved}
\end{figure}

\subsection{Cluster Visualizations}
In this section, we provide two visualizations of the \iten\ dataset. In Figure~\ref{fig:images_same_cluster}, each row presents randomly drawn images from each cluster. Images that have red-and-yellow borders are categorized incorrectly and should belong to different clusters. For example, the first image in the fourth row should be in the truck category, but it is categorized as an airliner. In the soccer ball category, there are two mistakes: the first image should be categorized as a truck, and the fourth image should be categorized as a dog. In the ninth row, the last image should be categorized as an airship, but it is categorized as a truck.

To check whether the learned representations that are closest to each other belong to the same category, we use a retrieval task. In Figure~\ref{fig:images_retrieved}, we show the results. The first image in each row was used as a query (random samples from the dataset), and the five images nearest to the query image were retrieved using their representations. Ideally, one would expect all retrieved images to belong to the same category as the query image. For the example from the soccer ball category, the first image retrieved does not belong to this category; however, both images have water as their main feature. In the last row, the second image retrieved is a penguin and ideally should not be one of the closest matches to an image in the orange category.

\section{Ablation Studies}
\label{sec:ablation}
Although the proposed method is trained in an end-to-end manner, each component of the method may have a different impact on the results. We conduct various controlled experiments to quantify the impact of the losses (Section~\ref{sec:loss_ablation}),
the data augmentation methods (Section~\ref{sec:ablation_augmentations}), the image resolution (Section~\ref{sec:ablation_resolution}), the number of transformations (Section~\ref{sec:ablation_components}), the dimensionality of each transformation (Section~\ref{sec:ablation_components}) and the architecture choice (Section~\ref{sec:ablation_architecture_choice}).

\subsection{{Effects of the Loss Terms}}
\label{sec:loss_ablation}

{
We consider the following scenarios: train with only loss $L_{\mathcal Z}$ (Consensus Loss),
train with only loss $L_{b}$ (ID), train with both losses $L_{b} + L_{\mathcal Z}$ (\pap). We do this for \cten,~\chundred~ datasets. We then compare the clustering performance of all three scenarios and observe that training with both losses improves the performance. 
Additionally, we also compare the loss trajectories during training. We compare the $L_b$ loss for the case when we train only $L_b$ and for the case when we train only $L_{\mathcal{Z}}$. We repeat this for loss $L_{\mathcal{Z}}$. The results are summarized in Table~\ref{table:loss_ablation}, where we observe that training with both the losses provides a much better clustering performance as compared to training with the losses individually.
}

\begin{table}[htb!]
\renewcommand\tabcolsep{3.1pt}
\caption{{Ablation on the losses during training}}
\label{table:loss_ablation}
\centering
\begin{tabular}{|c|ccc|ccc|}
\hline

\multirow{2}{*}{Loss used}             & 
\multicolumn{3}{c|}{\chundred}&\multicolumn{3}{c|}{\cten}\\
\cline{2-7} & ACC  & NMI  & ARI& ACC&NMI&ARI  \\
\hline\hline
        $L_b$ only \tablefootnote{For \cten, \chundred~ we provide results of rerun that is better than reported in \citep{taoclustering}} & 0.437 & 0.429 & 0.259 & 0.826 & 0.732 & 0.684  \\
        $L_{\mathcal{Z}}$ only & 0.302 & 0.333 & 0.145 & 0.739 & 0.667 & 0.539  \\
        $L_b+L_{\mathcal{Z}}$ (\pap) & \textbf{0.479} & \textbf{0.468} & \textbf{0.303}  & \textbf{0.846} & \textbf{0.762} & \textbf{0.715} \\  
\hline
\end{tabular}
\end{table}

\begin{figure}[h]
    \centering
    \subfigure[]{\includegraphics[width=0.47\textwidth]{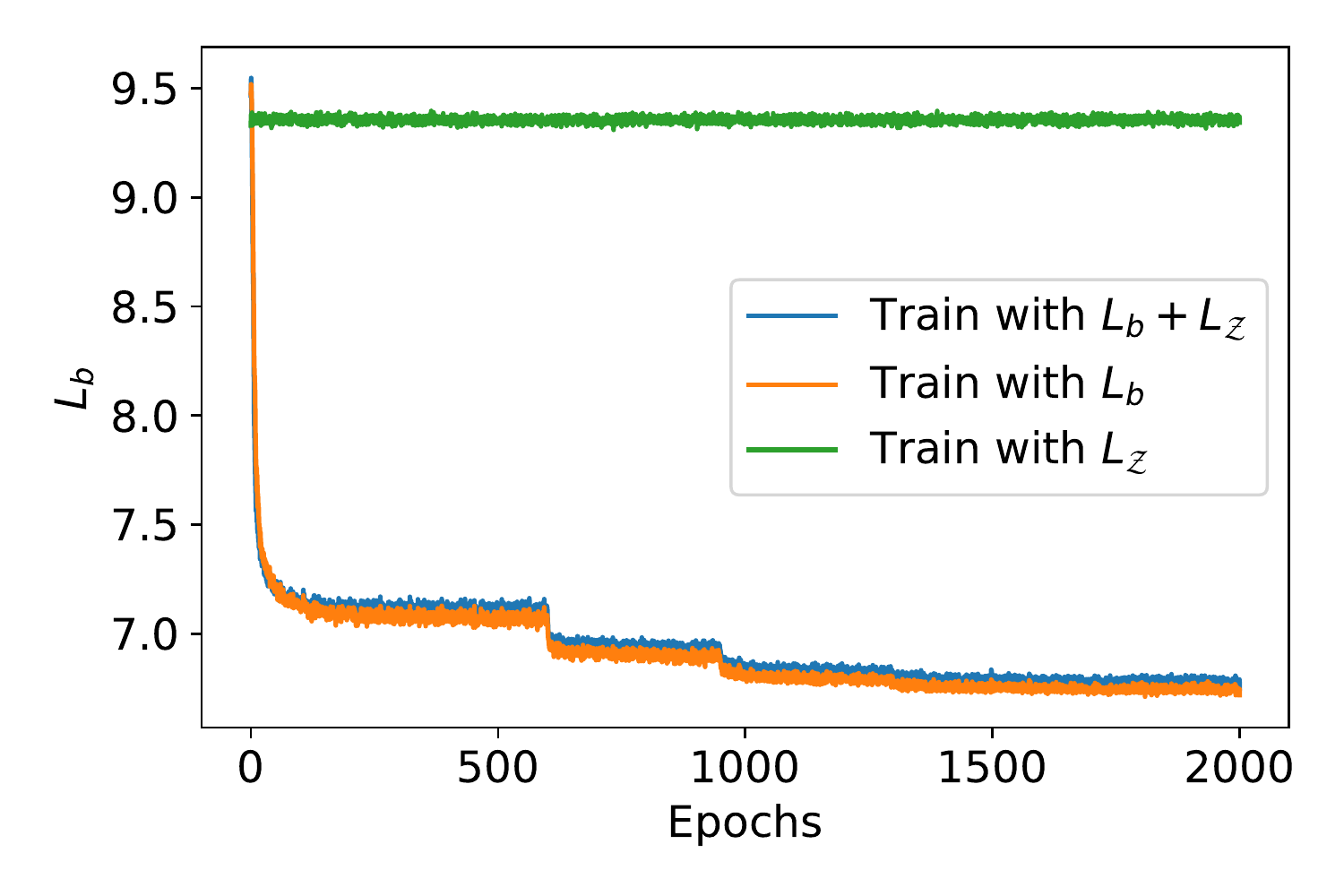}
    \label{fig:lb}
    } 
    \subfigure[]{\includegraphics[width=0.47\textwidth]{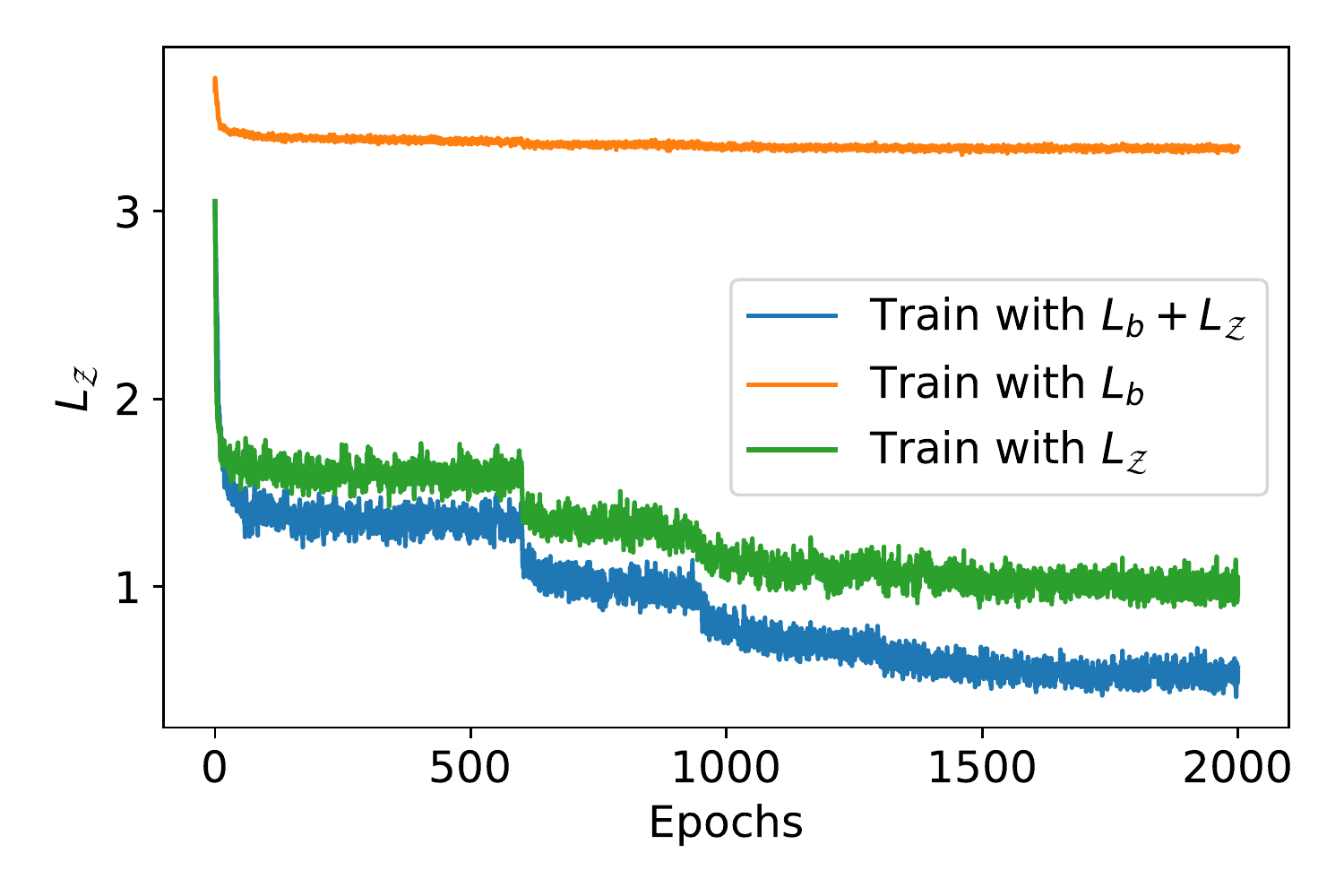}
    \label{fig:lz}
    } 
    \caption{{
We perform an ablation over the losses used and compare the loss values for each of the cases for \chundred. 
}}
    \label{fig:loss_ablation_loss}
\end{figure}

\begin{figure}[h]
    \centering
    \subfigure[]{\includegraphics[width=0.32\textwidth]{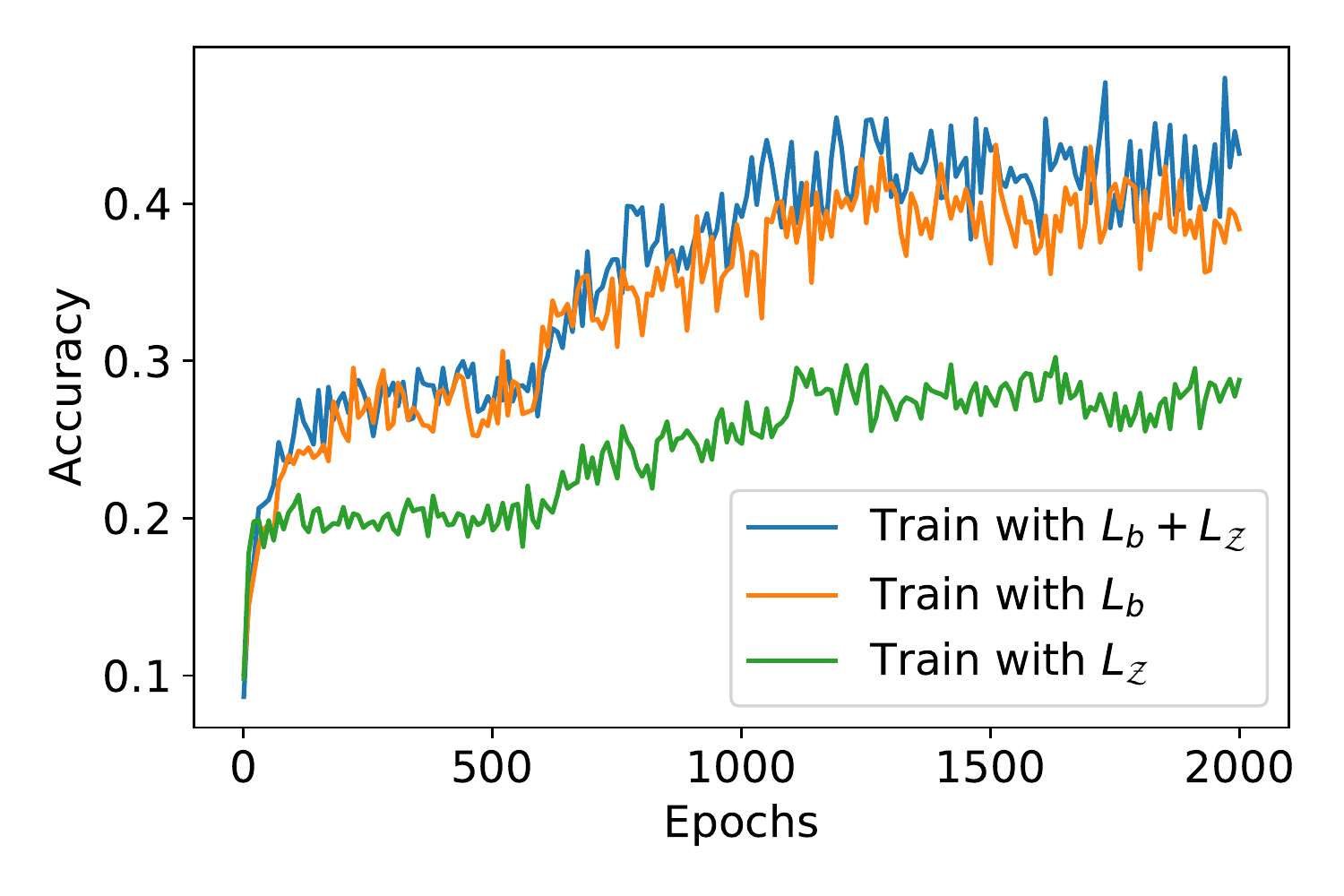}} 
    \subfigure[]{\includegraphics[width=0.32\textwidth]{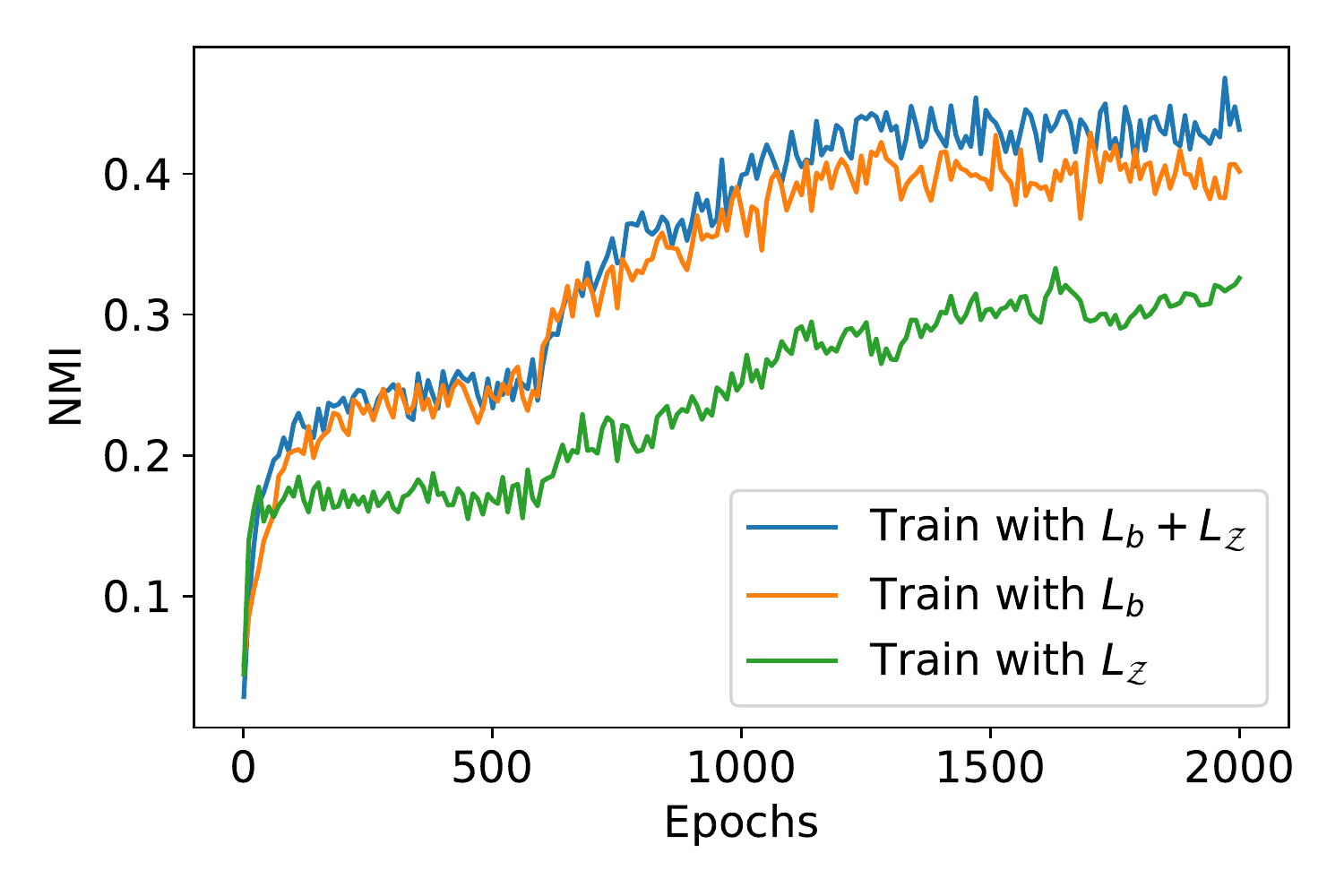}} 
    \subfigure[]{\includegraphics[width=0.32\textwidth]{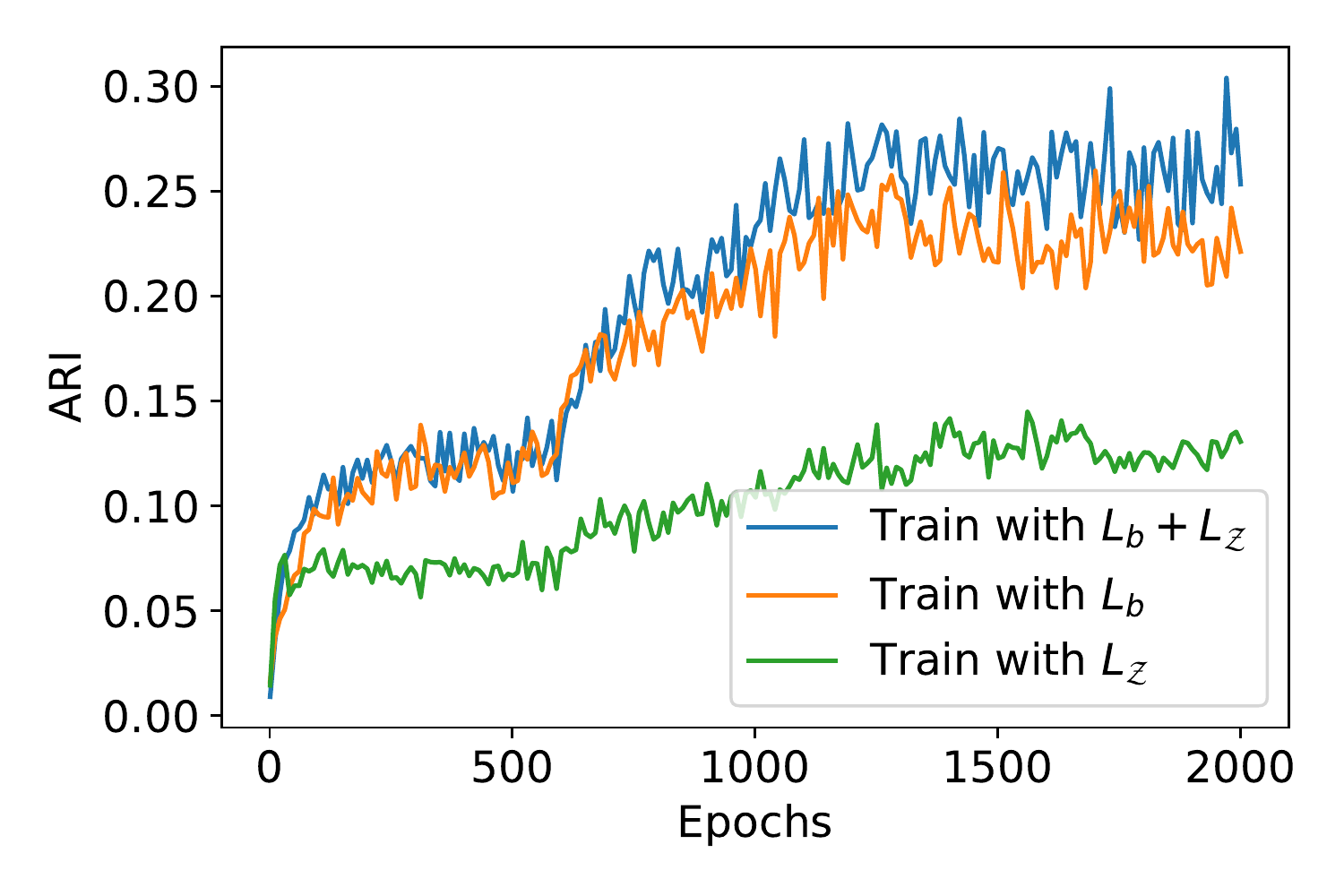}} 
    \caption{{
    We compare the clustering metrics for ablation over the losses used for \chundred~
    and observe that Train with $L_b+L_{\mathcal{Z}}$ (\pap) outperforms training with only $L_b$ or $L_{\mathcal{Z}}$. 
    }}
    \label{fig:loss_ablation_res}
\end{figure}

{
The exemplar consistency trains with the objective of classifying each data point into its own class. The population and consensus consistencies train without regarding for discrimination among the individual data points. An algorithm that is trained with only the consensus loss therefore is not effective in discriminating individual data points. From Figure~\ref{fig:lb}, we can observe that when we train with only $L_{\mathcal{Z}}$, we do not observe any improvement in the loss $L_{b}$; when we train with $L_b+L_{\mathcal{Z}}$, we observe a similar trajectory as training only with $L_b$. This shows that training with $L_{\mathcal{Z}}$ does not conflict with $L_b$ loss.
}

{
On the other hand, it is possible that an algorithm that is trained to discriminate individual data points can show some improvement on the consensus loss $L_{\mathcal{Z}}$. 
From Figure~\ref{fig:lz}, we can observe that $L_{\mathcal{Z}}$ is lesser when trained with $L_b+L_{\mathcal{Z}}$ than when training only with $L_{\mathcal{Z}}$. We observe a small decrease in $L_{\mathcal{Z}}$ value when trained only with $L_b$. 
This shows that optimizing $L_b$ helps to some extent in achieving a better $L_{\mathcal{Z}}$. 
}

{
From this discussion, we observe that both losses contribute differently to the training without much interference or conflicts. They indeed complement each other as we observe improved clustering performance for \pap~from Table~\ref{table:loss_ablation} and Figure~\ref{fig:loss_ablation_res}. 
}


\subsection{Effects of Data Augmentation Methods}
\label{sec:ablation_augmentations}

\begin{table}[h]
\caption{Data augmentation details }
\label{tab:DAType}
\centering
\begin{tabular}{|c | c | c | c |} 
\hline
Data Augmentation Type    & \textbf{Abbreviation}   & \textbf{\cten}  & \textbf{\chundred} \\ \hline
All data augmentations       & All DA   &  0.8459   & 0.4798   \\ \hline
No random horizontal flip                 & No RHF & 0.7689 & 0.4551 \\ \hline
No random gray scale  & No RGS     & 0.7750   &  0.4357     \\ \hline
No randomly applied blur  & No RAB    & 0.7907  & 0.4295     \\ \hline
No color jitter              & No CJ   & 0.6463  & 0.2536     \\ \hline
No random resized cropping             & No RRC   & 0.2988  & 0.1322    \\ \hline
\end{tabular}
\end{table}

\begin{figure}[h]
    \centering
    \includegraphics[width=0.8\columnwidth]{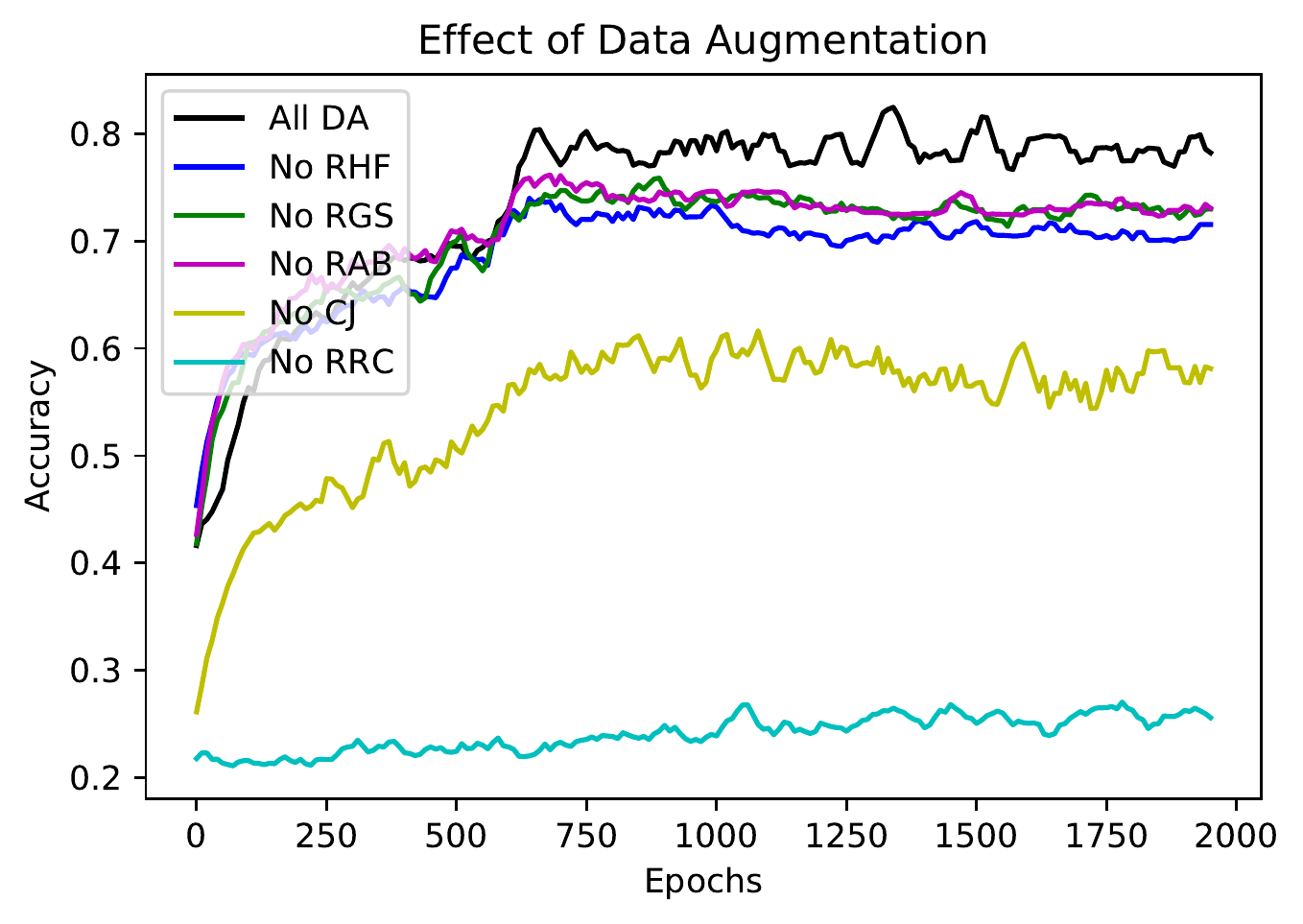}
    \caption{Effect of data augmentation on \cten}
    \label{fig:Effect_aug_CIFAR10}
\end{figure}

\begin{figure}[h]
    \centering
    \includegraphics[width=0.8\columnwidth]{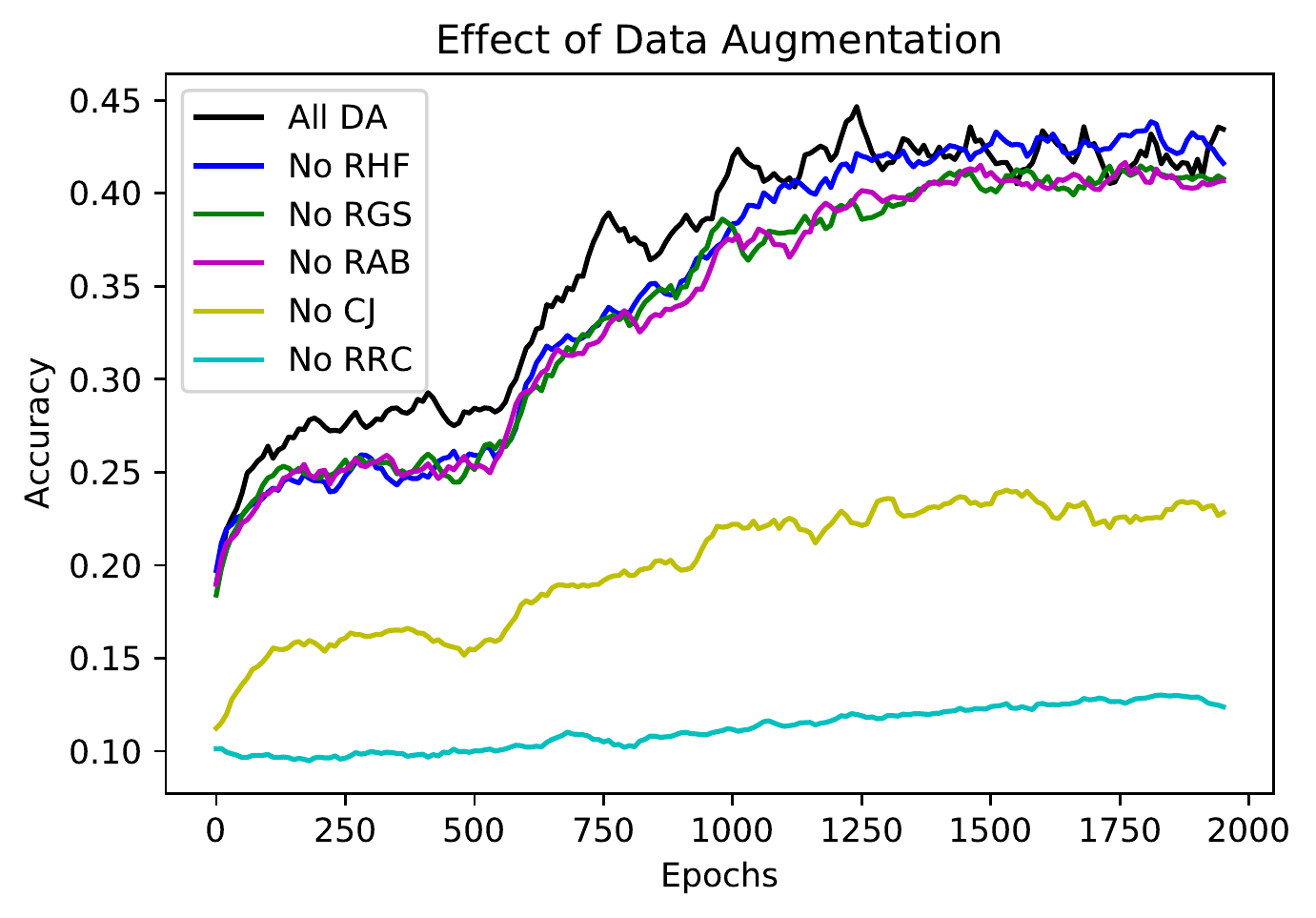}
    \caption{Effect of data augmentation on \chundred}
    \label{fig:Effect_aug_CIFAR100}
\end{figure}

Augmenting the training data is a standard technique for training deep learning methods \citep{ shorten2019survey}. The backbones used in this study rely on the different views that are generated by applying different augmentations to the input image. Recently, \cite{tian2020makes} investigated the impact of data augmentation on contrastive learning methods and shed some light on this topic. In our setting, we would like to quantify the impacts of several data augmentations on the consensus loss. 

In Table~\ref{tab:DAType}, we show the maximum accuracy achieved when all data augmentation approaches are used and when we skip one data augmentation technique at a time. When all data augmentation methods are used, the maximum accuracies achieved are 0.8459 and 0.4798 on \cten\ and \chundred, respectively. When random resized cropping data augmentation is dropped, we obtain the maximum accuracy drops for both datasets, followed by color jitter. Other data augmentation techniques are important for obtaining the best possible accuracy but do not have as much of an effect as color jitter and random resized cropping. In Figures \ref{fig:Effect_aug_CIFAR10} and \ref{fig:Effect_aug_CIFAR100}, we show how the running mean of accuracy progresses during training for each of the experiments in Table~\ref{tab:DAType}.

\subsection{Effect of Image Resolution}
\label{sec:ablation_resolution}
\begin{table}[h]
\centering
\caption{Effects of different resolutions for \stl, \iten\ and \idogs}
\label{tab:res}
\begin{tabular}{|c|c|c|c|c|c|}
\hline
& $32$ &$64$ & $96$ & $160$ & $224$ \\ \hline
\textbf{\stl}        & 0.531       & 0.724       & 0.749       & NA           & NA           \\ \hline
\textbf{\iten}   & NA          & NA          & 0.887       & 0.958        & 0.946        \\ \hline
\textbf{\idogs} & NA          & NA          & 0.629       & 0.695        & 0.679        \\ \hline
\end{tabular}
\end{table}

Image resolution is often considered a free parameter \citep{niu2020gatcluster}, and however, its effect on clustering performance is not evaluated rigorously in most works. We try to quantify the effects of different resolutions to the greatest extent possible, given that some datasets are available only at specific resolutions. For \stl, we use $32\times32$, $64\times64$ and $96\times96$ resolutions. For ImageNet-$10$ and ImageNet-Dog-$15$, we use $96\times96$, $160\times160$ and $224\times224$ resolutions. The results are given in Table \ref{tab:res}.

The best performance for \iten\ and \idogs\ is obtained at a resolution of 160, and for \stl, the best performance is obtained at a resolution of 96. It is not clear why \iten\ and \idogs\ do not yield the best performance at high resolutions, and further investigation is needed; we keep this as an open problem.

\subsection{Distribution of Accuracies Across the Set of Hyperparameters}
\label{sec:ablation_components}
\begin{table}[h]
\caption{Hyperparameters and the range values used for the experiments}
\label{tab:hyper}
\centering
\begin{tabular}{|c | c | c | c |} 
\hline
    & \textbf{min}   & \textbf{max}  & \textbf{step size} \\ \hline
$\tau$ = temperature parameter       & 0.3   & 1    & 0.1   \\ \hline
$l$ = learning rate                 & 0.015 & 0.09 & 0.015 \\ \hline
$\eta$ = natural log of the number of transformations  & 0     & 7    & 1     \\ \hline
$d$ = dimensionality of the projection space              & 80    & 168  & 8     \\ \hline
\end{tabular}
\end{table}

\begin{table}[h]
\caption{Hyperparameters for obtaining maximum performance}
\label{tab:hypermaxperf}
\centering
{
\begin{tabular}{|c | c | c | c | c| c|} 
\hline
    &  \stl  & \iten   & \idogs  & \cten  & \chundred \\ \hline
$\tau$       & 0.8   & 1.0   & 0.5 & 0.85 &  0.35 \\ \hline
$l$                  & 0.03 & 0.03 & 0.06 & 0.015 & 0.06  \\ \hline
$\exp(\eta)$  & 4     & 1    & 64 &  64 & 16  \\ \hline
$d$             & 80    & 104  & 136  & 152  & 32   \\ \hline
\end{tabular}
}
\end{table}

\begin{figure}[htb!]
    \centering
    \subfigure[\stl: log of the number of transformations  ]{\includegraphics[width=0.48\textwidth]{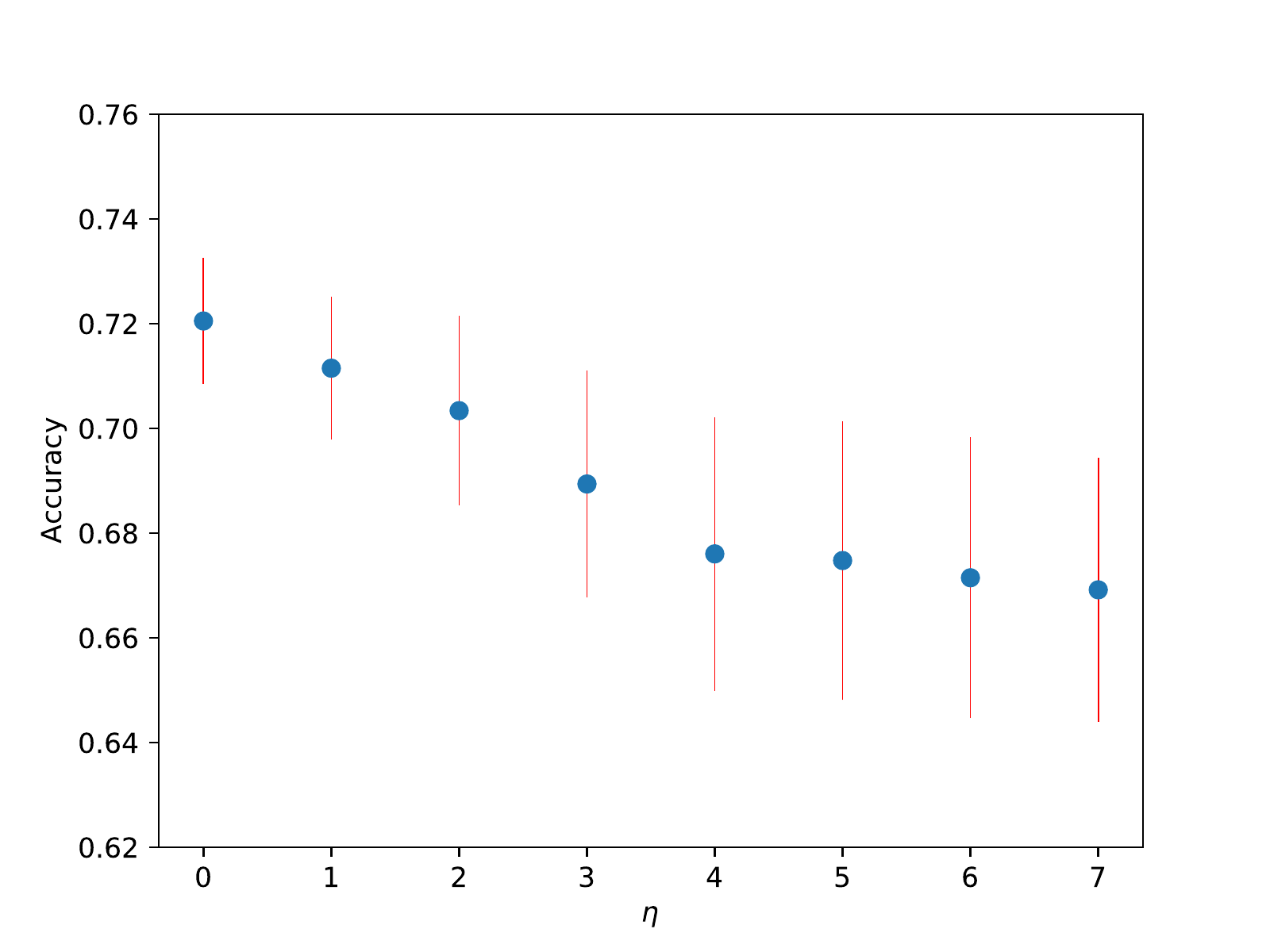}} 
    \subfigure[\stl: dimensionality of the projection space]{\includegraphics[width=0.48\textwidth]{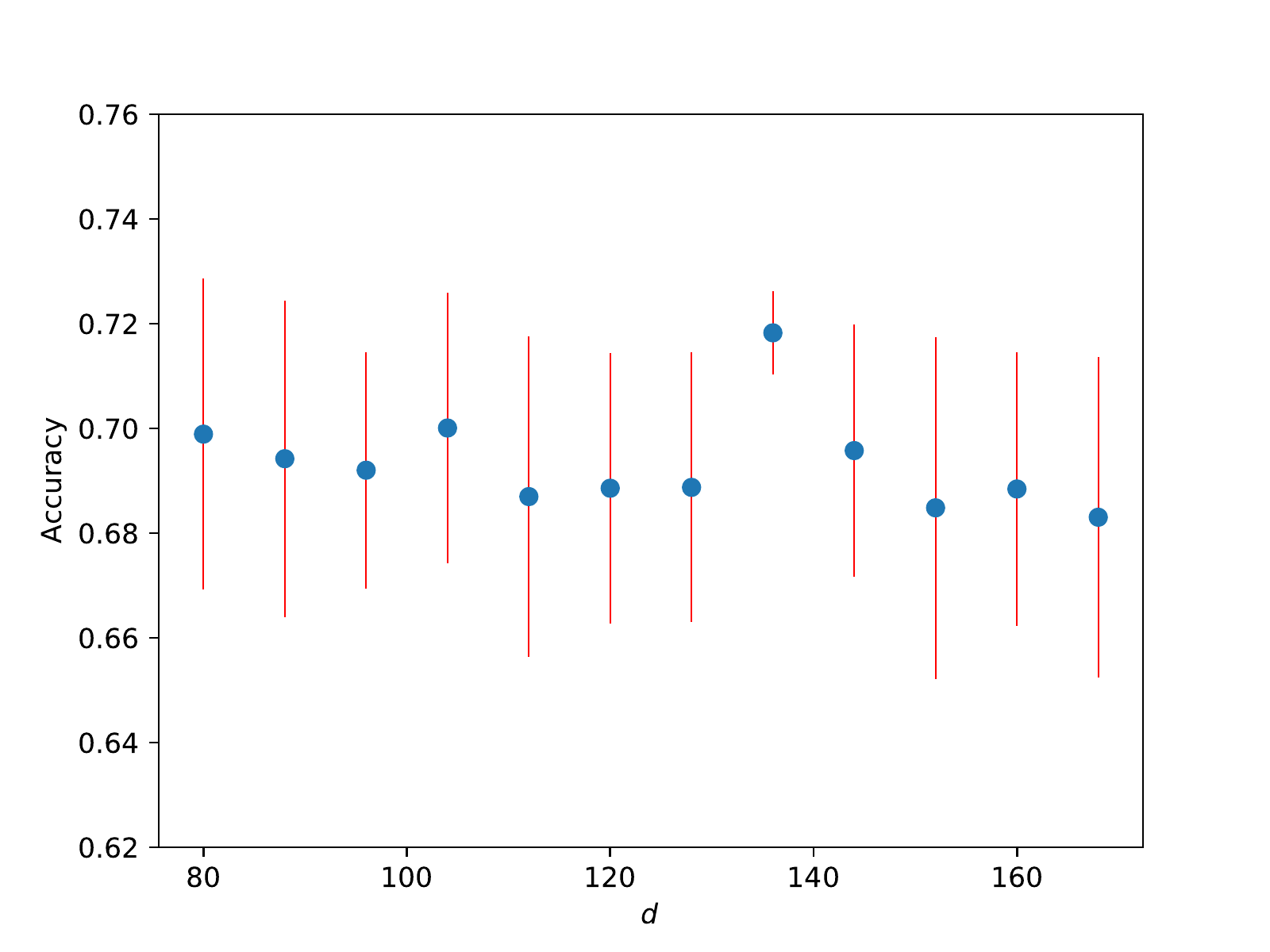}} 
    \subfigure[\chundred: log of the number of transformations ]{\includegraphics[width=0.48\textwidth]{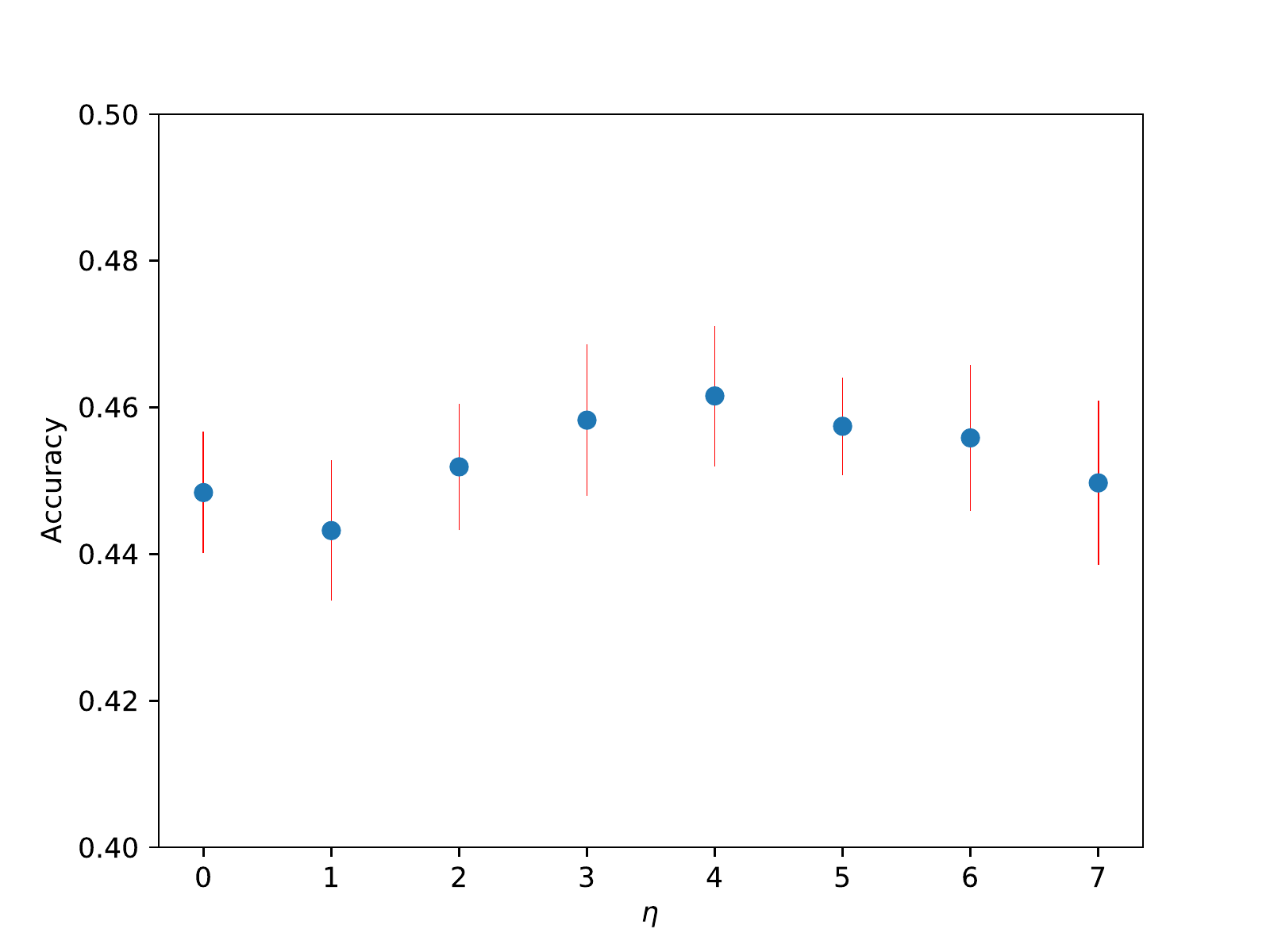}}
    \subfigure[\chundred: dimensionality of the projection space]{\includegraphics[width=0.48\textwidth]{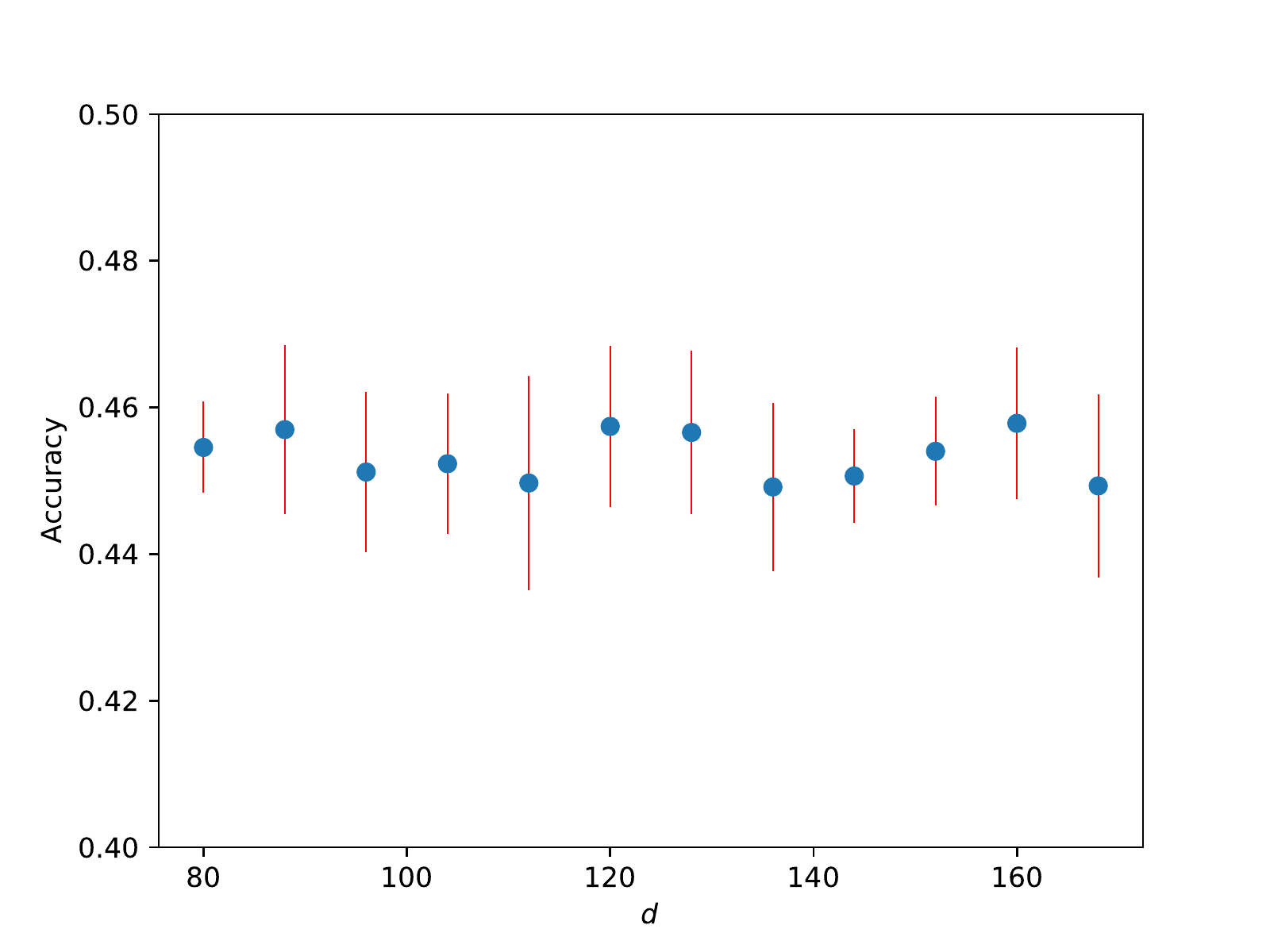}}
    \caption{Components of the consensus loss; ablation of \stl\ and \chundred }
    \label{fig:hyper}
\end{figure}

The proposed consensus loss has two parameters. The first is the number of transformations used, and the second is the dimensionality of the projection space. To understand the proposed loss, we conduct a detailed experimental study on \stl\ and \chundred\footnote{We conduct a similar but smaller study on the remaining dataset, and we observe similar trends.}. The hyperparameters used are given in Table \ref{tab:hyper}. 

Due to the sheer number of conducted experiments, we supply the summary statistics obtained on a random set. We report the empirical mean and standard deviation of the marginal distribution of the quantity under investigation. Let $P_{\tau,\eta,d,l}$ be the joint distribution over the hyperparameters $\tau$ (temperature parameter), $l$ (learning rate), $\eta$ (natural log of the number of transformations) and $d$ (dimensionality of the projection space). We consider $n_h$ as the number of distinct values used in the experiment for each hyperparameter $h \in \{ \tau,\eta,d,l \}$ based on Table~\ref{tab:hyper}. We the denote accuracy from each experiment based on the hyperparameters used as $a_{\tau,\eta,d,l}$ .
Let $P_{h_i \vert h_j}$ be the conditional marginal distribution of hyperparameter $h_i$ given $h_j$ and the conditional empirical mean of $P_{h_i \vert h_j}$ be $m(P_{h_i \vert h_j})$. In this case, the conditional empirical mean $m(P_{h_i \vert h_j})$ when $h_i = d$ and $h_j=\tau$ can be calculated using
$m(P_{d \vert \tau}) = \frac{1}{n_{\eta} \times n_{l}} \sum_{\eta} \sum_{l} a_{\tau,\eta,d,l}$. The conditional empirical means and standard deviations of other hyperparameters are calculated in the same way.
In Figure~\ref{fig:hyper}, we show each conditional empirical mean with a blue dot, and each red line around a dot represents a standard deviation. For both \stl\ and \chundred, we see a trend regarding the number of projections. For \stl, the smaller the number of random projections, the better the results are, and for \chundred, increasing the number of random projections is helpful for improving the clustering accuracy up to some point. Note that when the number of random projections is equal to zero, our setting is equivalent to the baseline ID model, and our approach always performs better than ID. This means that the optimal number of random projections is greater than or equal to one. There is no such clear trend in the number of dimensions of the random projections.  
\begin{figure}
    \centering
    \subfigure[]{\includegraphics[width=0.48\textwidth]{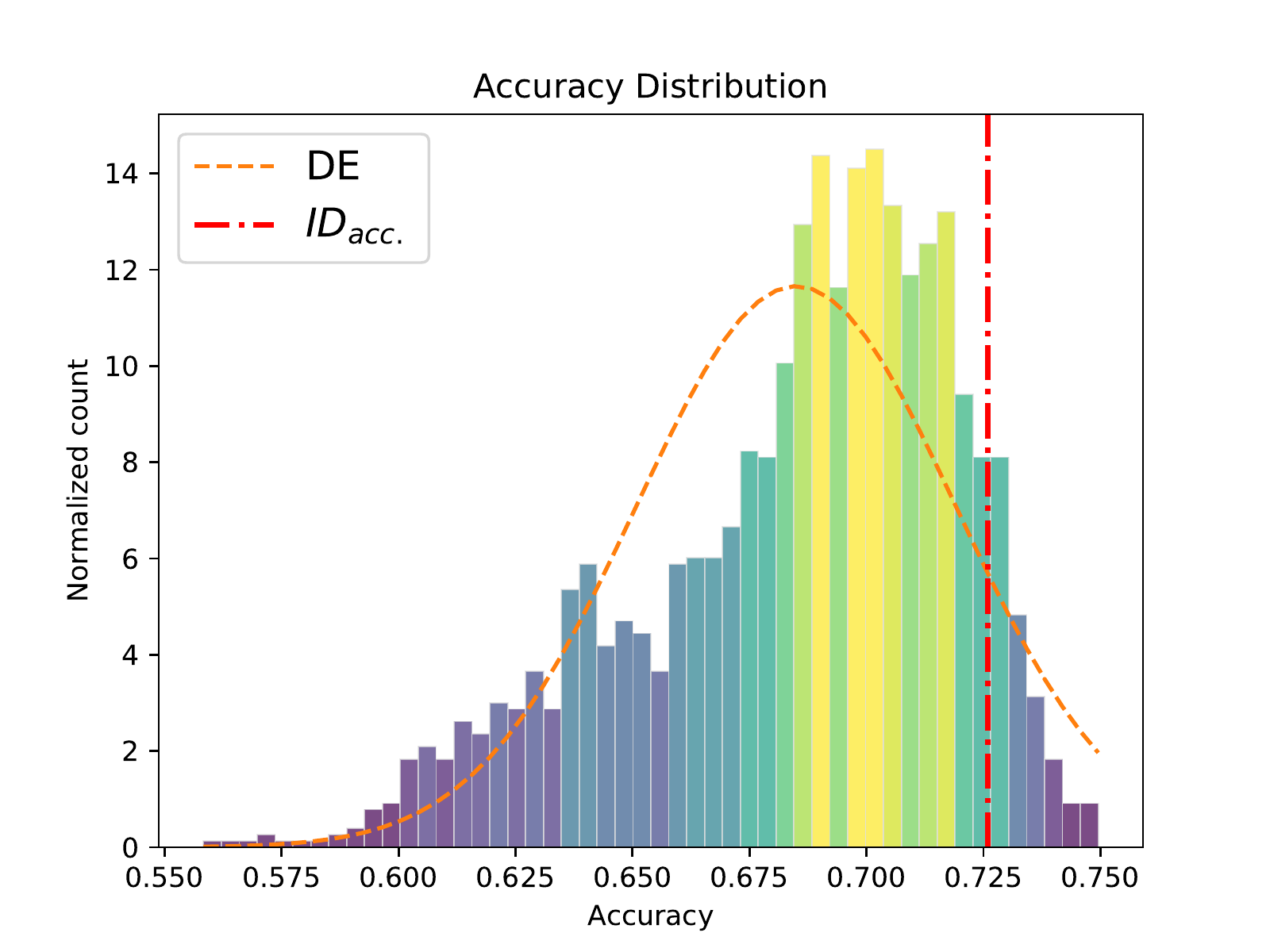}} 
    \subfigure[]{\includegraphics[width=0.48\textwidth]{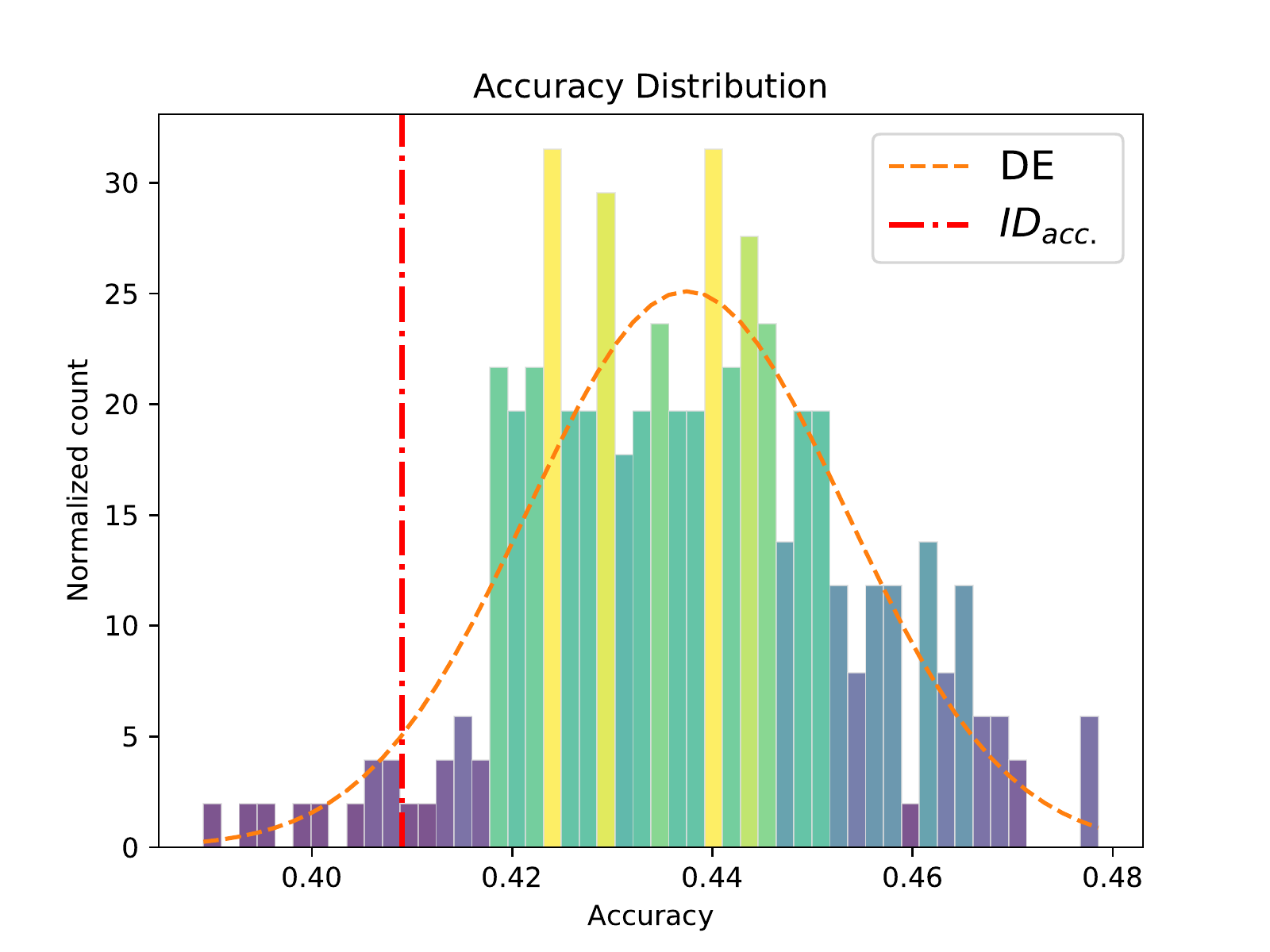}} 
    \caption{The dotted red lines show the accuracy of the baseline, i.e., ID~\cite{taoclustering}, on the corresponding dataset, and DE is the density estimate of the empirical distribution: (a) Empirical accuracy distribution for \stl, (b) Empirical accuracy distribution for \chundred }
     \label{fig:hyper_acc}
\end{figure}

The max-performance procedure provides some insights into the performance of the algorithms at hand, although it does not provide the whole picture because it does not consider the robustness of the performance differences. In Table~\ref{tab:hypermaxperf}, we give the hyperparameters that yield the max performance. In other words, finding a hyperparameter set that yields better performance than the baseline is the core idea behind the max-performance procedure. We ask the following question: given a hyperparameter grid, how likely is our method to achieve better accuracy than the baseline? In Figure \ref{fig:hyper_acc}, we report the empirical accuracy distributions on \stl\ and \chundred\ for all hyperparameters given in Table \ref{tab:hyper}. The red dotted lines show the corresponding baseline accuracy for each dataset. For \stl, only approximately $12.5\%$ of the hyperparameter sets yield better results than the baseline. On the other hand, for \chundred, approximately $90\%$ of the hyperparameter sets yield better results than the baseline. In other words, it does not require a significant amount of computational power to find a better model than the state-of-the art models for \chundred; however, the situation is the opposite for \stl. The results given in Figure \ref{fig:hyper_acc} suggest that when comparing models, multiple metrics need to be considered, not only the max-performance procedure.

\subsection{Effect of Architecture Choice}
\label{sec:ablation_architecture_choice}
\begin{figure}
    \centering
    \includegraphics[width=0.8\columnwidth]{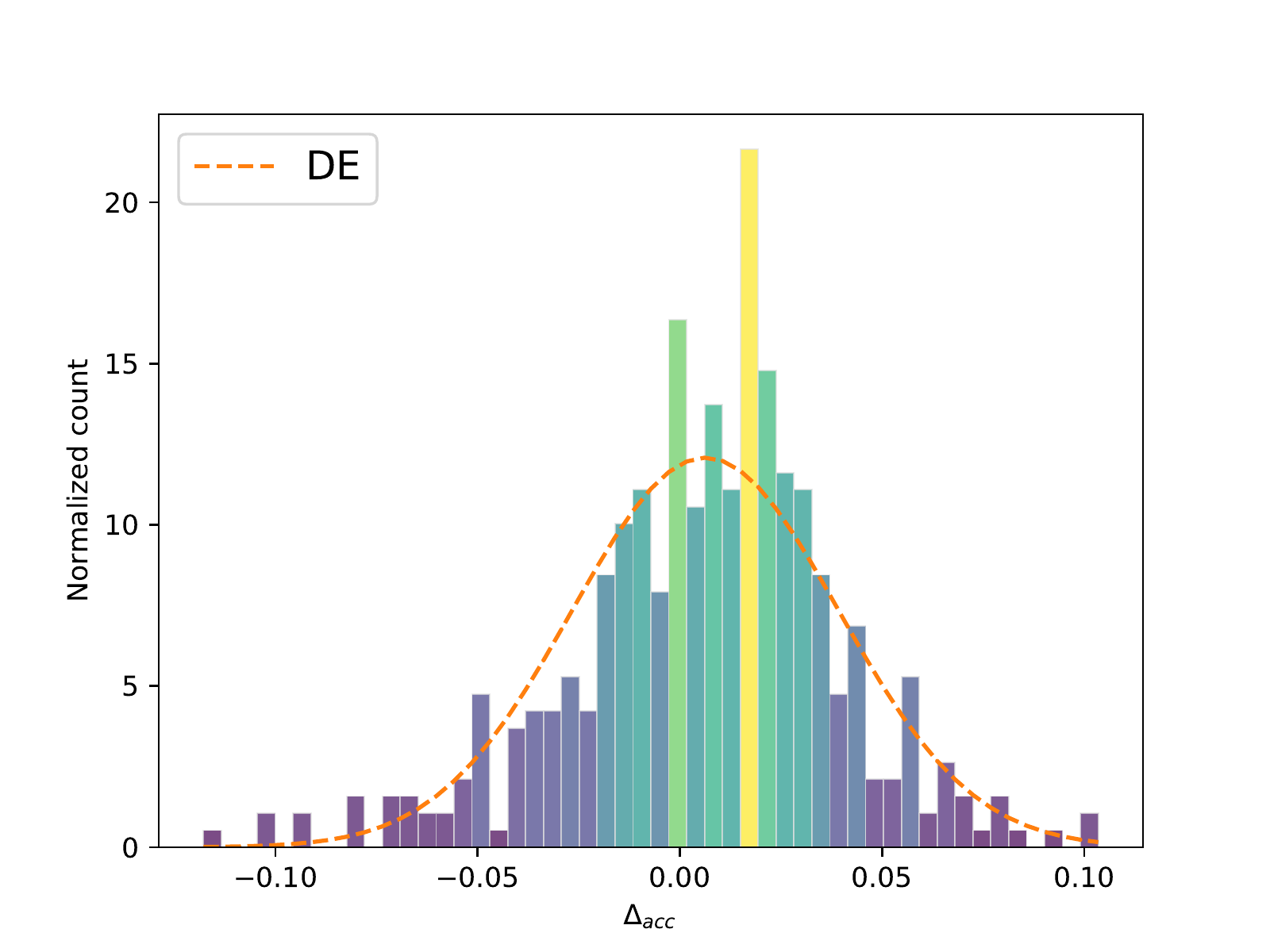}
    \caption{Empirical distribution of the performance difference between Residual Network (ResNet)-50 and ResNet-18. DE is the density estimate of the empirical distribution. }
    \label{fig:effect_of_arch}
\end{figure}
In this work, we use ResNet-18 and ResNet-50 as network architectures.
For both ResNet-18 and ResNet-50, we sweep over the same set of hyperparameter choices, i.e., the temperature, number of projections and projection dimensionality, and report the results {for \iten~dataset with image resolution of 160$\times 160$}. Figure \ref{fig:effect_of_arch} shows the distribution of $\Delta_{acc}$, which is defined as the accuracy difference between ResNet-50 and ResNet-18. Figure \ref{fig:effect_of_arch} indicates that ResNet-50 slightly outperforms ResNet-18, i.e., the mean difference is approximately $0.5\%$.

\subsection{{Runtime Comparison}}
\begin{figure}[h]
    \centering
    \includegraphics[width=0.5\textwidth]{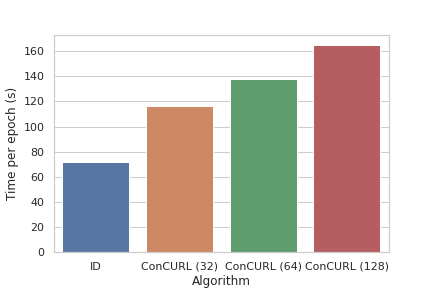}
    \caption{Comparison of the runtimes per epoch for ID and \pap~ on \cten~ dataset. We vary the number of random transformations in the computation of the consensus loss and is mentioned in paranthesis.}
    \label{fig:timing_comparison}
\end{figure}
{
To study the runtime of the proposed method, we compare the time taken per epoch for the baseline ID algorithm and the proposed algorithm. Due to the additional loss computation, the time taken to run the proposed algorithm is higher which can be observed from Figure~\ref{fig:timing_comparison}. The additional time taken is mainly due to computing the consensus loss for the different number of transformations. The current implementation computes the forward pass for the different transformations sequentially thus increasing the runtime. However, a more time efficient implementation where the forward passes for all the different random transformations are computed in parallel can make the runtime more comparable to the baseline ID algorithm.
}

\section{Conclusion}

In this work, we introduce different notions of the consistency constraints that are enforced in different unsupervised/self-supervised learning algorithms. We propose a novel clustering algorithm that seamlessly incorporates all three consistency constraints (exemplar, population and consensus) and achieves state-of-the-art clustering results for four out of five popular and challenging computer vision datasets. Our work on consensus clustering is significantly different from earlier consensus clustering works that do not learn representations. Moreover, we initiate a discussion on the adequacy of the currently used methods for evaluating clustering algorithms. 
We significantly extend the evaluation procedure for clustering algorithms, thereby reflecting the challenges of applying clustering to real-world tasks. We provide evaluation results for \pap\ and other state-of-the-art clustering algorithms based on
max-performance criteria, according to which ConCURL outperforms other algorithms on most datasets. However, its average performance according to out-of-distribution criteria highlights the need to use the proposed evaluation methods for deep clustering algorithms. 

\backmatter

\begin{appendices}

\section{Pseudo-code}
\label{sec:pseudo}
\begin{algorithm}[!htbp]
\caption{PyTorch-style pseudocode for \pap}
\label{alg:concurl}
\definecolor{codeblue}{rgb}{0.0,0.0,1.0}
\definecolor{codeblack}{rgb}{0.0,0.0,0.0}
\lstset{
    basicstyle=\fontsize{9pt}{9pt}\ttfamily\bfseries\color{codeblack},
    commentstyle=\fontsize{9pt}{9pt}\color{codeblue},
    keywordstyle=
}
\begin{lstlisting}[mathescape,escapechar=\%,language=python]
# f = feature network (e.g., ResNet18)
# g = MLP network that outputs embeddings
# c = prototypes - initialized randomly
# %\color{codeblue}$\tau$ % = temperature parameter
# list_A = random matrices for transformations

def ConsensusLoss(list_A, c, z1, z2, q1, q2):
    # Transform the embeddings and prototypes
    for counter, A in enumerate(list_A):  
        $\mathbf{\tilde{z}1}$= normalize(torch.mm(z1,A), p=2)
        $\mathbf{\tilde{z}2}$ = normalize(torch.mm(z2,A), p=2)
        with torch.no_grad():
            $\mathbf{\tilde{c}}$ = normalize(torch.mm(c,A), p=2)
        $\mathbf{\tilde{p}1}$= Softmax($\mathbf{\tilde{z}1}$,  $\mathbf{\tilde{c}}$), $\mathbf{\tilde{p}2}$= Softmax($\mathbf{\tilde{z}2}$,  $\mathbf{\tilde{c}}$)
        # CE = cross-entropy loss
        consensus_loss = CE(q2,$\mathbf{\tilde{p}1}$) + CE(q1,$\mathbf{\tilde{p}2}$)   

# training loop
for x, index in dataloader: 
    # Return embeddings from the model
    x1 = random_transformation(x), x2 = random_transformation(x) 
    # embeddings (ID loss)
    fx1 = f.forward(x1), fx2 = f.forward(x2) 
    # embeddings (clustering)
    z1 = g.forward(fx1), z2 = g.forward(fx2) 

    # Compute ID loss
    ID_loss = ID(normalized(fx1),index) # Using equation %\color{red}\ref{eqn:backbone_loss} %
    
    # Compute p and q vectors
    with torch.no_grad():
        c = normalize(c,p=2) 
    out1 = torch.mm(c,normalize(z1,p=2)
    out2 = torch.mm(c,normalize(z2,p=2)
    q1 = Sinkhorn(out1) # Using equation %\color{codeblue}\ref{eqn:sinkhorn} %
    q2 = Sinkhorn(out2) # Using equation %\color{codeblue}\ref{eqn:sinkhorn} %
    consensus_loss = ConsensusLoss(list_A, c, z1, z2, q1, q2)
    
    loss = %\color{codeblack}$\alpha$%*ID_loss +  %\color{codeblack}$\beta$%*consensus_loss
    optimizer.zero_grad(); loss.backward(); optimizer.step()
\end{lstlisting}
\end{algorithm}

\section{Evaluation Metrics}
\label{sec:evalmetrics}

We evaluate our algorithm by computing traditional clustering metrics (the ACC, NMI, and ARI), which we discuss below in detail.

\subsection{ACC} The ACC is computed by first computing a cluster partition of the input data. Once the partitions are computed and cluster indices are assigned to each input data point, a linear assignment map is computed using the Kuhn-Munkres (Hungarian) algorithm, which reassigns the cluster indices to the true labels of the data. The ACC is then given by
\[
ACC = \frac{ \sum_{i = 1}^N \mathbb{I}\{y_{true}(x_i) = c(x_i)\}}{N},
\]
where $y_{true}(x_i) $ is a true label of $x_i$ and $c(x_i)$ is the cluster assignment produced by an algorithm (after Hungarian mapping).

\subsection{NMI} For two clusters $U,V$, where each contains $|U|, |V|$ clusters and $|U_i|$ represents the number of samples in cluster $U_i$ of clustering result $U$ (similar for $V$), the MI is given by
\[
MI(U,V) = \sum_{i=1}^{\vert U \vert }\sum_{j=1}^{\vert V \vert } \frac{\vert U_i \cap V_i \vert }{N}\log \frac{N \vert U_i\cap V_j \vert }{\vert U_i\vert \vert V_j \vert}
\]
where $N$ is the number of data points under consideration. The NMI is defined as
\[
NMI(U,V) = \frac{MI(U,V)}{\sqrt{MI(U,U)MI(V,V)}}
\]

\subsection{ARI}\footnote{https://scikit-learn.org/stable/modules/clustering.html\#adjusted-rand-score} 
Suppose that R is the ground truth clustering result and that S is a partition. The RI of S is given as follows. Let $a$ be the number of pairs of elements that are in the same set in R and in S; let $b$ be the number of pairs of elements that are in different sets in R and in S. Then,
\[
RI = \frac{a + b}{{n \choose 2}}
\]
\[
ARI = \frac{RI - \mathbb{E}[RI] }{\max(RI) -  \mathbb{E}[RI] }
\]

\section{Implementation Details}
\label{sec:implementation}
In this section, we discuss the implementation of the proposed algorithm. We use PyTorch version 1.7.1 for the implementation. The ID block of the algorithm uses the code from the implementation of ID \citep{wu2018unsupervised} available at 
 \href{https://github.com/zhirongw/lemniscate.pytorch}{https://github.com/zhirongw/lemniscate.pytorch}. The repository uses PyTorch version 0.3, and appropriate changes are made to use it with the latest version of PyTorch. We use ResNet-18 and ResNet-50 blocks during our experiments. 
 In the ID block, the ResNet architecture is modified as follows. The final fully connected layer consists of 128 dimensions instead of the usual 1000 dimensions. The output of the final fully connected layer proceeds to compute the noise contrastive estimation (NCE) loss, and the feature representations (the layer before the fully connected layer) are fed to the clustering part.

 For the clustering part, the MLP projection head $g$ consists of a hidden layer of size 2048, followed by batch normalization and rectified linear unit (ReLU) layers, and an output layer of size 256. The prototypes are thus chosen to have 256 dimensions. 
Note that we fix the number of prototypes to be equal to the number of ground truth classes in the dataset. It has been shown, however, that overclustering leads to better representations \citep{caron2020unsupervised,ji2019invariant,asano2019self}, and we can extend our model to include an overclustering block with a larger set of prototypes \citep{ji2019invariant} and alternate the training procedure between the blocks.

We train the algorithm for 2000 epochs on all datasets. We use the SGD optimizer with a learning rate decay of 0.1 at prespecified epochs (600, 950, 1300, 1650, 2000) to perform the updates for all datasets. We perform a coarse learning rate search and find that 0.03 is the best-performing setting. We use a batch size of 128 for all the datasets. To evaluate the cluster accuracy, we compute the cluster assignments using {MiniBatchKMeans}\footnote{https://scikit-learn.org/stable/modules/clustering.html\#mini-batch-kmeans} with a batch size of 6000 and 20 random initializations.

\subsection{Image augmentations}
The different views $\mathcal{X}_b^1, \mathcal{X}_b^2$ are not the same as the views in multiview datasets \citep{schops2017multi}. The views referred to in this paper correspond to different augmented views that are generated by image augmentation techniques, such as RandomHorizontalFlip and RandomCrop.
We explain the generation process of multiple augmented views, which have been shown to be very effective in unsupervised learning \citep{chen2020simple}. Indeed, it is possible to use more than two augmented views, but we limit to the number to two for the sake of simplicity. \cite{caron2020unsupervised} proposed an augmentation technique (Multi-Crop) to use more than two views. In this work, we use the augmentation methods used in \cite{chen2020simple,grill2020bootstrap}.
We first crop a random patch of the input image with a scale ranging from 0.08 to 1.0 and resize the cropped patch to 224$\times$224 (96$\times$96 for smaller-resolution datasets such as STL10). The resulting image is then flipped horizontally with a probability of 0.5. We then apply color transformations, starting by applying grayscale with a probability of 0.2 followed by randomly changing the brightness, contrast, saturation and hue with a probability of 0.8. Then, we apply a Gaussian blur with a kernel size of 23$\times$23 and a sigma chosen uniformly and randomly between 0.1 and 2.0. The probabilities of applying Gaussian blur are 1.0 for view 1 and 0.5 for view 2. During the evaluation, we resize the image such that the smaller edge of the image is of size 256 (not required for \stl, \cten, and \chundred), and a center crop operation is performed with the resolution mentioned in the main paper. 
We finally normalize the image channels with the mean and standard deviation computed on ImageNet. Additionally, during training, we experiment with applying a Sobel filter after all the image augmentation steps are performed but before the forward pass. Applying a Sobel filter reduces the number of channels in the input images to 2. We also experiment by augmenting the RGB images with the output of the Sobel transform, resulting in 5-channel input images. In both of these cases, the input channels in the first convolution layer are modified accordingly. All image augmentations are computed using PyTorch's torchvision module (available in version 1.7.1).

\subsection{Random transformations}
To compute the random transformations on the embeddings $\mathbf{z}$, we follow two techniques. We use Gaussian random projections with an output dimensionality of $d$ and transform the embeddings $\mathbf{z}$ to the new space with a dimensionality of $d$. In Gaussian random projections, the projection matrix is generated by picking rows from a Gaussian distribution such that they are orthogonal. We also use diagonal transformation \citep{hsu2018unsupervised}, where we multiply $\mathbf{z}$ with a randomly generated diagonal matrix with the same dimensions as $\mathbf{z}$. We initialize $M$ random transformations at the beginning, and they are kept fixed throughout the training process.

\section{Comparison with LA}
\label{sec:local_aggr}

LA \citep{zhuang2019local} builds on nonparametric ID \citep{wu2018unsupervised} and uses a robust clustering objective (it uses a closest neighbors set generated using multiple runs of k-means) similar to consensus clustering to move statistically similar points closer in the representation space and dissimilar points farther away. By using the linear evaluation protocol on ImageNet, the authors demonstrate that the representations learned with LA are better than those obtained without LA. 

However, the performance of these features with respect to clustering was not discussed. Since LA is similar to our work in spirit, we perform a study on the clustering performance of the features learned using LA\footnote{using the official PyTorch implementation available at https://github.com/neuroailab/LocalAggregation-PyTorch/tree/master/config} on the ImageNet-10 and ImageNet-Dogs datasets. The results are presented in Table~\ref{tab:comparison_la_im}.

\begin{table}[htb!]
\begin{center}
\caption{Comparison with LA on ImageNet-10 } 
\label{tab:comparison_la_im}
\begin{tabular}{|c| ccc | ccc|}
    \hline
     Datasets & 
     \multicolumn{3}{c}{\iten} & \multicolumn{3}{c}{\idogs}
     \vline\\
    \hline
    Method\textbackslash Metrics & 
    Acc & NMI & ARI & Acc & NMI & ARI
    \\
    \hline
    LA &
        0.393 & 0.346 & 0.213 & 0.201 & 0.133 & 0.06
        \\
    \hline
    \pap &
        \textbf{0.958}&\textbf{0.907}&\textbf{0.909}&\textbf{0.695}&\textbf{0.63}&\textbf{0.531}\\
    \hline

\end{tabular}%
\end{center}
\end{table}

We observe that the clustering performance of our proposed ConCURL algorithm is much better than the clustering performance of the LA method. Note that the clustering performance of the ID features is better than that of LA, and our algorithm further improves upon the clustering performance of ID. One major difference between our work and LA is the way in which we generate the ensemble. Our approach allows us to control and measure the diversity of the ensemble, which can be useful in making algorithm design choices. Although LA controls the ensemble by varying
the number of clustering results and
the number of clusters in each clustering result, which aptly suits the objective that LA is solving, the resultant ensembles are limited to utilizing k-means clustering (the authors showed that other clustering approaches were either not scalable or not optimal). In our case, by applying feature space transformations, we have much more freedom in generating the ensemble. We use random projections and diagonal transformations, but there could be other transformations on the feature space that we have not yet explored.

\subsection{Implementation Details of LA}
We train the model for 500 epochs. Since the original implementation of LA was designed for ImageNet, we perform a hyperparameter search as follows. Using the config file from the official PyTorch implementation, we create 36 configuration files by varying the learning rate and k-means-k. In the original config file provided by the authors, k-means-k = 30000 (for 1.28 million images). We scale the k-means-k for ImageNet-10 (13000 images) and ImageNet-Dogs (19500 images) accordingly and try six different values. In particular, we try learning rates = [0.003, 0.005, 0.01, 0.03, 0.05, 0.1], and k-means-k = [10, 15, 100, 310, 452, 500].

The number of background neighbors is 4096, as used in the original paper. We run ResNet-18 experiments for the full hyperparameter search (36 experiments) and evaluate the cluster metrics.
Additionally, we choose the top 5 choices of hyperparameters from above and run ResNet-34 experiments with those parameters for both datasets. In Table~\ref{tab:comparison_la_im}, we present the best results obtained for each dataset.

The code repository uses a different version of ResNet (PreActResNet). Therefore, for evaluating the clustering performance, we take the output of the layer before the final dense layer. For ResNet-18, the output dimensions of this layer are (512,7,7), and we take the mean along the (1,2) dimensions and use the resulting 512-dimensional vector. We compute the k-means clustering results on these embeddings using faiss (\href{https://github.com/facebookresearch/faiss}{https://github.com/facebookresearch/faiss}) for the training split of the data and compute the cluster metrics as mentioned in our paper.
\end{appendices}


\section*{Declarations}

\begin{itemize}
\item Funding - Work was done at Microsoft with Microsoft's support. 
\item Conflict of interests - Authors are from Microsoft and The Ohio State University. There are no conflict of interests to disclose. 
\item Ethics approval - Not applicable 
\item Consent to participate - Yes
\item Consent for publication - Yes
\item Availability of data and materials - The datasets used are available at \href{CIFAR-10, CIFAR-100}{https://www.cs.toronto.edu/~kriz/cifar.html}, \href{STL10}{https://cs.stanford.edu/~acoates/stl10/} or can be requested at \href{ImageNet-10, ImageNet-Dogs}{https://image-net.org/}. All the trained models, their usage is available \href{here}{https://github.com/JayanthRR/ConCURL_NCE}.
\item Code availability - Code is available \href{here}{https://github.com/JayanthRR/ConCURL_NCE}. 
\item Authors' contributions - All authors - Aniket Anand Deshmukh, Jayanth Reddy Regatti, Eren Manavoglu and Urun Dogan contributed to this work and were essential to complete this submission. 
\end{itemize}

\bibliography{ConCURL}


\end{document}

%% file: math_commands.tex
\def\eqref#1{equation~\ref{#1}}

\def\1{\bm{1}}

\DeclareMathAlphabet{\mathsfit}{\encodingdefault}{\sfdefault}{m}{sl}
\SetMathAlphabet{\mathsfit}{bold}{\encodingdefault}{\sfdefault}{bx}{n}

\DeclareMathOperator*{\argmax}{arg\,max}

%% file: main_springer.bbl
\begin{thebibliography}{}
\providecommand{\doi}[1]{\url{https://doi.org/#1}}
\bibcommenthead

\bibitem [\protect \citeauthoryear {%
Asano%
, Rupprecht%
\BCBL {}\ \BBA {} Vedaldi%
}{%
Asano%
\ \protect \BOthers {.}}{%
{\protect \APACyear {2019}}%
}]{%
asano2019self}
\APACinsertmetastar {%
asano2019self}%
\begin{APACrefauthors}%
Asano, Y.M.%
, Rupprecht, C.%
\BCBL {} Vedaldi, A.%
\end{APACrefauthors}%
\unskip\
\newblock
\APACrefYearMonthDay{2019}{}{}.
\newblock
{\BBOQ}\APACrefatitle {Self-labelling via simultaneous clustering and
  representation learning} {Self-labelling via simultaneous clustering and
  representation learning}.{\BBCQ}
\newblock
\APACjournalVolNumPages{arXiv preprint arXiv:1911.05371}{}{}{}.
\newblock

\newblock

\PrintBackRefs{\CurrentBib}

\bibitem [\protect \citeauthoryear {%
Bengio%
, Lamblin%
, Popovici%
, Larochelle%
\BCBL {}\ \BBA {} Montreal%
}{%
Bengio%
\ \protect \BOthers {.}}{%
{\protect \APACyear {2007}}%
}]{%
Bengio2007Greedy}
\APACinsertmetastar {%
Bengio2007Greedy}%
\begin{APACrefauthors}%
Bengio, Y.%
, Lamblin, P.%
, Popovici, D.%
, Larochelle, H.%
\BCBL {} Montreal, U.%
\end{APACrefauthors}%
\unskip\
\newblock
\APACrefYearMonthDay{2007}{}{}.
\newblock
{\BBOQ}\APACrefatitle {Greedy layer-wise training of deep networks} {Greedy
  layer-wise training of deep networks}.{\BBCQ}
\newblock
\APACjournalVolNumPages{NeurIPS}{19}{}{153-160}.
\newblock

\newblock

\PrintBackRefs{\CurrentBib}

\bibitem [\protect \citeauthoryear {%
Cai%
, He%
, Wang%
, Bao%
\BCBL {}\ \BBA {} Han%
}{%
Cai%
\ \protect \BOthers {.}}{%
{\protect \APACyear {2009}}%
}]{%
NMF}
\APACinsertmetastar {%
NMF}%
\begin{APACrefauthors}%
Cai, D.%
, He, X.%
, Wang, X.%
, Bao, H.%
\BCBL {} Han, J.%
\end{APACrefauthors}%
\unskip\
\newblock
\APACrefYearMonthDay{2009}{}{}.
\newblock
{\BBOQ}\APACrefatitle {Locality Preserving Nonnegative Matrix Factorization}
  {Locality preserving nonnegative matrix factorization}.{\BBCQ}
\newblock
 \APACrefbtitle {IJCAI} {Ijcai}\ (\BPGS\ 1010--1015).
\newblock
\begin{APACrefURL} {http://ijcai.org/Proceedings/09/Papers/171.pdf}
  \end{APACrefURL}
\PrintBackRefs{\CurrentBib}

\bibitem [\protect \citeauthoryear {%
Caron%
, Bojanowski%
, Joulin%
\BCBL {}\ \BBA {} Douze%
}{%
Caron%
\ \protect \BOthers {.}}{%
{\protect \APACyear {2018}}%
{\protect \APACexlab {{\protect \BCnt {1}}}}}]{%
caron2018deep}
\APACinsertmetastar {%
caron2018deep}%
\begin{APACrefauthors}%
Caron, M.%
, Bojanowski, P.%
, Joulin, A.%
\BCBL {} Douze, M.%
\end{APACrefauthors}%
\unskip\
\newblock
\APACrefYearMonthDay{2018{\protect \BCnt {1}}}{}{}.
\newblock
{\BBOQ}\APACrefatitle {Deep clustering for unsupervised learning of visual
  features} {Deep clustering for unsupervised learning of visual
  features}.{\BBCQ}
\newblock
 \APACrefbtitle {Proceedings of the European Conference on Computer Vision
  (ECCV)} {Proceedings of the european conference on computer vision (eccv)}\
  (\BPGS\ 132--149).
\PrintBackRefs{\CurrentBib}

\bibitem [\protect \citeauthoryear {%
Caron%
, Bojanowski%
, Joulin%
\BCBL {}\ \BBA {} Douze%
}{%
Caron%
\ \protect \BOthers {.}}{%
{\protect \APACyear {2018}}%
{\protect \APACexlab {{\protect \BCnt {2}}}}}]{%
vf2018}
\APACinsertmetastar {%
vf2018}%
\begin{APACrefauthors}%
Caron, M.%
, Bojanowski, P.%
, Joulin, A.%
\BCBL {} Douze, M.%
\end{APACrefauthors}%
\unskip\
\newblock
\APACrefYearMonthDay{2018{\protect \BCnt {2}}}{}{}.
\newblock
{\BBOQ}\APACrefatitle {Deep Clustering for Unsupervised Learning of Visual
  Features.} {Deep clustering for unsupervised learning of visual
  features.}{\BBCQ}
\newblock
 \APACrefbtitle {ECCV} {Eccv}\ (\BVOL\ 11218, \BPG~139-156).
\PrintBackRefs{\CurrentBib}

\bibitem [\protect \citeauthoryear {%
Caron%
\ \protect \BOthers {.}}{%
Caron%
\ \protect \BOthers {.}}{%
{\protect \APACyear {2020}}%
}]{%
caron2020unsupervised}
\APACinsertmetastar {%
caron2020unsupervised}%
\begin{APACrefauthors}%
Caron, M.%
, Misra, I.%
, Mairal, J.%
, Goyal, P.%
, Bojanowski, P.%
\BCBL {} Joulin, A.%
\end{APACrefauthors}%
\unskip\
\newblock
\APACrefYearMonthDay{2020}{}{}.
\newblock
{\BBOQ}\APACrefatitle {Unsupervised learning of visual features by contrasting
  cluster assignments} {Unsupervised learning of visual features by contrasting
  cluster assignments}.{\BBCQ}
\newblock
\APACjournalVolNumPages{arXiv preprint arXiv:2006.09882}{}{}{}.
\newblock

\newblock

\PrintBackRefs{\CurrentBib}

\bibitem [\protect \citeauthoryear {%
Chang%
\ \protect \BOthers {.}}{%
Chang%
\ \protect \BOthers {.}}{%
{\protect \APACyear {2019}}%
}]{%
chang2019deep}
\APACinsertmetastar {%
chang2019deep}%
\begin{APACrefauthors}%
Chang, J.%
, Guo, Y.%
, Wang, L.%
, Meng, G.%
, Xiang, S.%
\BCBL {} Pan, C.%
\end{APACrefauthors}%
\unskip\
\newblock
\APACrefYearMonthDay{2019}{}{}.
\newblock
\APACrefbtitle {Deep Discriminative Clustering Analysis.} {Deep discriminative
  clustering analysis.}
\PrintBackRefs{\CurrentBib}

\bibitem [\protect \citeauthoryear {%
Chang%
, Wang%
, Meng%
, Xiang%
\BCBL {}\ \BBA {} Pan%
}{%
Chang%
\ \protect \BOthers {.}}{%
{\protect \APACyear {2017}}%
}]{%
Chang_2017_ICCV}
\APACinsertmetastar {%
Chang_2017_ICCV}%
\begin{APACrefauthors}%
Chang, J.%
, Wang, L.%
, Meng, G.%
, Xiang, S.%
\BCBL {} Pan, C.%
\end{APACrefauthors}%
\unskip\
\newblock
\APACrefYearMonthDay{2017}{Oct}{}.
\newblock
{\BBOQ}\APACrefatitle {Deep Adaptive Image Clustering} {Deep adaptive image
  clustering}.{\BBCQ}
\newblock
 \APACrefbtitle {The IEEE International Conference on Computer Vision (ICCV).}
  {The ieee international conference on computer vision (iccv).}
\PrintBackRefs{\CurrentBib}

\bibitem [\protect \citeauthoryear {%
{Chang}%
, {Wang}%
, {Meng}%
, {Xiang}%
\BCBL {}\ \BBA {} {Pan}%
}{%
{Chang}%
\ \protect \BOthers {.}}{%
{\protect \APACyear {2017}}%
}]{%
DAIC2017}
\APACinsertmetastar {%
DAIC2017}%
\begin{APACrefauthors}%
{Chang}, J.%
, {Wang}, L.%
, {Meng}, G.%
, {Xiang}, S.%
\BCBL {} {Pan}, C.%
\end{APACrefauthors}%
\unskip\
\newblock
\APACrefYearMonthDay{2017}{Oct}{}.
\newblock
{\BBOQ}\APACrefatitle {Deep Adaptive Image Clustering} {Deep adaptive image
  clustering}.{\BBCQ}
\newblock
 \APACrefbtitle {ICCV} {Iccv}\ (\BPG~5880-5888).
\newblock
\begin{APACrefDOI} \doi{10.1109/ICCV.2017.626} \end{APACrefDOI}
\PrintBackRefs{\CurrentBib}

\bibitem [\protect \citeauthoryear {%
Chen%
, Kornblith%
, Norouzi%
\BCBL {}\ \BBA {} Hinton%
}{%
Chen%
\ \protect \BOthers {.}}{%
{\protect \APACyear {2020}}%
}]{%
chen2020simple}
\APACinsertmetastar {%
chen2020simple}%
\begin{APACrefauthors}%
Chen, T.%
, Kornblith, S.%
, Norouzi, M.%
\BCBL {} Hinton, G.%
\end{APACrefauthors}%
\unskip\
\newblock
\APACrefYearMonthDay{2020}{}{}.
\newblock
{\BBOQ}\APACrefatitle {A simple framework for contrastive learning of visual
  representations} {A simple framework for contrastive learning of visual
  representations}.{\BBCQ}
\newblock
\APACjournalVolNumPages{arXiv preprint arXiv:2002.05709}{}{}{}.
\newblock

\newblock

\PrintBackRefs{\CurrentBib}

\bibitem [\protect \citeauthoryear {%
Cuturi%
}{%
Cuturi%
}{%
{\protect \APACyear {2013}}%
}]{%
cuturi2013sinkhorn}
\APACinsertmetastar {%
cuturi2013sinkhorn}%
\begin{APACrefauthors}%
Cuturi, M.%
\end{APACrefauthors}%
\unskip\
\newblock
\APACrefYearMonthDay{2013}{}{}.
\newblock
{\BBOQ}\APACrefatitle {Sinkhorn distances: Lightspeed computation of optimal
  transport} {Sinkhorn distances: Lightspeed computation of optimal
  transport}.{\BBCQ}
\newblock
 \APACrefbtitle {Advances in neural information processing systems} {Advances
  in neural information processing systems}\ (\BPGS\ 2292--2300).
\PrintBackRefs{\CurrentBib}

\bibitem [\protect \citeauthoryear {%
Deng%
\ \protect \BOthers {.}}{%
Deng%
\ \protect \BOthers {.}}{%
{\protect \APACyear {2009}}%
}]{%
deng2009imagenet}
\APACinsertmetastar {%
deng2009imagenet}%
\begin{APACrefauthors}%
Deng, J.%
, Dong, W.%
, Socher, R.%
, Li, L\BHBI J.%
, Li, K.%
\BCBL {} Fei-Fei, L.%
\end{APACrefauthors}%
\unskip\
\newblock
\APACrefYearMonthDay{2009}{}{}.
\newblock
{\BBOQ}\APACrefatitle {Imagenet: A large-scale hierarchical image database}
  {Imagenet: A large-scale hierarchical image database}.{\BBCQ}
\newblock
 \APACrefbtitle {2009 IEEE conference on computer vision and pattern
  recognition} {2009 ieee conference on computer vision and pattern
  recognition}\ (\BPGS\ 248--255).
\PrintBackRefs{\CurrentBib}

\bibitem [\protect \citeauthoryear {%
Fern%
\ \BBA {} Brodley%
}{%
Fern%
\ \BBA {} Brodley%
}{%
{\protect \APACyear {2003}}%
}]{%
fern2003random}
\APACinsertmetastar {%
fern2003random}%
\begin{APACrefauthors}%
Fern, X.Z.%
\BCBT {}\ \BBA {} Brodley, C.E.%
\end{APACrefauthors}%
\unskip\
\newblock
\APACrefYearMonthDay{2003}{}{}.
\newblock
{\BBOQ}\APACrefatitle {Random projection for high dimensional data clustering:
  A cluster ensemble approach} {Random projection for high dimensional data
  clustering: A cluster ensemble approach}.{\BBCQ}
\newblock
 \APACrefbtitle {Proceedings of the 20th international conference on machine
  learning (ICML-03)} {Proceedings of the 20th international conference on
  machine learning (icml-03)}\ (\BPGS\ 186--193).
\PrintBackRefs{\CurrentBib}

\bibitem [\protect \citeauthoryear {%
{Franti}%
, {Virmajoki}%
\BCBL {}\ \BBA {} {Hautamaki}%
}{%
{Franti}%
\ \protect \BOthers {.}}{%
{\protect \APACyear {2006}}%
}]{%
Pasi2006Fast}
\APACinsertmetastar {%
Pasi2006Fast}%
\begin{APACrefauthors}%
{Franti}, P.%
, {Virmajoki}, O.%
\BCBL {} {Hautamaki}, V.%
\end{APACrefauthors}%
\unskip\
\newblock
\APACrefYearMonthDay{2006}{Nov}{}.
\newblock
{\BBOQ}\APACrefatitle {Fast Agglomerative Clustering Using a k-Nearest Neighbor
  Graph} {Fast agglomerative clustering using a k-nearest neighbor
  graph}.{\BBCQ}
\newblock
\APACjournalVolNumPages{IEEE Transactions on Pattern Analysis and Machine
  Intelligence}{28}{11}{1875-1881}.
\newblock

\newblock

\PrintBackRefs{\CurrentBib}

\bibitem [\protect \citeauthoryear {%
Fred%
\ \BBA {} Jain%
}{%
Fred%
\ \BBA {} Jain%
}{%
{\protect \APACyear {2005}}%
}]{%
fred2005combining}
\APACinsertmetastar {%
fred2005combining}%
\begin{APACrefauthors}%
Fred, A.L.%
\BCBT {}\ \BBA {} Jain, A.K.%
\end{APACrefauthors}%
\unskip\
\newblock
\APACrefYearMonthDay{2005}{}{}.
\newblock
{\BBOQ}\APACrefatitle {Combining multiple clusterings using evidence
  accumulation} {Combining multiple clusterings using evidence
  accumulation}.{\BBCQ}
\newblock
\APACjournalVolNumPages{IEEE transactions on pattern analysis and machine
  intelligence}{27}{6}{835--850}.
\newblock

\newblock

\PrintBackRefs{\CurrentBib}

\bibitem [\protect \citeauthoryear {%
Frey%
\ \BBA {} Dueck%
}{%
Frey%
\ \BBA {} Dueck%
}{%
{\protect \APACyear {2007}}%
}]{%
frey2007clustering}
\APACinsertmetastar {%
frey2007clustering}%
\begin{APACrefauthors}%
Frey, B.J.%
\BCBT {}\ \BBA {} Dueck, D.%
\end{APACrefauthors}%
\unskip\
\newblock
\APACrefYearMonthDay{2007}{}{}.
\newblock
{\BBOQ}\APACrefatitle {Clustering by passing messages between data points}
  {Clustering by passing messages between data points}.{\BBCQ}
\newblock
\APACjournalVolNumPages{science}{315}{5814}{972--976}.
\newblock

\newblock

\PrintBackRefs{\CurrentBib}

\bibitem [\protect \citeauthoryear {%
Ghosh%
\ \BBA {} Acharya%
}{%
Ghosh%
\ \BBA {} Acharya%
}{%
{\protect \APACyear {2011}}%
}]{%
ghosh2011cluster}
\APACinsertmetastar {%
ghosh2011cluster}%
\begin{APACrefauthors}%
Ghosh, J.%
\BCBT {}\ \BBA {} Acharya, A.%
\end{APACrefauthors}%
\unskip\
\newblock
\APACrefYearMonthDay{2011}{}{}.
\newblock
{\BBOQ}\APACrefatitle {Cluster ensembles} {Cluster ensembles}.{\BBCQ}
\newblock
\APACjournalVolNumPages{Wiley Interdisciplinary Reviews: Data Mining and
  Knowledge Discovery}{1}{4}{305--315}.
\newblock

\newblock

\PrintBackRefs{\CurrentBib}

\bibitem [\protect \citeauthoryear {%
Grill%
\ \protect \BOthers {.}}{%
Grill%
\ \protect \BOthers {.}}{%
{\protect \APACyear {2020}}%
}]{%
grill2020bootstrap}
\APACinsertmetastar {%
grill2020bootstrap}%
\begin{APACrefauthors}%
Grill, J\BHBI B.%
, Strub, F.%
, Altch{\'e}, F.%
, Tallec, C.%
, Richemond, P.H.%
, Buchatskaya, E.%
\BDBL {}others%
\end{APACrefauthors}%
\unskip\
\newblock
\APACrefYearMonthDay{2020}{}{}.
\newblock
{\BBOQ}\APACrefatitle {Bootstrap your own latent: A new approach to
  self-supervised learning} {Bootstrap your own latent: A new approach to
  self-supervised learning}.{\BBCQ}
\newblock
\APACjournalVolNumPages{arXiv preprint arXiv:2006.07733}{}{}{}.
\newblock

\newblock

\PrintBackRefs{\CurrentBib}

\bibitem [\protect \citeauthoryear {%
Haeusser%
, Plapp%
, Golkov%
, Aljalbout%
\BCBL {}\ \BBA {} Cremers%
}{%
Haeusser%
\ \protect \BOthers {.}}{%
{\protect \APACyear {2019}}%
}]{%
ADC2019}
\APACinsertmetastar {%
ADC2019}%
\begin{APACrefauthors}%
Haeusser, P.%
, Plapp, J.%
, Golkov, V.%
, Aljalbout, E.%
\BCBL {} Cremers, D.%
\end{APACrefauthors}%
\unskip\
\newblock
\APACrefYearMonthDay{2019}{}{}.
\newblock
{\BBOQ}\APACrefatitle {Associative Deep Clustering: Training a Classification
  Network with No Labels} {Associative deep clustering: Training a
  classification network with no labels}.{\BBCQ}
\newblock
 T.~Brox, A.~Bruhn\BCBL {}\ \BBA {} M.~Fritz\ (\BEDS), \APACrefbtitle {Pattern
  Recognition} {Pattern recognition}\ (\BPGS\ 18--32).
\newblock
\APACaddressPublisher{Cham}{Springer International Publishing}.
\PrintBackRefs{\CurrentBib}

\bibitem [\protect \citeauthoryear {%
He%
, Fan%
, Wu%
, Xie%
\BCBL {}\ \BBA {} Girshick%
}{%
He%
\ \protect \BOthers {.}}{%
{\protect \APACyear {2019}}%
}]{%
he2019moco}
\APACinsertmetastar {%
he2019moco}%
\begin{APACrefauthors}%
He, K.%
, Fan, H.%
, Wu, Y.%
, Xie, S.%
\BCBL {} Girshick, R.%
\end{APACrefauthors}%
\unskip\
\newblock
\APACrefYearMonthDay{2019}{}{}.
\newblock
{\BBOQ}\APACrefatitle {Momentum Contrast for Unsupervised Visual Representation
  Learning} {Momentum contrast for unsupervised visual representation
  learning}.{\BBCQ}
\newblock
\APACjournalVolNumPages{arXiv preprint arXiv:1911.05722}{}{}{}.
\newblock

\newblock

\PrintBackRefs{\CurrentBib}

\bibitem [\protect \citeauthoryear {%
He%
, Fan%
, Wu%
, Xie%
\BCBL {}\ \BBA {} Girshick%
}{%
He%
\ \protect \BOthers {.}}{%
{\protect \APACyear {2020}}%
}]{%
he2020momentum}
\APACinsertmetastar {%
he2020momentum}%
\begin{APACrefauthors}%
He, K.%
, Fan, H.%
, Wu, Y.%
, Xie, S.%
\BCBL {} Girshick, R.%
\end{APACrefauthors}%
\unskip\
\newblock
\APACrefYearMonthDay{2020}{}{}.
\newblock
{\BBOQ}\APACrefatitle {Momentum contrast for unsupervised visual representation
  learning} {Momentum contrast for unsupervised visual representation
  learning}.{\BBCQ}
\newblock
 \APACrefbtitle {Proceedings of the IEEE/CVF Conference on Computer Vision and
  Pattern Recognition} {Proceedings of the ieee/cvf conference on computer
  vision and pattern recognition}\ (\BPGS\ 9729--9738).
\PrintBackRefs{\CurrentBib}

\bibitem [\protect \citeauthoryear {%
Hsu%
, Levine%
\BCBL {}\ \BBA {} Finn%
}{%
Hsu%
\ \protect \BOthers {.}}{%
{\protect \APACyear {2018}}%
}]{%
hsu2018unsupervised}
\APACinsertmetastar {%
hsu2018unsupervised}%
\begin{APACrefauthors}%
Hsu, K.%
, Levine, S.%
\BCBL {} Finn, C.%
\end{APACrefauthors}%
\unskip\
\newblock
\APACrefYearMonthDay{2018}{}{}.
\newblock
{\BBOQ}\APACrefatitle {Unsupervised learning via meta-learning} {Unsupervised
  learning via meta-learning}.{\BBCQ}
\newblock
\APACjournalVolNumPages{arXiv preprint arXiv:1810.02334}{}{}{}.
\newblock

\newblock

\PrintBackRefs{\CurrentBib}

\bibitem [\protect \citeauthoryear {%
Huang%
, Gong%
\BCBL {}\ \BBA {} Zhu%
}{%
Huang%
\ \protect \BOthers {.}}{%
{\protect \APACyear {2020}}%
{\protect \APACexlab {{\protect \BCnt {1}}}}}]{%
huang2020deep}
\APACinsertmetastar {%
huang2020deep}%
\begin{APACrefauthors}%
Huang, J.%
, Gong, S.%
\BCBL {} Zhu, X.%
\end{APACrefauthors}%
\unskip\
\newblock
\APACrefYearMonthDay{2020{\protect \BCnt {1}}}{}{}.
\newblock
{\BBOQ}\APACrefatitle {Deep Semantic Clustering by Partition Confidence
  Maximisation} {Deep semantic clustering by partition confidence
  maximisation}.{\BBCQ}
\newblock
 \APACrefbtitle {Proceedings of the IEEE/CVF Conference on Computer Vision and
  Pattern Recognition} {Proceedings of the ieee/cvf conference on computer
  vision and pattern recognition}\ (\BPGS\ 8849--8858).
\PrintBackRefs{\CurrentBib}

\bibitem [\protect \citeauthoryear {%
Huang%
, Gong%
\BCBL {}\ \BBA {} Zhu%
}{%
Huang%
\ \protect \BOthers {.}}{%
{\protect \APACyear {2020}}%
{\protect \APACexlab {{\protect \BCnt {2}}}}}]{%
Huang_2020_CVPR}
\APACinsertmetastar {%
Huang_2020_CVPR}%
\begin{APACrefauthors}%
Huang, J.%
, Gong, S.%
\BCBL {} Zhu, X.%
\end{APACrefauthors}%
\unskip\
\newblock
\APACrefYearMonthDay{2020{\protect \BCnt {2}}}{June}{}.
\newblock
{\BBOQ}\APACrefatitle {Deep Semantic Clustering by Partition Confidence
  Maximisation} {Deep semantic clustering by partition confidence
  maximisation}.{\BBCQ}
\newblock
 \APACrefbtitle {CVPR.} {Cvpr.}
\PrintBackRefs{\CurrentBib}

\bibitem [\protect \citeauthoryear {%
Jain%
\ \BBA {} Dubes%
}{%
Jain%
\ \BBA {} Dubes%
}{%
{\protect \APACyear {1988}}%
}]{%
jain1988algorithms}
\APACinsertmetastar {%
jain1988algorithms}%
\begin{APACrefauthors}%
Jain, A.K.%
\BCBT {}\ \BBA {} Dubes, R.C.%
\end{APACrefauthors}%
\unskip\
\newblock
\APACrefYear{1988}.
\newblock
\APACrefbtitle {Algorithms for clustering data} {Algorithms for clustering
  data}.
\newblock
\APACaddressPublisher{}{Prentice-Hall, Inc.}
\PrintBackRefs{\CurrentBib}

\bibitem [\protect \citeauthoryear {%
Jain%
, Murty%
\BCBL {}\ \BBA {} Flynn%
}{%
Jain%
\ \protect \BOthers {.}}{%
{\protect \APACyear {1999}}%
}]{%
jain1999data}
\APACinsertmetastar {%
jain1999data}%
\begin{APACrefauthors}%
Jain, A.K.%
, Murty, M.N.%
\BCBL {} Flynn, P.J.%
\end{APACrefauthors}%
\unskip\
\newblock
\APACrefYearMonthDay{1999}{}{}.
\newblock
{\BBOQ}\APACrefatitle {Data clustering: a review} {Data clustering: a
  review}.{\BBCQ}
\newblock
\APACjournalVolNumPages{ACM computing surveys (CSUR)}{31}{3}{264--323}.
\newblock

\newblock

\PrintBackRefs{\CurrentBib}

\bibitem [\protect \citeauthoryear {%
Ji%
, Henriques%
\BCBL {}\ \BBA {} Vedaldi%
}{%
Ji%
\ \protect \BOthers {.}}{%
{\protect \APACyear {2019}}%
{\protect \APACexlab {{\protect \BCnt {1}}}}}]{%
ji2019invariant}
\APACinsertmetastar {%
ji2019invariant}%
\begin{APACrefauthors}%
Ji, X.%
, Henriques, J.F.%
\BCBL {} Vedaldi, A.%
\end{APACrefauthors}%
\unskip\
\newblock
\APACrefYearMonthDay{2019{\protect \BCnt {1}}}{}{}.
\newblock
{\BBOQ}\APACrefatitle {Invariant information clustering for unsupervised image
  classification and segmentation} {Invariant information clustering for
  unsupervised image classification and segmentation}.{\BBCQ}
\newblock
 \APACrefbtitle {Proceedings of the IEEE International Conference on Computer
  Vision} {Proceedings of the ieee international conference on computer
  vision}\ (\BPGS\ 9865--9874).
\PrintBackRefs{\CurrentBib}

\bibitem [\protect \citeauthoryear {%
Ji%
, Henriques%
\BCBL {}\ \BBA {} Vedaldi%
}{%
Ji%
\ \protect \BOthers {.}}{%
{\protect \APACyear {2019}}%
{\protect \APACexlab {{\protect \BCnt {2}}}}}]{%
IIC2019}
\APACinsertmetastar {%
IIC2019}%
\begin{APACrefauthors}%
Ji, X.%
, Henriques, J.F.%
\BCBL {} Vedaldi, A.%
\end{APACrefauthors}%
\unskip\
\newblock
\APACrefYearMonthDay{2019{\protect \BCnt {2}}}{October}{}.
\newblock
{\BBOQ}\APACrefatitle {Invariant Information Clustering for Unsupervised Image
  Classification and Segmentation} {Invariant information clustering for
  unsupervised image classification and segmentation}.{\BBCQ}
\newblock
 \APACrefbtitle {ICCV.} {Iccv.}
\PrintBackRefs{\CurrentBib}

\bibitem [\protect \citeauthoryear {%
Kuhn%
}{%
Kuhn%
}{%
{\protect \APACyear {1955}}%
}]{%
kuhn1955hungarian}
\APACinsertmetastar {%
kuhn1955hungarian}%
\begin{APACrefauthors}%
Kuhn, H.W.%
\end{APACrefauthors}%
\unskip\
\newblock
\APACrefYearMonthDay{1955}{}{}.
\newblock
{\BBOQ}\APACrefatitle {The Hungarian method for the assignment problem} {The
  hungarian method for the assignment problem}.{\BBCQ}
\newblock
\APACjournalVolNumPages{Naval research logistics quarterly}{2}{1-2}{83--97}.
\newblock

\newblock

\PrintBackRefs{\CurrentBib}

\bibitem [\protect \citeauthoryear {%
Kuhn%
}{%
Kuhn%
}{%
{\protect \APACyear {1956}}%
}]{%
kuhn1956variants}
\APACinsertmetastar {%
kuhn1956variants}%
\begin{APACrefauthors}%
Kuhn, H.W.%
\end{APACrefauthors}%
\unskip\
\newblock
\APACrefYearMonthDay{1956}{}{}.
\newblock
{\BBOQ}\APACrefatitle {Variants of the Hungarian method for assignment
  problems} {Variants of the hungarian method for assignment problems}.{\BBCQ}
\newblock
\APACjournalVolNumPages{Naval Research Logistics Quarterly}{3}{4}{253--258}.
\newblock

\newblock

\PrintBackRefs{\CurrentBib}

\bibitem [\protect \citeauthoryear {%
LeCun%
, Bengio%
\BCBL {}\ \BBA {} Hinton%
}{%
LeCun%
\ \protect \BOthers {.}}{%
{\protect \APACyear {2015}}%
}]{%
lecun2015deep}
\APACinsertmetastar {%
lecun2015deep}%
\begin{APACrefauthors}%
LeCun, Y.%
, Bengio, Y.%
\BCBL {} Hinton, G.%
\end{APACrefauthors}%
\unskip\
\newblock
\APACrefYearMonthDay{2015}{}{}.
\newblock
{\BBOQ}\APACrefatitle {Deep learning} {Deep learning}.{\BBCQ}
\newblock
\APACjournalVolNumPages{nature}{521}{7553}{436--444}.
\newblock

\newblock

\PrintBackRefs{\CurrentBib}

\bibitem [\protect \citeauthoryear {%
Li%
\ \protect \BOthers {.}}{%
Li%
\ \protect \BOthers {.}}{%
{\protect \APACyear {2021}}%
}]{%
cons_clus}
\APACinsertmetastar {%
cons_clus}%
\begin{APACrefauthors}%
Li, Y.%
, Hu, P.%
, Liu, Z.%
, Peng, D.%
, Zhou, J.T.%
\BCBL {} Peng, X.%
\end{APACrefauthors}%
\unskip\
\newblock
\APACrefYearMonthDay{2021}{}{}.
\newblock
{\BBOQ}\APACrefatitle {Contrastive Clustering} {Contrastive clustering}.{\BBCQ}
\newblock
 \APACrefbtitle {AAAI.} {Aaai.}
\PrintBackRefs{\CurrentBib}

\bibitem [\protect \citeauthoryear {%
Macqueen%
}{%
Macqueen%
}{%
{\protect \APACyear {1967}}%
}]{%
kmeans1967}
\APACinsertmetastar {%
kmeans1967}%
\begin{APACrefauthors}%
Macqueen, J.%
\end{APACrefauthors}%
\unskip\
\newblock
\APACrefYearMonthDay{1967}{}{}.
\newblock
{\BBOQ}\APACrefatitle {Some methods for classification and analysis of
  multivariate observations} {Some methods for classification and analysis of
  multivariate observations}.{\BBCQ}
\newblock
 \APACrefbtitle {In 5-th Berkeley Symposium on Mathematical Statistics and
  Probability} {In 5-th berkeley symposium on mathematical statistics and
  probability}\ (\BPGS\ 281--297).
\PrintBackRefs{\CurrentBib}

\bibitem [\protect \citeauthoryear {%
Masulli%
\ \BBA {} Schenone%
}{%
Masulli%
\ \BBA {} Schenone%
}{%
{\protect \APACyear {1999}}%
}]{%
masulli1999fuzzy}
\APACinsertmetastar {%
masulli1999fuzzy}%
\begin{APACrefauthors}%
Masulli, F.%
\BCBT {}\ \BBA {} Schenone, A.%
\end{APACrefauthors}%
\unskip\
\newblock
\APACrefYearMonthDay{1999}{}{}.
\newblock
{\BBOQ}\APACrefatitle {A fuzzy clustering based segmentation system as support
  to diagnosis in medical imaging} {A fuzzy clustering based segmentation
  system as support to diagnosis in medical imaging}.{\BBCQ}
\newblock
\APACjournalVolNumPages{Artificial intelligence in medicine}{16}{2}{129--147}.
\newblock

\newblock

\PrintBackRefs{\CurrentBib}

\bibitem [\protect \citeauthoryear {%
Ng%
, Jordan%
\BCBL {}\ \BBA {} Weiss%
}{%
Ng%
\ \protect \BOthers {.}}{%
{\protect \APACyear {2002}}%
}]{%
Spectral2002}
\APACinsertmetastar {%
Spectral2002}%
\begin{APACrefauthors}%
Ng, A.Y.%
, Jordan, M.I.%
\BCBL {} Weiss, Y.%
\end{APACrefauthors}%
\unskip\
\newblock
\APACrefYearMonthDay{2002}{}{}.
\newblock
{\BBOQ}\APACrefatitle {On Spectral Clustering: Analysis and an algorithm} {On
  spectral clustering: Analysis and an algorithm}.{\BBCQ}
\newblock
 T.G.~Dietterich, S.~Becker\BCBL {}\ \BBA {} Z.~Ghahramani\ (\BEDS),
  \APACrefbtitle {NeurIPS} {Neurips}\ (\BPGS\ 849--856).
\newblock
\APACaddressPublisher{}{MIT Press}.
\newblock
\begin{APACrefURL}
  {http://papers.nips.cc/paper/2092-on-spectral-clustering-analysis-and-an-algorithm.pdf}
  \end{APACrefURL}
\PrintBackRefs{\CurrentBib}

\bibitem [\protect \citeauthoryear {%
Niu%
, Shan%
\BCBL {}\ \BBA {} Wang%
}{%
Niu%
\ \protect \BOthers {.}}{%
{\protect \APACyear {2021}}%
}]{%
niu2021spice}
\APACinsertmetastar {%
niu2021spice}%
\begin{APACrefauthors}%
Niu, C.%
, Shan, H.%
\BCBL {} Wang, G.%
\end{APACrefauthors}%
\unskip\
\newblock
\APACrefYearMonthDay{2021}{}{}.
\newblock
{\BBOQ}\APACrefatitle {Spice: Semantic pseudo-labeling for image clustering}
  {Spice: Semantic pseudo-labeling for image clustering}.{\BBCQ}
\newblock
\APACjournalVolNumPages{arXiv preprint arXiv:2103.09382}{}{}{}.
\newblock

\newblock

\PrintBackRefs{\CurrentBib}

\bibitem [\protect \citeauthoryear {%
Niu%
, Zhang%
, Wang%
\BCBL {}\ \BBA {} Liang%
}{%
Niu%
\ \protect \BOthers {.}}{%
{\protect \APACyear {2020}}%
{\protect \APACexlab {{\protect \BCnt {1}}}}}]{%
niu2020gatcluster}
\APACinsertmetastar {%
niu2020gatcluster}%
\begin{APACrefauthors}%
Niu, C.%
, Zhang, J.%
, Wang, G.%
\BCBL {} Liang, J.%
\end{APACrefauthors}%
\unskip\
\newblock
\APACrefYearMonthDay{2020{\protect \BCnt {1}}}{}{}.
\newblock
{\BBOQ}\APACrefatitle {GATCluster: Self-Supervised Gaussian-Attention Network
  for Image Clustering} {Gatcluster: Self-supervised gaussian-attention network
  for image clustering}.{\BBCQ}
\newblock
\APACjournalVolNumPages{}{}{}{735--751}.
\newblock

\newblock

\PrintBackRefs{\CurrentBib}

\bibitem [\protect \citeauthoryear {%
Niu%
, Zhang%
, Wang%
\BCBL {}\ \BBA {} Liang%
}{%
Niu%
\ \protect \BOthers {.}}{%
{\protect \APACyear {2020}}%
{\protect \APACexlab {{\protect \BCnt {2}}}}}]{%
gatcluster}
\APACinsertmetastar {%
gatcluster}%
\begin{APACrefauthors}%
Niu, C.%
, Zhang, J.%
, Wang, G.%
\BCBL {} Liang, J.%
\end{APACrefauthors}%
\unskip\
\newblock
\APACrefYearMonthDay{2020{\protect \BCnt {2}}}{}{}.
\newblock
{\BBOQ}\APACrefatitle {GATCluster: Self-supervised Gaussian-Attention Network
  for Image Clustering} {Gatcluster: Self-supervised gaussian-attention network
  for image clustering}.{\BBCQ}
\newblock
 \APACrefbtitle {ECCV} {Eccv}\ (\BPGS\ 735--751).
\PrintBackRefs{\CurrentBib}

\bibitem [\protect \citeauthoryear {%
Regatti%
, Deshmukh%
, Manavoglu%
\BCBL {}\ \BBA {} Dogan%
}{%
Regatti%
\ \protect \BOthers {.}}{%
{\protect \APACyear {2021}}%
}]{%
regatti2020consensus}
\APACinsertmetastar {%
regatti2020consensus}%
\begin{APACrefauthors}%
Regatti, J.R.%
, Deshmukh, A.A.%
, Manavoglu, E.%
\BCBL {} Dogan, U.%
\end{APACrefauthors}%
\unskip\
\newblock
\APACrefYearMonthDay{2021}{}{}.
\newblock
{\BBOQ}\APACrefatitle {Consensus Clustering with Unsupervised Representation
  Learning} {Consensus clustering with unsupervised representation
  learning}.{\BBCQ}
\newblock
\APACjournalVolNumPages{International Joint Conference on Neural Networks
  (IJCNN) arXiv preprint arXiv:2010.01245}{}{}{}.
\newblock

\newblock

\PrintBackRefs{\CurrentBib}

\bibitem [\protect \citeauthoryear {%
Schops%
\ \protect \BOthers {.}}{%
Schops%
\ \protect \BOthers {.}}{%
{\protect \APACyear {2017}}%
}]{%
schops2017multi}
\APACinsertmetastar {%
schops2017multi}%
\begin{APACrefauthors}%
Schops, T.%
, Schonberger, J.L.%
, Galliani, S.%
, Sattler, T.%
, Schindler, K.%
, Pollefeys, M.%
\BCBL {} Geiger, A.%
\end{APACrefauthors}%
\unskip\
\newblock
\APACrefYearMonthDay{2017}{}{}.
\newblock
{\BBOQ}\APACrefatitle {A multi-view stereo benchmark with high-resolution
  images and multi-camera videos} {A multi-view stereo benchmark with
  high-resolution images and multi-camera videos}.{\BBCQ}
\newblock
 \APACrefbtitle {Proceedings of the IEEE Conference on Computer Vision and
  Pattern Recognition} {Proceedings of the ieee conference on computer vision
  and pattern recognition}\ (\BPGS\ 3260--3269).
\PrintBackRefs{\CurrentBib}

\bibitem [\protect \citeauthoryear {%
Shah%
\ \BBA {} Koltun%
}{%
Shah%
\ \BBA {} Koltun%
}{%
{\protect \APACyear {2018}}%
}]{%
shah2018deep}
\APACinsertmetastar {%
shah2018deep}%
\begin{APACrefauthors}%
Shah, S.A.%
\BCBT {}\ \BBA {} Koltun, V.%
\end{APACrefauthors}%
\unskip\
\newblock
\APACrefYearMonthDay{2018}{}{}.
\newblock
{\BBOQ}\APACrefatitle {Deep continuous clustering} {Deep continuous
  clustering}.{\BBCQ}
\newblock
\APACjournalVolNumPages{arXiv preprint arXiv:1803.01449}{}{}{}.
\newblock

\newblock

\PrintBackRefs{\CurrentBib}

\bibitem [\protect \citeauthoryear {%
Shorten%
\ \BBA {} Khoshgoftaar%
}{%
Shorten%
\ \BBA {} Khoshgoftaar%
}{%
{\protect \APACyear {2019}}%
}]{%
shorten2019survey}
\APACinsertmetastar {%
shorten2019survey}%
\begin{APACrefauthors}%
Shorten, C.%
\BCBT {}\ \BBA {} Khoshgoftaar, T.M.%
\end{APACrefauthors}%
\unskip\
\newblock
\APACrefYearMonthDay{2019}{}{}.
\newblock
{\BBOQ}\APACrefatitle {A survey on image data augmentation for deep learning}
  {A survey on image data augmentation for deep learning}.{\BBCQ}
\newblock
\APACjournalVolNumPages{Journal of Big Data}{6}{1}{1--48}.
\newblock

\newblock

\PrintBackRefs{\CurrentBib}

\bibitem [\protect \citeauthoryear {%
Strehl%
\ \BBA {} Ghosh%
}{%
Strehl%
\ \BBA {} Ghosh%
}{%
{\protect \APACyear {2002}}%
}]{%
strehl2002cluster}
\APACinsertmetastar {%
strehl2002cluster}%
\begin{APACrefauthors}%
Strehl, A.%
\BCBT {}\ \BBA {} Ghosh, J.%
\end{APACrefauthors}%
\unskip\
\newblock
\APACrefYearMonthDay{2002}{}{}.
\newblock
{\BBOQ}\APACrefatitle {Cluster ensembles---a knowledge reuse framework for
  combining multiple partitions} {Cluster ensembles---a knowledge reuse
  framework for combining multiple partitions}.{\BBCQ}
\newblock
\APACjournalVolNumPages{Journal of machine learning
  research}{3}{Dec}{583--617}.
\newblock

\newblock

\PrintBackRefs{\CurrentBib}

\bibitem [\protect \citeauthoryear {%
Tao%
, Takagi%
\BCBL {}\ \BBA {} Nakata%
}{%
Tao%
\ \protect \BOthers {.}}{%
{\protect \APACyear {2021}}%
}]{%
taoclustering}
\APACinsertmetastar {%
taoclustering}%
\begin{APACrefauthors}%
Tao, Y.%
, Takagi, K.%
\BCBL {} Nakata, K.%
\end{APACrefauthors}%
\unskip\
\newblock
\APACrefYearMonthDay{2021}{}{}.
\newblock
{\BBOQ}\APACrefatitle {CLUSTERING-FRIENDLY REPRESENTATION LEARN-ING VIA
  INSTANCE DISCRIMINATION AND FEATURE DECORRELATION} {Clustering-friendly
  representation learn-ing via instance discrimination and feature
  decorrelation}.{\BBCQ}
\newblock
\APACjournalVolNumPages{arXiv preprint arXiv:2106.00131}{}{}{}.
\newblock

\newblock

\PrintBackRefs{\CurrentBib}

\bibitem [\protect \citeauthoryear {%
Tian%
\ \protect \BOthers {.}}{%
Tian%
\ \protect \BOthers {.}}{%
{\protect \APACyear {2020}}%
}]{%
tian2020makes}
\APACinsertmetastar {%
tian2020makes}%
\begin{APACrefauthors}%
Tian, Y.%
, Sun, C.%
, Poole, B.%
, Krishnan, D.%
, Schmid, C.%
\BCBL {} Isola, P.%
\end{APACrefauthors}%
\unskip\
\newblock
\APACrefYearMonthDay{2020}{}{}.
\newblock
{\BBOQ}\APACrefatitle {What makes for good views for contrastive learning}
  {What makes for good views for contrastive learning}.{\BBCQ}
\newblock
\APACjournalVolNumPages{arXiv preprint arXiv:2005.10243}{}{}{}.
\newblock

\newblock

\PrintBackRefs{\CurrentBib}

\bibitem [\protect \citeauthoryear {%
Van~Gansbeke%
, Vandenhende%
, Georgoulis%
, Proesmans%
\BCBL {}\ \BBA {} Van~Gool%
}{%
Van~Gansbeke%
\ \protect \BOthers {.}}{%
{\protect \APACyear {2020}}%
}]{%
van2020scan}
\APACinsertmetastar {%
van2020scan}%
\begin{APACrefauthors}%
Van~Gansbeke, W.%
, Vandenhende, S.%
, Georgoulis, S.%
, Proesmans, M.%
\BCBL {} Van~Gool, L.%
\end{APACrefauthors}%
\unskip\
\newblock
\APACrefYearMonthDay{2020}{}{}.
\newblock
{\BBOQ}\APACrefatitle {Scan: Learning to classify images without labels} {Scan:
  Learning to classify images without labels}.{\BBCQ}
\newblock

\newblock

\newblock

\PrintBackRefs{\CurrentBib}

\bibitem [\protect \citeauthoryear {%
Vincent%
, Larochelle%
, Lajoie%
, Bengio%
\BCBL {}\ \BBA {} Manzagol%
}{%
Vincent%
\ \protect \BOthers {.}}{%
{\protect \APACyear {2010}}%
}]{%
SDAE2010}
\APACinsertmetastar {%
SDAE2010}%
\begin{APACrefauthors}%
Vincent, P.%
, Larochelle, H.%
, Lajoie, I.%
, Bengio, Y.%
\BCBL {} Manzagol, P.A.%
\end{APACrefauthors}%
\unskip\
\newblock
\APACrefYearMonthDay{2010}{}{}.
\newblock
{\BBOQ}\APACrefatitle {Stacked Denoising Autoencoders: Learning Useful
  Representations in a Deep Network with a Local Denoising Criterion.} {Stacked
  denoising autoencoders: Learning useful representations in a deep network
  with a local denoising criterion.}{\BBCQ}
\newblock
\APACjournalVolNumPages{Journal of Machine Learning
  Research}{11}{12}{3371-3408}.
\newblock

\newblock

\PrintBackRefs{\CurrentBib}

\bibitem [\protect \citeauthoryear {%
J.~Wu%
\ \protect \BOthers {.}}{%
J.~Wu%
\ \protect \BOthers {.}}{%
{\protect \APACyear {2019}}%
{\protect \APACexlab {{\protect \BCnt {1}}}}}]{%
wu2019deep}
\APACinsertmetastar {%
wu2019deep}%
\begin{APACrefauthors}%
Wu, J.%
, Long, K.%
, Wang, F.%
, Qian, C.%
, Li, C.%
, Lin, Z.%
\BCBL {} Zha, H.%
\end{APACrefauthors}%
\unskip\
\newblock
\APACrefYearMonthDay{2019{\protect \BCnt {1}}}{}{}.
\newblock
{\BBOQ}\APACrefatitle {Deep comprehensive correlation mining for image
  clustering} {Deep comprehensive correlation mining for image
  clustering}.{\BBCQ}
\newblock
 \APACrefbtitle {Proceedings of the IEEE International Conference on Computer
  Vision} {Proceedings of the ieee international conference on computer
  vision}\ (\BPGS\ 8150--8159).
\PrintBackRefs{\CurrentBib}

\bibitem [\protect \citeauthoryear {%
J.~Wu%
\ \protect \BOthers {.}}{%
J.~Wu%
\ \protect \BOthers {.}}{%
{\protect \APACyear {2019}}%
{\protect \APACexlab {{\protect \BCnt {2}}}}}]{%
Wu_2019_ICCV}
\APACinsertmetastar {%
Wu_2019_ICCV}%
\begin{APACrefauthors}%
Wu, J.%
, Long, K.%
, Wang, F.%
, Qian, C.%
, Li, C.%
, Lin, Z.%
\BCBL {} Zha, H.%
\end{APACrefauthors}%
\unskip\
\newblock
\APACrefYearMonthDay{2019{\protect \BCnt {2}}}{October}{}.
\newblock
{\BBOQ}\APACrefatitle {Deep Comprehensive Correlation Mining for Image
  Clustering} {Deep comprehensive correlation mining for image
  clustering}.{\BBCQ}
\newblock
 \APACrefbtitle {ICCV.} {Iccv.}
\PrintBackRefs{\CurrentBib}

\bibitem [\protect \citeauthoryear {%
Z.~Wu%
, Xiong%
, Yu%
\BCBL {}\ \BBA {} Lin%
}{%
Z.~Wu%
\ \protect \BOthers {.}}{%
{\protect \APACyear {2018}}%
}]{%
wu2018unsupervised}
\APACinsertmetastar {%
wu2018unsupervised}%
\begin{APACrefauthors}%
Wu, Z.%
, Xiong, Y.%
, Yu, S.X.%
\BCBL {} Lin, D.%
\end{APACrefauthors}%
\unskip\
\newblock
\APACrefYearMonthDay{2018}{}{}.
\newblock
{\BBOQ}\APACrefatitle {Unsupervised feature learning via non-parametric
  instance discrimination} {Unsupervised feature learning via non-parametric
  instance discrimination}.{\BBCQ}
\newblock
 \APACrefbtitle {Proceedings of the IEEE Conference on Computer Vision and
  Pattern Recognition} {Proceedings of the ieee conference on computer vision
  and pattern recognition}\ (\BPGS\ 3733--3742).
\PrintBackRefs{\CurrentBib}

\bibitem [\protect \citeauthoryear {%
Xie%
, Girshick%
\BCBL {}\ \BBA {} Farhadi%
}{%
Xie%
\ \protect \BOthers {.}}{%
{\protect \APACyear {2016}}%
{\protect \APACexlab {{\protect \BCnt {1}}}}}]{%
xie2016unsupervised}
\APACinsertmetastar {%
xie2016unsupervised}%
\begin{APACrefauthors}%
Xie, J.%
, Girshick, R.%
\BCBL {} Farhadi, A.%
\end{APACrefauthors}%
\unskip\
\newblock
\APACrefYearMonthDay{2016{\protect \BCnt {1}}}{}{}.
\newblock
{\BBOQ}\APACrefatitle {Unsupervised deep embedding for clustering analysis}
  {Unsupervised deep embedding for clustering analysis}.{\BBCQ}
\newblock
 \APACrefbtitle {International conference on machine learning} {International
  conference on machine learning}\ (\BPGS\ 478--487).
\PrintBackRefs{\CurrentBib}

\bibitem [\protect \citeauthoryear {%
Xie%
, Girshick%
\BCBL {}\ \BBA {} Farhadi%
}{%
Xie%
\ \protect \BOthers {.}}{%
{\protect \APACyear {2016}}%
{\protect \APACexlab {{\protect \BCnt {2}}}}}]{%
Xie2016}
\APACinsertmetastar {%
Xie2016}%
\begin{APACrefauthors}%
Xie, J.%
, Girshick, R.%
\BCBL {} Farhadi, A.%
\end{APACrefauthors}%
\unskip\
\newblock
\APACrefYearMonthDay{2016{\protect \BCnt {2}}}{}{}.
\newblock
{\BBOQ}\APACrefatitle {Unsupervised Deep Embedding for Clustering Analysis}
  {Unsupervised deep embedding for clustering analysis}.{\BBCQ}
\newblock
 \APACrefbtitle {ICML} {Icml}\ (\BPGS\ 478--487).
\newblock
\APACaddressPublisher{}{JMLR.org}.
\newblock
\begin{APACrefURL} {http://dl.acm.org/citation.cfm?id=3045390.3045442}
  \end{APACrefURL}
\PrintBackRefs{\CurrentBib}

\bibitem [\protect \citeauthoryear {%
Xu%
\ \BBA {} Wunsch%
}{%
Xu%
\ \BBA {} Wunsch%
}{%
{\protect \APACyear {2005}}%
}]{%
xu2005survey}
\APACinsertmetastar {%
xu2005survey}%
\begin{APACrefauthors}%
Xu, R.%
\BCBT {}\ \BBA {} Wunsch, D.%
\end{APACrefauthors}%
\unskip\
\newblock
\APACrefYearMonthDay{2005}{}{}.
\newblock
{\BBOQ}\APACrefatitle {Survey of clustering algorithms} {Survey of clustering
  algorithms}.{\BBCQ}
\newblock
\APACjournalVolNumPages{IEEE Transactions on neural networks}{16}{3}{645--678}.
\newblock

\newblock

\PrintBackRefs{\CurrentBib}

\bibitem [\protect \citeauthoryear {%
Yang%
, Parikh%
\BCBL {}\ \BBA {} Batra%
}{%
Yang%
\ \protect \BOthers {.}}{%
{\protect \APACyear {2016}}%
}]{%
Yang2016Joint}
\APACinsertmetastar {%
Yang2016Joint}%
\begin{APACrefauthors}%
Yang, J.%
, Parikh, D.%
\BCBL {} Batra, D.%
\end{APACrefauthors}%
\unskip\
\newblock
\APACrefYearMonthDay{2016}{}{}.
\newblock
{\BBOQ}\APACrefatitle {Joint Unsupervised Learning of Deep Representations and
  Image Clusters} {Joint unsupervised learning of deep representations and
  image clusters}.{\BBCQ}
\newblock
 \APACrefbtitle {CVPR.} {Cvpr.}
\PrintBackRefs{\CurrentBib}

\bibitem [\protect \citeauthoryear {%
Zeiler%
, Krishnan%
, Taylor%
\BCBL {}\ \BBA {} Fergus%
}{%
Zeiler%
\ \protect \BOthers {.}}{%
{\protect \APACyear {2010}}%
}]{%
Zeiler2010Deconvolutional}
\APACinsertmetastar {%
Zeiler2010Deconvolutional}%
\begin{APACrefauthors}%
Zeiler, M.D.%
, Krishnan, D.%
, Taylor, G.W.%
\BCBL {} Fergus, R.%
\end{APACrefauthors}%
\unskip\
\newblock
\APACrefYearMonthDay{2010}{}{}.
\newblock
{\BBOQ}\APACrefatitle {Deconvolutional networks} {Deconvolutional
  networks}.{\BBCQ}
\newblock
 \APACrefbtitle {Computer Vision and Pattern Recognition.} {Computer vision and
  pattern recognition.}
\PrintBackRefs{\CurrentBib}

\bibitem [\protect \citeauthoryear {%
Zhuang%
, Zhai%
\BCBL {}\ \BBA {} Yamins%
}{%
Zhuang%
\ \protect \BOthers {.}}{%
{\protect \APACyear {2019}}%
}]{%
zhuang2019local}
\APACinsertmetastar {%
zhuang2019local}%
\begin{APACrefauthors}%
Zhuang, C.%
, Zhai, A.L.%
\BCBL {} Yamins, D.%
\end{APACrefauthors}%
\unskip\
\newblock
\APACrefYearMonthDay{2019}{}{}.
\newblock
{\BBOQ}\APACrefatitle {Local aggregation for unsupervised learning of visual
  embeddings} {Local aggregation for unsupervised learning of visual
  embeddings}.{\BBCQ}
\newblock
 \APACrefbtitle {Proceedings of the IEEE International Conference on Computer
  Vision} {Proceedings of the ieee international conference on computer
  vision}\ (\BPGS\ 6002--6012).
\PrintBackRefs{\CurrentBib}

\end{thebibliography}
